\documentclass{article}

\PassOptionsToPackage{numbers, compress}{natbib}
\usepackage[preprint]{neurips_2026}
\makeatletter
\renewcommand{\@noticestring}{}
\makeatother

\usepackage[utf8]{inputenc}
\usepackage[T1]{fontenc}
\usepackage[hidelinks]{hyperref}
\usepackage{url}
\usepackage{booktabs}
\usepackage{array}
\usepackage{amsfonts}
\usepackage{amsmath}
\usepackage{amssymb}
\usepackage{amsthm}
\usepackage{nicefrac}
\usepackage{microtype}
\usepackage[table]{xcolor}
\usepackage{graphicx}
\usepackage{subcaption}
\usepackage{placeins}

\newcommand{\msci}[3]{\ensuremath{\vcenter{\hbox{\shortstack{#1\\[-1.5pt]{\scriptsize(#2,\,#3)}}}}}}
\newcommand{\rowstrut}{\rule[-8pt]{0pt}{20.8pt}}

\newtheorem{proposition}{Proposition}

\title{Practical Scaling Laws:\\ Converting Compute into Performance in a Data-Constrained World}

\author{%
  Christopher M. Bryant \\
  Arena Physica \\
  \texttt{chris@arenaphysica.com}
  \And
  Hao Liu \\
  Arena Physica \\
  \texttt{hao@arenaphysica.com}
}

\begin{document}

\maketitle

\begin{abstract}
The scaling laws guiding modern model training were calibrated for a single regime: data-rich, single-epoch pretraining. The dominant such scaling law form, Chinchilla's $L = E + A/N^\alpha + B/D^\beta$, has three structural limitations outside that regime: it diverges as unique data shrinks instead of saturating at the uninformed baseline; it cannot represent overfitting when capacity exceeds the data; and it conflates total examples seen with unique examples available. We propose a closed-form extension, $L(N, D, T) = E + (L_0 - E)\,h/(1+h)$ with $h = a/N^\alpha + b/T^\beta + c\,N^\gamma/D^\delta$, that decomposes loss into undercapacity, undertraining, and overfitting terms. It saturates between the irreducible loss $E$ and an uninformed baseline $L_0$ fixed by the loss type, and reduces to Chinchilla in the data-rich, single-epoch limit. We validate it on four multi-epoch experiments spanning four architecture families (MLPs, ResNets, Fourier neural operators, and transformers) across vision, scientific ML, and language domains, and refit it to five published LLM scaling-law grids. Extrapolating to higher compute and larger unique data than seen at fit time, our form achieves state-of-the-art RMSE on every published LLM grid we evaluate and on most cells of our constructed experiments. Once calibrated, the form admits a cost-aware allocation that recovers Chinchilla's optimum when data is free and shifts toward smaller corpora and more epochs as data grows expensive.
\end{abstract}

\section{Introduction}

For most modern ML, training data is expensive. Physics surrogates require running an expensive PDE solver to generate each example~\citep{ashton2025fluid}; medical-imaging foundation models require expert-labeled data~\citep{ilse2025radiology}; reinforcement-learning agents pay for every transition~\citep{pearce2024scaling}; LLM finetuning operates at fractions of pretraining data budgets~\citep{zhang2024when}. Even frontier LLM pretraining, where data has historically been the cheap input relative to compute, is starting to revisit tokens as scraped text runs out~\citep{villalobos2024run,muennighoff2023scaling}. In all of these settings, the practitioner faces the same tradeoff: spend dollars to acquire more unique data, or spend compute to revisit the data already in hand. Reasoning about that tradeoff is what scaling laws are for, but the dominant scaling law was calibrated in a regime where the tradeoff did not exist.

Scaling laws fit a parametric model of the loss surface from cheap small-scale runs and use it to predict loss at scales practitioners cannot afford to run. The most-cited instance is the Chinchilla form~\citep{hoffmann2022training}:
\begin{equation}
L(N, D) = E + \frac{A}{N^\alpha} + \frac{B}{D^\beta}.
\label{eq:chinchilla}
\end{equation}
Variants of this power-law-plus-constant template recur across language modeling~\citep{kaplan2020scaling}, vision~\citep{alabdulmohsin2022revisiting}, transfer learning~\citep{hernandez2021scaling}, downstream tasks~\citep{caballero2023broken,gadre2025language}, and theoretical models~\citep{bahri2024explaining,maloney2022solvable,bordelon2024dynamical}, and inherit its structural commitments.

\paragraph{Three structural failures.} Though Equation~\eqref{eq:chinchilla} and its close relatives have been used to fit training runs across many orders of magnitude in the data-rich, single-epoch regime, these forms lack three crucial properties needed to generalize beyond this regime:
\begin{itemize}
\itemsep -0.05em
  \item \textbf{No baseline saturation.} The loss should be bounded above by $L_0$, the loss of a model that has not learned anything. Chinchilla has $L \to \infty$ as $D \to 0$ instead of saturating at $L_0$.
  \item \textbf{Overfitting unreachable.} At fixed $D$, the form is strictly decreasing in $N$, but neural networks with $N \gg D$ overfit~\citep{hernandez2022repeated,muennighoff2023scaling}: validation loss can rise with capacity past a point.
  \item \textbf{No compute axis independent of $(N, D)$.} The form cannot distinguish between total examples seen $T$ and unique examples $D$, precluding the analysis of multi-epoch training runs.
\end{itemize}  

The paper's contributions are:
\begin{itemize}
\itemsep -0.05em
  \item A single parametric form for $L(N, D, T)$ on $\mathbb{R}_{>0}^3$ that addresses these failures by decomposing loss into undercapacity, undertraining, and overfitting terms wrapped in a saturating bound between the irreducible loss $E$ and an uninformed baseline $L_0$ (Section~\ref{sec:form}).
  \item A set of scaling experiments testing the form across four architecture-domain pairs that span vision, scientific machine learning, and language.
  \item An empirical demonstration of state-of-the-art extrapolation capabilities across prior published LLM scaling-law datasets.
  \item A strategy that chooses $(N^*, D^*, T^*)$ to minimize loss given a budget constraint, while optimally allocating that budget between data acquisition and training compute.
\end{itemize}

\section{The extended scaling law}
\label{sec:form}

Motivated by the boundary failures of Equation~\eqref{eq:chinchilla}, we enumerate a list of limiting behaviors (Table~\ref{tab:limits}) that any form in $(N, D, T)$ should exhibit when $N$, $D$, or $T$ goes to either $0$ or $\infty$. Rows 1--5 require $L \to L_0$ when training collapses and row 6 characterizes optimal behavior.

\paragraph{Proposed form.} We seek a simple closed-form extension of the Chinchilla functional form that obeys the desired limiting behaviors, decomposes into terms with clear physical meaning, and reduces to Equation~\eqref{eq:chinchilla} in the small-$h$ single-epoch limit. Working from those constraints, we arrive at the following form (full derivations are in Appendix~\ref{app:proposed-form}):
\begin{equation}
L(N, D, T) = E + (L_0 - E) \cdot \frac{h(N, D, T)}{1 + h(N, D, T)},
\qquad
h = \frac{a}{N^\alpha} + \frac{b}{T^\beta} + c \, \frac{N^\gamma}{D^\delta}.
\label{eq:ours}
\end{equation}

\paragraph{Notation.} $N$ is parameters, $D$ unique training examples, $T$ total training examples seen (counted with repetition). The three terms in $h$ represent \emph{undercapacity} ($a/N^\alpha$), \emph{undertraining} ($b/T^\beta$), and \emph{overfitting} ($c\,N^\gamma/D^\delta$); the saturating wrapper $h/(1+h)$ bounds $L$ in $[E, L_0]$, where $E$ is the irreducible loss and $L_0$ is the uninformed baseline loss. When necessary for analysis, compute $C$ is approximated as $C \approx kNT$ with architecture-dependent constant $k$ (though the form makes no assumption about the functional relationship between $C$, $N$, and $T$), and $T/D$ is the number of training epochs. The eight free parameters are $(E, a, b, c, \alpha, \beta, \gamma, \delta)$.

\paragraph{What $L$ represents.} Throughout, $L(N, D, T)$ is the \emph{lowest validation loss seen during training}: the minimum over checkpoints, not the final-checkpoint value. The convention matters in the overfitting regime ($N \gg D$), where validation loss can rise as the model memorizes; we fit to the achievable minimum, which is what a real deployment would early-stop at.

\paragraph{Scope of a fit.} A single fit of $L$ is conditional on a fixed architecture family, training-set distribution, and training recipe (optimizer, LR schedule, regularization, batch size, hyperparameter-selection protocol). These choices reshape the loss surface and are absorbed into the fitted constants.

\begin{table}[t]
\centering
\caption{Target limiting behaviors for a parametric loss $L(N, D, T)$. Rows where Chinchilla (Eq.~\eqref{eq:chinchilla}) fails are marked; our form (Eq.~\eqref{eq:ours}) satisfies all rows.}
\label{tab:limits}
\resizebox{\linewidth}{!}{%
\begin{tabular}{@{}clll@{\hspace{1em}}l@{}}
\toprule
\# & Limit & Expected $L$ & Reason & Chinchilla \\
\midrule
1 & $N \to 0$; $D$ and $T$ finite & $L_0$ & No capacity to represent any distribution & \textbf{fails} ($L \to \infty$) \\
2 & $D \to 0$; $N$ and $T$ finite & $L_0$ & No training data & \textbf{fails} ($L \to \infty$) \\
3 & $T \to 0$; $N$ and $D$ finite & $L_0$ & No training, random initialization & \textbf{fails} ($T = D$, so $L \to \infty$ as $D \to 0$) \\
4 & $N \to \infty$; $T$ and $D$ finite & $L_0$ & Infinite-capacity model, finite training & \textbf{fails} ($L \to E + B/D^\beta$) \\
5 & $N, T \to \infty$; $D$ finite & $L_0$ & Complete overfitting & \textbf{fails} (unreachable, since $T = D$) \\
6 & $N, D, T \to \infty$ jointly & $E$ & Irreducible loss reachable when no resource bottlenecks & passes \\
\bottomrule
\end{tabular}%
}
\end{table}

\subsection{Three decomposed terms}

The difficulty $h$ is a sum of power laws that represent three distinct underperformance phenomena:
\begin{equation}
h(N, D, T) = \underbrace{\frac{a}{N^\alpha}}_\text{undercapacity} + \underbrace{\frac{b}{T^\beta}}_\text{undertraining} + \underbrace{c \, \frac{N^\gamma}{D^\delta}}_\text{overfitting}.
\label{eq:h}
\end{equation}

\emph{Undercapacity} $a/N^\alpha$ is the only term that does not vanish as $D$ and $T$ grow without bound. It captures the residual cost of a model with finite capacity, and its exponent $\alpha$ is the standard Chinchilla capacity exponent. The precise way that $N$ maps to capacity depends on the model architecture, but within a given architecture, capacity increases monotonically with $N$.

\emph{Undertraining} $b/T^\beta$ decays with the total examples seen $T$. In practice, $T$ is often taken as $\approx C/(kN)$ with $k$ architecture-dependent, but the form does not require any strict relationship between $C$ and $T$. Per Chinchilla single-epoch fits, $\beta$ is the data-axis exponent: at $T = D$, $b/T^\beta$ collapses to $b/D^\beta$ and recovers the Chinchilla data term. Outside the single-epoch grid the two mechanisms separate: undertraining decays in $T$ regardless of whether each gradient step sees a fresh example or a repeat.

\emph{Overfitting} $c\,N^\gamma/D^\delta$ depends only on capacity $N$ and unique-data count $D$, not on training duration. The two-exponent shape is more flexible than the symmetric ratio $(N/D)^\gamma$: it allows $D$-scaling and $N$-scaling to be calibrated independently, since they enter generalization through different mechanisms (sample-size variance of the empirical-risk minimizer in $D$, and capacity-driven excess representation in $N$), each with its own scaling rate.

At fixed $D$, the interplay between the overfitting term $cN^\gamma/D^\delta$ and undercapacity term $a/N^\alpha$ produces an interior minimum in $L(N)$ in the limit as $T \to \infty$ (Appendix~\ref{app:nopt}), which implies that for a finite data size, there is a model size beyond which the best achievable performance not only saturates, but \emph{degrades}. This U-shaped behavior is a structural feature absent from the other parametric forms surveyed in Section~\ref{sec:related}. We provide additional remarks on the relationship between ``undercapacity'' and ``overfitting'' in Appendix~\ref{app:terms}.

\subsection{The saturating wrapper}

The wrapper $h / (1 + h)$ is a monotone bijection from $[0, \infty)$ to $[0, 1)$ with two properties we want:
\begin{enumerate}
\itemsep -0.05em
  \item \emph{Small-$h$ linearity.} For $h \ll 1$, $h/(1+h) = h - h^2 + h^3 - \cdots$, so Equation~\eqref{eq:ours} reduces at leading order to $L \approx E + (L_0 - E) h$. This is the regime in which Chinchilla was fit, and our form becomes Chinchilla-like there: the $(L_0 - E)$ prefactor is absorbed into the fitted coefficients of the power-law terms in $h$.
  \item \emph{Large-$h$ saturation.} For $h \to \infty$, $h/(1+h) \to 1$ and $L \to L_0$. The form cannot exceed the baseline.
\end{enumerate}
Other wrappers satisfy the same endpoints; we compare against the natural alternative $1 - e^{-h}$ in Appendix~\ref{app:wrapper}.

\subsection{The baseline loss \texorpdfstring{$L_0$}{L0}}

$L_0$ is the loss of an \emph{uninformed baseline}: a model that extracts no information from the data. It is determined by the loss type and the output normalization, not fitted to the loss surface. For cross-entropy over $K$ possible outcomes, $L_0 = \ln K$ (Appendix~\ref{app:l0}): $K = V$ for next-token prediction over a vocabulary of size $V$, and $K$ is the number of classes for classification. For relative-$L_2$ regression on $z$-normalized targets, $L_0 = 1$ exactly, since zero predictions give $\|0 - u\|_2/\|u\|_2 = 1$. For other bounded losses, $L_0$ is the corresponding uninformed-baseline limit. We use $L_0$ as a structural upper bound on the loss surface; per-dataset values are listed in Section~\ref{sec:empirical-setup}.

\subsection{Qualitative behavior}
\label{sec:qualitative-behavior}

\begin{figure}[t]
  \centering
  \includegraphics[width=\linewidth]{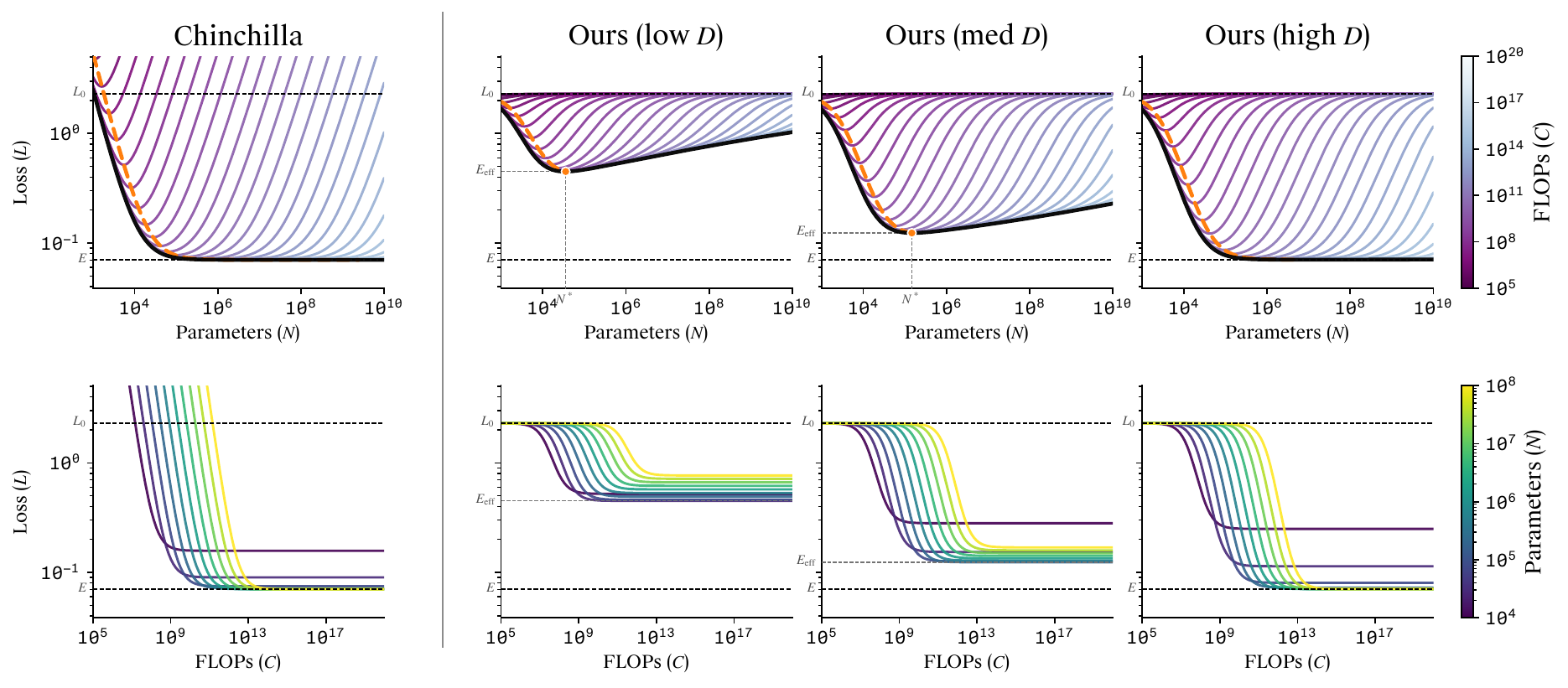}
  \caption{Qualitative shape of our form (right three columns) compared to Chinchilla (left). \emph{Top}: loss vs $N$ at fixed-$C$ iso-curves; the black curve is the infinite-compute envelope (the $C \to \infty$ limit isoFLOP), and the orange dashed curve is the compute-optimal envelope (the locus of isoFLOP minima). \emph{Bottom}: loss vs $C$ at fixed $N$. Chinchilla is unbounded above and admits no overfitting behavior. Our form saturates at the uninformed baseline $L_0$, has an interior minimum at $N^*$ where overfitting meets undercapacity (the floor $E_\text{eff}>E$), and recovers Chinchilla's compute-optimal envelope as $D$ grows; the floor $E_\text{eff}$ walks toward $E$ as $D$ increases.}
  \label{fig:theory-grid}
\end{figure}

To gain intuition for the form, in Figure~\ref{fig:theory-grid}, we contrast our loss surface with Chinchilla's. At a low, medium, and high value of $D$, we plot two complementary views: isoFLOP curves $L(N \mid C)$ at fixed compute, and training trajectories $L(C \mid N)$ at fixed model size.

\emph{IsoFLOP curves $L(N \mid C)$.} Chinchilla's isoFLOPs are U-shaped: they diverge to infinity on both tails and bottom out at a single minimum that defines the compute-optimal model size for that budget. As $C$ grows, this minimum shifts rightward toward larger model sizes and its loss approaches the irreducible floor $E$. Our form's isoFLOPs share that U-shape near the optimum, but differ at both extremes. The tails of our isoFLOPs saturate above at $L_0$, and for any finite $D$ they approach a different minimum loss value $E_\text{eff} > E$ at a critical model size $N^*$. Beyond that model size, the best achievable loss increases, even with infinite compute. As $D$ grows, $N^*$ shifts rightward and $E_\text{eff}$ descends toward $E$. In the infinite-data-and-compute limit away from $L_0$, the envelope of our isoFLOPs coincides with Chinchilla's.

\emph{Training trajectories $L(C \mid N)$.} By holding $N$ fixed instead, we can watch the loss evolve as compute increases. Chinchilla's trajectories begin at infinity and decay monotonically: smaller models drop faster early on but settle at higher plateaus than larger ones. Ours instead begin at the finite ceiling $L_0$ and likewise decay to an asymptote, but as $N$ grows, that asymptote first falls, reaching its minimum $E_\text{eff}$ at $N = N^*$, then rises again.

\section{Experiments}
\label{sec:experiments}

\begin{figure}[t]
  \centering
  \includegraphics[width=\linewidth]{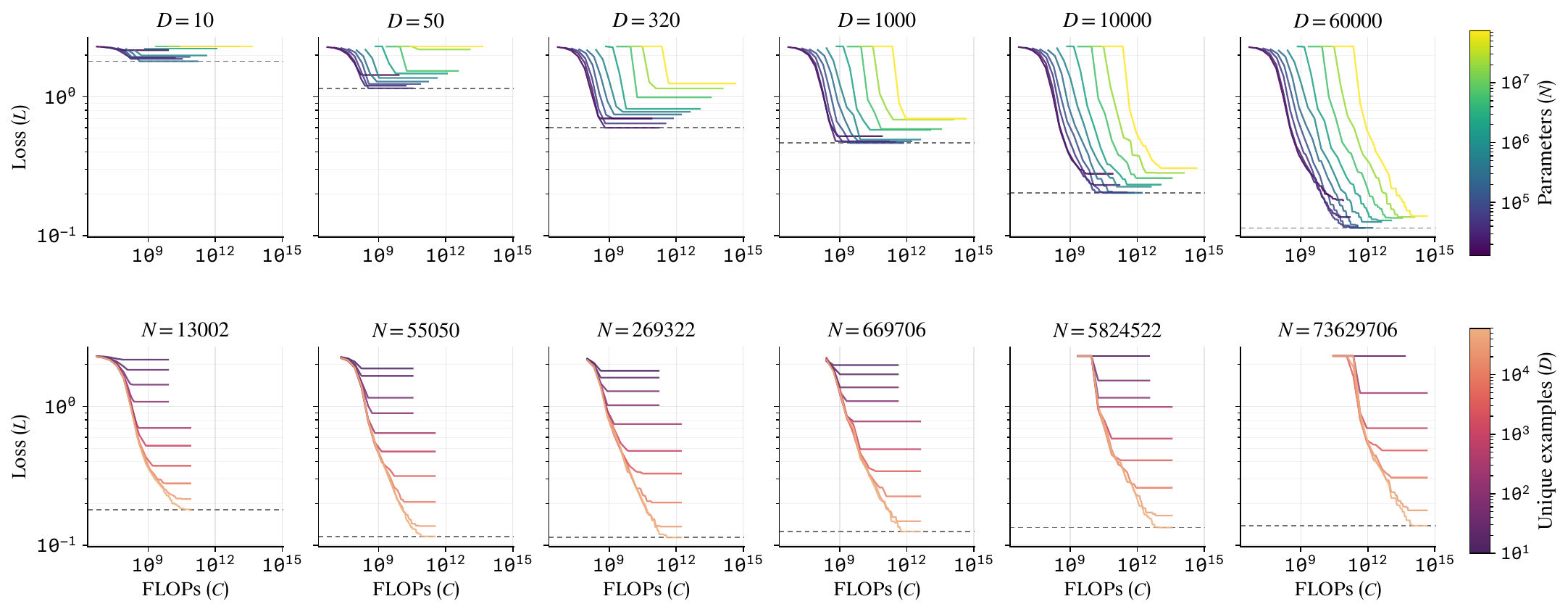}
  \caption{Empirical MNIST running-min validation-loss surface across a selection of $(N, D, C)$ cells. \emph{Top}: loss vs FLOPs at six fixed unique-data sizes $D \in \{10, 50, 320, 1000, 10000, 60000\}$, colored by $N$. \emph{Bottom}: the same surface at six fixed $N$, colored by $D$. The loss floor descends with increasing $D$ and models exhibit overfitting with increasing $N$. For $D=60\text{k}$, $N^*$ appears around 250k.}
  \label{fig:mnist-facets}
\end{figure}

We empirically validate our proposed form by comparing it to prior forms across a variety of different calibration sets of $(N, D, T)$ points. We evaluate extrapolation performance on held-out high-$C$ or high-$D$ runs since extrapolation to these unseen regimes is the primary practical use of a scaling law. Our validation consists of two distinct sets of experiments:

\paragraph{Multi-domain, multi-epoch experiments (Appendix~\ref{app:our-experiments}).} First, we train four different families of model architectures in four different domains and control the full $(N, D, T)$ sweep, spanning a regime and design space largely unexplored by prior scaling-law work: \emph{MLPs on MNIST} (10-way image classification: a toy problem where every limiting behavior is cheap to access; Figure~\ref{fig:mnist-facets}), \emph{PreActResNets~\citep{he2016identity} on CIFAR-100} (100-way image classification), \emph{Fourier Neural Operators on PDEBench 2D Darcy flow}~\citep{li2021fourier,takamoto2022pdebench} (physical PDE simulation), and \emph{decoder-only transformers on TinyStories}~\citep{eldan2023tinystories} (next-token prediction). In each of these domains, we fit our proposed form to the loss surface to evaluate the quality of the fit and the extrapolation error of the form, demonstrating its broad applicability. Full experimental details are in Appendix~\ref{app:our-experiments}.

\paragraph{Refits to prior LLM scaling studies (Appendix~\ref{app:external-refits}).} Second, since the majority of prior scaling-law work focuses on language modeling with transformer architectures~\citep{kaplan2020scaling,hoffmann2022training,muennighoff2023scaling,gadre2025language,li2025farseer}, we refit our form to five published LLM scaling-law grids, each chosen to probe a qualitatively different regime: \emph{Chinchilla}~\citep{hoffmann2022training} as the canonical compute-optimal reference; \emph{Muennighoff}~\citep{muennighoff2023scaling} for multi-epoch repetition (the only public grid with $T > D$); \emph{Gadre}~\citep{gadre2025language} for the over-trained regime ($D/N$ up to $\sim\!640$); \emph{Porian}~\citep{porian2024resolving} as a Kaplan stand-in (Kaplan's raw data is not public; Porian re-runs Chinchilla-style training under a Kaplan-style LR schedule on RefinedWeb); and \emph{Farseer}~\citep{li2025farseer} as the largest open single-recipe grid at modern scale. Dataset and grid descriptions are in Appendix~\ref{app:external-refits}.

\subsection{Setup}
\label{sec:empirical-setup}

\paragraph{Forms compared.} On each dataset, we evaluate our proposed form (Equation~\eqref{eq:ours}) against five refitted parametric scaling law baselines: \emph{Chinchilla}~\citep{hoffmann2022training} (the canonical additive $E + A/N^\alpha + B/D^\beta$ form), \emph{Muennighoff}~\citep{muennighoff2023scaling} (extends Chinchilla to multi-epoch via an exponentially-saturating effective-data substitution $D \to D'$), \emph{M4}~\citep{alabdulmohsin2022revisiting} (single-axis implicit form bounding $L$ in $[E, L_0]$; closest structural analog of our wrapper), \emph{BNSL}~\citep{caballero2023broken} ($3{+}3k$-parameter smoothly-broken power-law in a composite scalar; the only baseline able to fit non-monotone $L(N)$, via $k$ empirical breakpoints), and \emph{Farseer}~\citep{li2025farseer} ($L(N, D)$ form with $N$-dependent exponents; reports state-of-the-art far-compute extrapolation, which we reproduce in Appendix~\ref{app:far-extrapolation}). Appendix~\ref{app:forms} lists each form's equation, motivation, and free parameters. We also evaluate four ablations of our form in Appendix~\ref{app:full-results}: dropping the saturating wrapper, dropping the overfitting term, swapping $h/(1+h)$ for $1{-}e^{-h}$, and replacing the two-exponent overfitting term $c\,N^\gamma/D^\delta$ with a single-exponent ratio $c\,(N/D)^\gamma$.

\paragraph{Baseline loss values.} Following the loss-type rule of Section~\ref{sec:form}, we fix $L_0$ per dataset: $\ln 10 \approx 2.30$ for MNIST and $\ln 100 \approx 4.61$ for CIFAR-100 (classification), $L_0 = 1$ for Darcy (relative-$L_2$ on $z$-normalized targets), $\ln 2000 \approx 7.60$ for TinyStories (our custom 2K BPE tokenizer trained on the TinyStories corpus), and for the public LLM grids $\ln 32{,}000 \approx 10.37$ (Chinchilla, Porian; SentencePiece), $\ln 50{,}257 \approx 10.82$ (Muennighoff; GPT-NeoX/GPT-2 vocabulary), $\ln 50{,}432 \approx 10.83$ (Gadre), and $\log_2(65{,}536)/4 = 4.0$ for Farseer, whose grid we score in bits per character rather than nats, to match the statistic its authors report.

\paragraph{Training recipe.} The form's fitted constants are recipe-dependent (Section~\ref{sec:form}), so we assume a single training recipe within each experiment family and report exponents and coefficients conditional on that recipe. For three of our four training experiments, following~\citet{hagele2024scaling}, we adopt a Warmup-Stable-Decay (WSD) schedule with brief cooldown forks at log-spaced compute targets, so a single stable run feeds multiple final-loss measurements at different total-compute budgets. We train our MNIST models with no LR schedule at all, so every step is a valid end state and no cooldown forks are needed. Per-family optimizer, LR, batch size, and warmup choices are in Appendix~\ref{app:our-experiments}. For the five published LLM grids we filter each source down to a single training recipe before fitting; per-grid filter rules are in Appendix~\ref{app:external-refits}.

\paragraph{Form fitting.} Each form's free parameters are fit by minimizing Huber loss on log-residuals with BFGS, with $L_0$ fixed by the loss type and multistart random initializations to mitigate local optima (Appendix~\ref{app:optimization}). Full details are in Appendix~\ref{app:fitting}.

\paragraph{Extrapolation protocols.} Both holdouts target $\sim$10\% of rows, constructed \emph{groupwise} on the relevant axis so no axis value is split across folds. \emph{high-$C$ holdout} (predicting expensive runs from cheap ones): groupwise on $C$, effectively the largest-$C$ 10\% of rows since $C$ is near-continuous. \emph{high-$D$ holdout} (predicting large-$D$ performance from small-$D$ data): groupwise on $D$; since $D$ is highly discrete, the realized holdout is the top-$D$ group for all four of our experiments (ranging from 8\% to 20\%).

\paragraph{Evaluation.} We report the root-mean-square residual in log loss over the $M$ held-out rows,
\begin{equation}
\mathrm{RMSE}_{\log} \;=\; \sqrt{\frac{1}{M} \sum\nolimits_{i=1}^{M} \big(\ln \hat{L}_i - \ln L_i\big)^2},
\label{eq:rmse-log}
\end{equation}
where $L_i$ is row $i$'s observed loss and $\hat{L}_i$ the fitted form's prediction. Every reported interval is the 2.5\%--97.5\% quantile range over 200 bootstrap resamples of the training set, each refit from scratch. When comparing two forms directly to each other, we report a paired win-rate: draw a bootstrap resample of the training set, fit the two forms to it, note which form has the lower RMSE, repeat 200 times, and report the fraction of times the first form has the lower RMSE. Where a win-rate is called significant, the $p$-value comes from an exact two-sided sign test of the win count against $\mathrm{Bin}(200, \tfrac{1}{2})$.

\subsection{Results}
\label{sec:empirical-results}

Figure~\ref{fig:mnist-facets} shows MNIST training run validation loss curves as a function of training FLOPs (averaged across 10 random seeds to reduce noise), split by $N$ and $D$. These curves exhibit the qualitative behavior we anticipate in Section~\ref{sec:qualitative-behavior}, consistent with the shape predicted by the proposed form. Table~\ref{tab:holdout-axis-headline} reports held-out RMSE in log space across all forms under the high-$C$ and high-$D$ protocols defined above. Table~\ref{tab:paired-diff} reports the paired win-rates of our form compared to each of the other forms. For completeness, we provide an ablation of all parts of our form, as well as random 5-fold cross-validation and in-sample metrics, in Appendix~\ref{app:full-results}.

\begin{table}[p]
\centering
\caption{\textbf{high-$C$ / high-$D$ extrapolation.} Held-out RMSE (log space, Eq.~\eqref{eq:rmse-log}) under the holdout protocols of \S\ref{sec:empirical-setup}. Lower is better. Per column within each pane, \textbf{best} and \underline{second best} mark the median. Values are the median over 200 bootstrap resamples of the training set, each refit from scratch; parentheses give their 2.5\% and 97.5\% quantiles.}
\label{tab:holdout-axis-headline}
\resizebox{\linewidth}{!}{%
\setlength{\aboverulesep}{0pt}\setlength{\belowrulesep}{0pt}%
\begin{tabular}{@{}>{\rowstrut}l|cccc|ccccc@{}}
\toprule
\rowcolor{black!10} \multicolumn{10}{c}{\rowstrut \textbf{high-$C$ holdout}} \\
\midrule
Form $\downarrow$ & MNIST & CIFAR-100 & Darcy & TinyStories & Chinchilla & Muennighoff & Gadre & Porian & Farseer \\
\midrule
Chinchilla~\citep{hoffmann2022training}      & \msci{0.331}{0.328}{0.334} & \msci{0.156}{0.137}{0.184} & \msci{0.322}{0.281}{0.380} & \msci{0.167}{0.134}{0.222} & \msci{0.023}{0.016}{0.027} & \msci{0.093}{0.072}{0.115} & \msci{0.037}{0.013}{0.060} & \msci{0.098}{0.071}{0.125} & \msci{0.022}{0.020}{0.025} \\
Muennighoff~\citep{muennighoff2023scaling}   & \msci{0.335}{0.316}{1.126} & \underline{\msci{0.091}{0.077}{0.142}} & \underline{\msci{0.195}{0.123}{0.267}} & \textbf{\msci{0.079}{0.057}{0.273}} & \msci{0.019}{0.012}{0.044} & \underline{\msci{0.083}{0.068}{0.102}} & \underline{\msci{0.024}{0.008}{0.055}} & \underline{\msci{0.083}{0.023}{0.125}} & \msci{0.017}{0.014}{0.107} \\
M4 ($D$-axis)~\citep{alabdulmohsin2022revisiting} & \underline{\msci{0.295}{0.267}{0.299}} & \msci{0.125}{0.096}{0.180} & \msci{0.404}{0.364}{0.475} & \msci{0.331}{0.287}{0.412} & \msci{0.067}{0.058}{0.080} & \msci{0.094}{0.077}{0.128} & \msci{0.121}{0.071}{0.183} & \msci{0.112}{0.097}{0.145} & \msci{0.104}{0.091}{0.115} \\
BNSL $k{=}2$~\citep{caballero2023broken}     & \msci{0.308}{0.305}{0.313} & \msci{0.138}{0.092}{0.191} & \msci{0.411}{0.367}{0.483} & \msci{0.336}{0.289}{0.420} & \msci{0.067}{0.058}{0.082} & \msci{0.095}{0.076}{0.126} & \msci{0.127}{0.070}{0.304} & \msci{0.122}{0.102}{0.181} & \msci{0.104}{0.091}{0.117} \\
Farseer~\citep{li2025farseer}                & \msci{1.363}{0.989}{1.497} & \msci{0.621}{0.606}{0.875} & \msci{0.343}{0.305}{0.535} & \underline{\msci{0.127}{0.110}{0.261}} & \underline{\msci{0.015}{0.009}{0.032}} & \msci{0.098}{0.092}{0.132} & \msci{0.037}{0.014}{0.114} & \msci{0.084}{0.050}{0.144} & \underline{\msci{0.007}{0.003}{0.021}} \\
\midrule
\textbf{Ours} (Eq.~\eqref{eq:ours})          & \textbf{\msci{0.127}{0.125}{0.129}} & \textbf{\msci{0.083}{0.067}{0.108}} & \textbf{\msci{0.169}{0.140}{0.200}} & \msci{0.183}{0.150}{0.232} & \textbf{\msci{0.008}{0.005}{0.012}} & \textbf{\msci{0.060}{0.046}{0.097}} & \textbf{\msci{0.014}{0.005}{0.091}} & \textbf{\msci{0.063}{0.045}{0.092}} & \textbf{\msci{0.005}{0.004}{0.007}} \\
\midrule[\heavyrulewidth]
\addlinespace[1em]
\midrule[\heavyrulewidth]
\rowcolor{black!10} \multicolumn{10}{c}{\rowstrut \textbf{high-$D$ holdout}} \\
\midrule
Form $\downarrow$ & MNIST & CIFAR-100 & Darcy & TinyStories & Chinchilla & Muennighoff & Gadre & Porian & Farseer \\
\midrule
Chinchilla~\citep{hoffmann2022training}      & \textbf{\msci{0.124}{0.122}{0.167}} & \msci{0.179}{0.148}{0.215} & \msci{0.385}{0.327}{0.455} & \msci{0.063}{0.041}{0.094} & \msci{0.027}{0.018}{0.032} & \msci{0.109}{0.095}{0.140} & \msci{0.037}{0.013}{0.060} & \msci{0.114}{0.084}{0.148} & \msci{0.015}{0.014}{0.018} \\
Muennighoff~\citep{muennighoff2023scaling}   & \msci{0.156}{0.115}{0.171} & \underline{\msci{0.121}{0.060}{0.283}} & \underline{\msci{0.209}{0.138}{0.369}} & \msci{0.063}{0.034}{0.157} & \msci{0.021}{0.009}{0.026} & \underline{\msci{0.069}{0.037}{0.101}} & \underline{\msci{0.024}{0.008}{0.055}} & \msci{0.104}{0.047}{0.145} & \msci{0.014}{0.013}{0.065} \\
M4 ($D$-axis)~\citep{alabdulmohsin2022revisiting} & \msci{0.193}{0.186}{0.313} & \msci{0.262}{0.244}{0.297} & \msci{0.361}{0.350}{0.414} & \msci{0.222}{0.210}{0.271} & \msci{0.038}{0.036}{0.054} & \msci{0.101}{0.097}{0.129} & \msci{0.121}{0.071}{0.183} & \underline{\msci{0.081}{0.078}{0.097}} & \msci{0.064}{0.062}{0.075} \\
BNSL $k{=}2$~\citep{caballero2023broken}     & \msci{0.173}{0.168}{0.444} & \msci{0.283}{0.244}{0.352} & \msci{0.368}{0.349}{0.610} & \msci{0.221}{0.209}{0.285} & \msci{0.038}{0.036}{0.056} & \msci{0.103}{0.097}{0.135} & \msci{0.127}{0.070}{0.304} & \msci{0.090}{0.079}{0.136} & \msci{0.064}{0.061}{0.076} \\
Farseer~\citep{li2025farseer}                & \msci{1.484}{1.449}{1.513} & \msci{0.883}{0.668}{1.073} & \msci{0.471}{0.371}{0.591} & \underline{\msci{0.062}{0.035}{0.146}} & \underline{\msci{0.013}{0.007}{0.030}} & \msci{0.092}{0.058}{0.149} & \msci{0.037}{0.014}{0.114} & \msci{0.101}{0.080}{0.136} & \textbf{\msci{0.007}{0.006}{0.021}} \\
\midrule
\textbf{Ours} (Eq.~\eqref{eq:ours})          & \underline{\msci{0.137}{0.128}{0.147}} & \textbf{\msci{0.078}{0.052}{0.120}} & \textbf{\msci{0.171}{0.145}{0.200}} & \textbf{\msci{0.056}{0.035}{0.090}} & \textbf{\msci{0.008}{0.006}{0.012}} & \textbf{\msci{0.046}{0.039}{0.073}} & \textbf{\msci{0.014}{0.005}{0.091}} & \textbf{\msci{0.033}{0.028}{0.043}} & \textbf{\msci{0.007}{0.006}{0.008}} \\
\bottomrule
\end{tabular}
}
\end{table}

\begin{table}[p]
\centering
\caption{\textbf{Paired bootstrap comparison against \textbf{Ours} (Eq.~\eqref{eq:ours}).} Every form is refit on the same 200 resampled training sets, so each baseline is compared to ours within the same resample. Each cell is the fraction of those resamples in which our form has the lower held-out RMSE: blue above $0.5$, red below. Higher is better.}
\label{tab:paired-diff}
\resizebox{\linewidth}{!}{
\setlength{\aboverulesep}{0pt}\setlength{\belowrulesep}{0pt}%
\begin{tabular}{@{}>{\rowstrut}l|cccc|ccccc@{}}
\toprule
\rowcolor{black!10} \multicolumn{10}{c}{\rowstrut \textbf{high-$C$ holdout}} \\
\midrule
Baseline $\downarrow$ & \makebox[39.08pt]{MNIST} & \makebox[47.04pt]{CIFAR-100} & \makebox[39.08pt]{Darcy} & \makebox[46.17pt]{TinyStories} & \makebox[41.67pt]{Chinchilla} & \makebox[52.52pt]{Muennighoff} & \makebox[39.08pt]{Gadre} & \makebox[39.08pt]{Porian} & \makebox[39.08pt]{Farseer} \\
\midrule
Chinchilla~\citep{hoffmann2022training}      & \cellcolor{blue!16}1.00 & \cellcolor{blue!16}1.00 & \cellcolor{blue!16}1.00 & \cellcolor{red!5}0.35 & \cellcolor{blue!16}1.00 & \cellcolor{blue!14}0.94 & \cellcolor{blue!12}0.86 & \cellcolor{blue!16}0.98 & \multicolumn{1}{>{\columncolor{blue!16}[\tabcolsep][0pt]}c@{}}{1.00} \\
Muennighoff~\citep{muennighoff2023scaling}   & \cellcolor{blue!16}1.00 & \cellcolor{blue!8}0.73 & \cellcolor{blue!8}0.74 & \cellcolor{red!14}0.06 & \cellcolor{blue!16}0.99 & \cellcolor{blue!13}0.92 & \cellcolor{blue!6}0.69 & \cellcolor{blue!4}0.62 & \multicolumn{1}{>{\columncolor{blue!16}[\tabcolsep][0pt]}c@{}}{1.00} \\
M4 ($D$-axis)~\citep{alabdulmohsin2022revisiting} & \cellcolor{blue!16}1.00 & \cellcolor{blue!16}0.99 & \cellcolor{blue!16}1.00 & \cellcolor{blue!16}1.00 & \cellcolor{blue!16}1.00 & \cellcolor{blue!14}0.95 & \cellcolor{blue!15}0.98 & \cellcolor{blue!16}1.00 & \multicolumn{1}{>{\columncolor{blue!16}[\tabcolsep][0pt]}c@{}}{1.00} \\
BNSL $k{=}2$~\citep{caballero2023broken}     & \cellcolor{blue!16}1.00 & \cellcolor{blue!16}0.98 & \cellcolor{blue!16}1.00 & \cellcolor{blue!16}1.00 & \cellcolor{blue!16}1.00 & \cellcolor{blue!15}0.95 & \cellcolor{blue!16}0.98 & \cellcolor{blue!16}1.00 & \multicolumn{1}{>{\columncolor{blue!16}[\tabcolsep][0pt]}c@{}}{1.00} \\
Farseer~\citep{li2025farseer}                & \cellcolor{blue!16}1.00 & \cellcolor{blue!16}1.00 & \cellcolor{blue!16}1.00 & \cellcolor{red!11}0.14 & \cellcolor{blue!15}0.96 & \cellcolor{blue!15}0.97 & \cellcolor{blue!11}0.84 & \cellcolor{blue!11}0.84 & \multicolumn{1}{>{\columncolor{blue!9}[\tabcolsep][0pt]}c@{}}{0.80} \\
\midrule[\heavyrulewidth]
\addlinespace[1em]
\midrule[\heavyrulewidth]
\rowcolor{black!10} \multicolumn{10}{c}{\rowstrut \textbf{high-$D$ holdout}} \\
\midrule
Baseline $\downarrow$ & \makebox[39.08pt]{MNIST} & \makebox[47.04pt]{CIFAR-100} & \makebox[39.08pt]{Darcy} & \makebox[46.17pt]{TinyStories} & \makebox[41.67pt]{Chinchilla} & \makebox[52.52pt]{Muennighoff} & \makebox[39.08pt]{Gadre} & \makebox[39.08pt]{Porian} & \makebox[39.08pt]{Farseer} \\
\midrule
Chinchilla~\citep{hoffmann2022training}      & \cellcolor{red!11}0.16 & \cellcolor{blue!16}1.00 & \cellcolor{blue!16}1.00 & \cellcolor{blue!5}0.65 & \cellcolor{blue!16}0.99 & \cellcolor{blue!16}0.99 & \cellcolor{blue!12}0.86 & \cellcolor{blue!16}1.00 & \multicolumn{1}{>{\columncolor{blue!16}[\tabcolsep][0pt]}c@{}}{1.00} \\
Muennighoff~\citep{muennighoff2023scaling}   & \cellcolor{blue!7}0.73 & \cellcolor{blue!7}0.72 & \cellcolor{blue!9}0.78 & \cellcolor{blue!5}0.64 & \cellcolor{blue!16}0.98 & \cellcolor{blue!13}0.90 & \cellcolor{blue!6}0.69 & \cellcolor{blue!16}1.00 & \multicolumn{1}{>{\columncolor{blue!16}[\tabcolsep][0pt]}c@{}}{1.00} \\
M4 ($D$-axis)~\citep{alabdulmohsin2022revisiting} & \cellcolor{blue!16}1.00 & \cellcolor{blue!16}1.00 & \cellcolor{blue!16}1.00 & \cellcolor{blue!16}1.00 & \cellcolor{blue!16}1.00 & \cellcolor{blue!16}0.98 & \cellcolor{blue!15}0.98 & \cellcolor{blue!16}1.00 & \multicolumn{1}{>{\columncolor{blue!16}[\tabcolsep][0pt]}c@{}}{1.00} \\
BNSL $k{=}2$~\citep{caballero2023broken}     & \cellcolor{blue!16}1.00 & \cellcolor{blue!16}1.00 & \cellcolor{blue!16}1.00 & \cellcolor{blue!16}1.00 & \cellcolor{blue!16}1.00 & \cellcolor{blue!16}0.99 & \cellcolor{blue!16}0.98 & \cellcolor{blue!16}1.00 & \multicolumn{1}{>{\columncolor{blue!16}[\tabcolsep][0pt]}c@{}}{1.00} \\
Farseer~\citep{li2025farseer}                & \cellcolor{blue!16}1.00 & \cellcolor{blue!16}1.00 & \cellcolor{blue!16}1.00 & \cellcolor{blue!5}0.66 & \cellcolor{blue!11}0.85 & \cellcolor{blue!15}0.96 & \cellcolor{blue!11}0.84 & \cellcolor{blue!16}1.00 & \multicolumn{1}{>{\columncolor{blue!7}[\tabcolsep][0pt]}c@{}}{0.72} \\
\bottomrule
\end{tabular}
}
\end{table}

\begin{figure}[t]
  \centering
  \includegraphics[width=\linewidth]{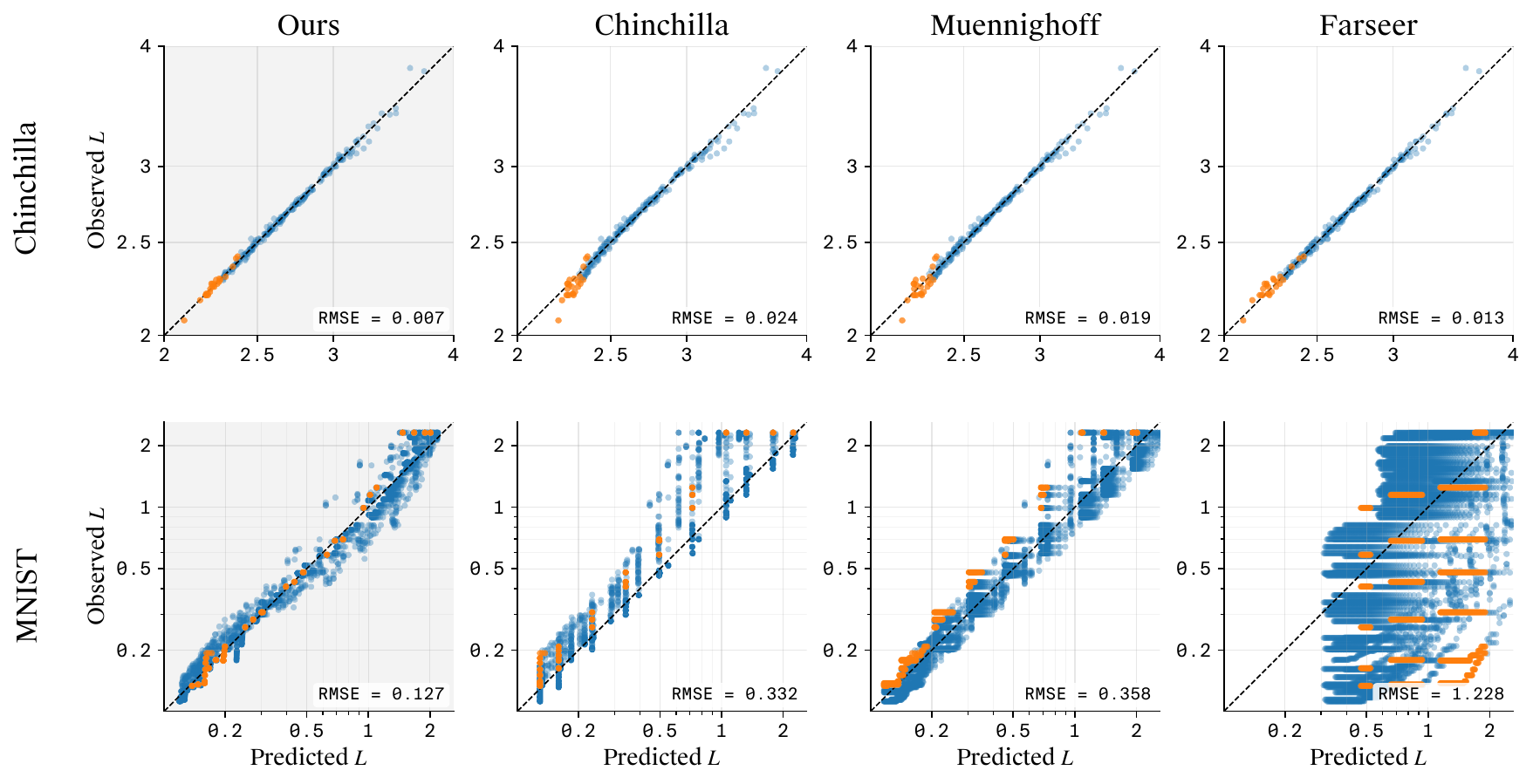}
  \caption{Observed vs. predicted loss for our form (leftmost column) and the three closest competitors, on the data-rich Chinchilla LLM grid (top) and the overfitting-dominated MNIST grid (bottom). Blue: training points; orange: high-$C$ holdout. Our form is tightest in both regimes.}
  \label{fig:obs-vs-pred}
\end{figure}

\paragraph{Summary.} Our form has the lowest median held-out RMSE in 16 of the 18 high-$C$ and high-$D$ cells, ahead of the best competing baseline in those cells by a median factor of $1.5\times$. Ours is also ahead of the best competing baseline in a majority of paired resamples in those same 16 cells, with wins that are all significant on a two-sided sign test at $p < 10^{-3}$. In addition, our form has the best predictive performance on the Farseer far-extrapolation task (detailed in Appendix~\ref{app:far-extrapolation}). The only two cells where our form underperforms a baseline are TinyStories high-$C$ and MNIST high-$D$.

\paragraph{Take-away.} Our results confirm that our form performs well in the multi-epoch data-constrained regimes the form was designed for, when compared to other forms. Furthermore, its benefits are not confined to these regimes: it significantly outperforms all other forms when evaluated on traditional large-scale LLM scaling prediction tasks, indicating that it can be used as a general-purpose scaling law for LLMs. The underperformance on two cells from our experiments motivates the practical need to quantify the loss prediction uncertainty of any fitted form, so we detail a conformal prediction approach in Appendix~\ref{app:conformal-prediction}.

\subsection{Compute footprint}
\label{sec:empirical-cost}

All training ran on AWS Batch with one NVIDIA A10G GPU per job (g5.2xlarge). The campaign consumed $\sim$206 GPU-hours total: MNIST $\sim$46, CIFAR-100 $\sim$14, Darcy $\sim$48, and TinyStories $\sim$98. Aggregate measured training compute is $\sim$$6.0 \times 10^{18}$ FLOPs, dominated by TinyStories ($4.8 \times 10^{18}$).

\section{Cost-aware allocation}
\label{sec:cost}

Beyond fitting held-out loss, a calibrated $L(N, D, T)$ answers two reciprocal questions of training economics. At unit prices $\rho_D$ per unique example and $\rho_C$ per FLOP, with total spend $\mathcal{B}(N, D, T) = \rho_D D + \rho_C k N T$: what is the minimum cost required to reach a target loss $L^*$, and what is the minimum loss achievable within a fixed budget $\mathcal{B}_\text{max}$? Under our form, both $h$ and $\mathcal{B}$ are convex in $(\log N, \log D, \log T)$ (each is a sum of exponentials of affine functions), so both are convex programs with a unique optimum on a single Pareto frontier $\{(\mathcal{B}^*, L^*)\}$. The Lagrange first-order condition for the budget-constrained problem is
\begin{equation}
\frac{\partial L / \partial N}{\rho_C k T} \;=\; \frac{\partial L / \partial D}{\rho_D} \;=\; \frac{\partial L / \partial T}{\rho_C k N},
\label{eq:cost-foc}
\end{equation}
i.e., \emph{at the optimum the marginal loss reduction per dollar is equal across all three axes}. The target-loss problem inverts these ratios and shares the same optimal allocation $(N^*, D^*, T^*)$. Appendix~\ref{app:cost} gives the full derivation and a numerical recipe.

\paragraph{Asymptotic regimes.} The dollar-cost ratio $\eta = \rho_D / \rho_C$ traces a continuum between two limits. When data is free ($\eta \to 0$), the overfitting term vanishes and $(N^*, T^*)$ reduces to Chinchilla compute-optimal, with our $T$-axis playing the role of Chinchilla's $D$-axis. When data is expensive (each example a simulator run, paid label, or licensed token), the overfitting term binds and the optimum walks toward smaller $D$, more epochs, and a smaller $N^*$. Prior cost-aware analyses~\citep{sardana2024beyond, ashton2025fluid, bansal2024smaller} inherit a Chinchilla-style loss with no overfitting term, leaving this data-expensive corner unrepresented.

\paragraph{Illustrative example.} At fixed total budget, we solve for the minimum-loss allocation on Chinchilla's fitted exponents $(\alpha, \beta) = (0.34, 0.28)$~\citep{hoffmann2022training} and illustrative overfitting constants $(c, \gamma, \delta) = (2 \times 10^3, 0.5, 1.0)$ chosen so the overfitting term binds at LLM-scale $(N, D)$ (Table~\ref{tab:cost-allocation}). Increasing $\eta$ by $1000\times$ from web-token economics to expert-label economics shrinks $D^*$ by two orders of magnitude, shrinks $N^*$ by $\sim 20\times$, walks $T^*/D^*$ from $\sim$3 to $\sim 1{,}250$ epochs, and shifts the optimal data-budget share from 12\% to 86\%.

\begin{table}[h]
\centering
\small
\caption{Cost-optimal allocation (illustrative) at fixed budget $\mathcal{B}_\text{max}$ as the data-to-compute price ratio $\eta = \rho_D / \rho_C$ grows.}
\label{tab:cost-allocation}
\begin{tabular}{l rrrrr}
\toprule
$\eta$ & $N^*$ & $D^*$ & $T^*/D^*$ & $L^*$ & data \$ share \\
\midrule
$0$ (Chinchilla CO)         & $5.2 \times 10^9$ & $3.2 \times 10^{11}$ & $1.0$     & $2.14$ & $0\%$ \\
$10^{10}$ (web tokens)      & $4.4 \times 10^9$ & $1.2 \times 10^{11}$ & $2.7$     & $2.14$ & $12\%$ \\
$10^{12}$ (licensed corpora)& $1.2 \times 10^9$ & $6.3 \times 10^{9}$  & $80$      & $2.30$ & $63\%$ \\
$10^{13}$ (expert / sim)    & $2.1 \times 10^8$ & $8.6 \times 10^{8}$  & $1{,}250$ & $2.66$ & $86\%$ \\
\bottomrule
\end{tabular}
\end{table}

\section{Related work}
\label{sec:related}

\paragraph{Parametric scaling laws and Chinchilla extensions.} Power-law generalization curves~\citep{hestness2017deep,rosenfeld2020constructive,henighan2020scaling} were unified into a joint $L(N, D)$ by \citet{kaplan2020scaling}, whose Equation~1.5 implies an overfitting ratio of $N^{0.74}/D$ structurally close to our $N^\gamma/D^\delta$ term, but bundled inside a non-separable power so the $N$- and $D$-axis exponents cannot be fit independently. \citet{hoffmann2022training} simplified this to the additive Equation~\eqref{eq:chinchilla} we recover as a small-$h$ limit. Subsequent work refines the protocol around Chinchilla's form without changing its structural commitments: \citet{gadre2025language} reparameterize via the token multiplier $M = D/N$ for over-trained models, \citet{porian2024resolving} and \citet{pearce2024reconciling} re-examine fits under tighter optimization control, \citet{besiroglu2024chinchilla} flag numerical issues in the original fits, \citet{kumar2025scaling} layer a precision axis, \citet{sardana2024beyond} add an inference-FLOPs term to the cost objective, and \citet{hernandez2021scaling} introduce an effective-data notion under transfer. \citet{li2025farseer} introduce \emph{Farseer}, a 9-parameter $L(N, D)$ form with $N$-dependent exponents reporting large extrapolation-error reductions over Chinchilla. None of these forms includes an explicit baseline $L_0$ or a training-duration axis independent of $(N, D)$.

\paragraph{Saturating and non-monotone forms.} A line of work moves beyond the pure power-law-plus-constant template by introducing saturation. \citet{muennighoff2023scaling} extend Chinchilla to multi-epoch training by replacing $D$ with an effective-data quantity that saturates exponentially in repetitions (and analogously $N$ with a saturating capacity quantity); \citet{liew2025reusing} document related saturation in continual pretraining. This treats $T$ via a fixed exponential curve in the data axis rather than as an independent power-law axis with its own exponent. \citet{alabdulmohsin2022revisiting}'s M4 family instead saturates in the loss axis, bounding $L$ between $E$ and $L_0$ as our wrapper does, but is single-axis (one scalar $x$ rather than a joint $h(N, D, T)$) and stated implicitly in $L$. \citet{caballero2023broken}'s Broken Neural Scaling Laws (BNSL), a $3{+}3k$-parameter smoothly-broken power-law, can express saturation, inflection, and double descent via empirical breakpoints; standard practice fits BNSL with a composite scalar $x(N, D, C)$, yielding a single curve in $x$ rather than an $(N, D, T)$ decomposition, and bounding $L$ in $[E, L_0]$ only approximately. To our knowledge, the $N^\gamma/D^\delta$ term wrapped in $h/(1+h)$ is the first explicit closed-form overfitting term in this literature. Surveys~\citep{li2025misfitting,choshen2025hitchhiker} catalog further variants; Appendix~\ref{app:forms} collects the equation, motivation, and parameter count of every form we compare against.

\paragraph{Theory and other domains.} Theoretical analyses derive scaling exponents from first principles via random-feature, kernel, and dynamical models~\citep{maloney2022solvable,bahri2024explaining,bordelon2024dynamical,brill2024neural}; these target the undercapacity and undertraining exponents and predict no overfitting term of our form. Parametric scaling outside language~\citep{zhai2022scaling,subramanian2023towards,pearce2024scaling,ilse2025radiology} typically reuses or lightly adapts the LLM template and inherits its structural limitations.

\section{Limitations}
\label{sec:limitations}

\paragraph{In-distribution validation loss only.} As in most of the scaling-law literature, our form predicts in-distribution validation loss (loss measured on held-out samples drawn from the same distribution as the training data). Downstream task performance scaling~\citep{gadre2025language}, task-level metrics, emergent capabilities, and out-of-distribution behavior are separate layers in the scaling-law stack and out of scope.

\paragraph{Double descent.} Classical double descent~\citep{belkin2019reconciling,nakkiran2021deep} studies what happens \emph{at} the interpolation threshold $N \approx D$ rather than asymptotically far from it: test error spikes at the threshold and falls again past it, and theoretical results in linear regression and random features~\citep{bartlett2020benign,hastie2022surprises,mei2022generalization} characterize when interpolation is benign. Our form is asymptotic and smooth, so it does not predict a peak at $N \approx D$ and does not attempt to capture this structure. We also do not observe the phenomenon in our experiments, likely because tracking best-so-far validation loss suppresses the model-wise peak~\citep{heckel2021early,nakkiran2021deep} and the peak's typically narrow $N$ range falls between cells on our log-spaced grids.

\paragraph{Universality of fits.} While our form is generally applicable across architectures and domains, the fitted constants $(E, a, b, c, \alpha, \beta, \gamma, \delta)$ are not. Several choices made before fitting are absorbed into these constants and can shift values materially across settings: the parameter-counting convention for $N$, architecture family, and the training recipe (optimizer, schedule, regularization). A scaling law that unifies a fitted form across these choices is beyond the scope of this paper. Practitioners refitting our form should hold these choices fixed across the calibration grid and recalibrate per setting.

\section{Conclusion}

We introduced a closed-form scaling law $L(N, D, T) = E + (L_0 - E)\,h/(1+h)$ that decomposes loss into undercapacity, undertraining, and overfitting terms, bounded between the irreducible loss $E$ and an uninformed baseline $L_0$ fixed by the loss type. The form satisfies six target limiting behaviors, five of which Chinchilla violates, reduces to Chinchilla in the data-rich single-epoch limit, and includes an explicit closed-form overfitting term $c\,N^\gamma/D^\delta$ with no analog in prior parametric scaling laws. Across nine calibration grids spanning four architecture families and three domains, it wins 16 of 18 high-$C$ and high-$D$ extrapolation cells, with state-of-the-art held-out RMSE on every published LLM grid.

A calibrated $L(N, D, T)$ converts a single price ratio $\eta = \rho_D / \rho_C$ into a concrete cost-optimal allocation $(N^*, D^*, T^*)$: the data-free limit recovers Chinchilla compute-optimal, and as data grows expensive (a simulator run, an expert label, a licensed token), the allocation walks toward smaller corpora, more epochs, and smaller models, reaching budget shares prior cost-aware analyses (which inherit Chinchilla's loss form) cannot describe.

We expect the structural ingredients introduced here (baseline saturation, an independent training-duration axis, and an explicit overfitting term) to inform new parametric scaling laws as practitioners move further into the compute-rich, data-constrained regime that motivates this work.

\bibliographystyle{plainnat}
\bibliography{lit_review/references}

\clearpage
\appendix

\addtocontents{toc}{\protect\setcounter{tocdepth}{2}}
\makeatletter
\renewcommand*\l@section[2]{%
  \ifnum \c@tocdepth >\z@
    \addpenalty\@secpenalty
    \addvspace{0.35em \@plus\p@}%
    \setlength\@tempdima{1.5em}%
    \begingroup
      \parindent \z@ \rightskip \@pnumwidth
      \parfillskip -\@pnumwidth
      \leavevmode \bfseries
      \advance\leftskip\@tempdima
      \hskip -\leftskip
      #1\nobreak\hfil \nobreak\hb@xt@\@pnumwidth{\hss #2}\par
    \endgroup
  \fi}
\makeatother
\renewcommand{\contentsname}{Appendix}
\tableofcontents

\clearpage

\section{Our proposed form}
\label{app:proposed-form}

\subsection{The uninformed baseline \texorpdfstring{$L_0$}{L0}}
\label{app:l0}

We derive the value $L_0 = \ln V$ used in Equation~\eqref{eq:ours} as the cross-entropy loss of an \emph{uninformed baseline}: a model that extracts no information from the data and outputs a uniform distribution over the $V$ possible next tokens (analogously, $L_0 = \ln K$ for $K$-way classification).

\paragraph{Setup.} Let the data distribution be $p(x, y)$ over (context, next-token) pairs $(x, y)$ with $y \in \{1, \ldots, V\}$. The cross-entropy loss of a model $q(\cdot \mid x)$ is
\[
L(q) \;=\; \mathbb{E}_{(x, y) \sim p}\bigl[-\log q(y \mid x)\bigr].
\]

\paragraph{Uninformed baseline.} Define the uninformed baseline $q_0$ by $q_0(y \mid x) = 1/V$ for all $y$ and all $x$ (the uniform output is the maximum-entropy distribution given no information about the input). Substituting,
\[
L_0 \;\equiv\; L(q_0) \;=\; \mathbb{E}_{(x, y) \sim p}\bigl[-\log(1/V)\bigr] \;=\; \log V,
\]
where the last equality uses that $-\log(1/V) = \log V$ is a constant and pulls out of the expectation. The result is independent of $p(x, y)$: it does not assume tokens are uniformly distributed in the data, nor any structure on the conditional $p(y \mid x)$.

\paragraph{Classification analog.} For a $K$-way classifier with uniform output $q_0(y \mid x) = 1/K$, the same argument gives $L_0 = \log K$. Next-token prediction over a vocabulary of size $V$ is the special case $K = V$, with $L_0 = \log V$. Per-dataset numerical values are listed in Section~\ref{sec:empirical-setup}.

\paragraph{Relative-$L_2$ analog.} For dense regression on $z$-normalized targets evaluated under relative-$L_2$ loss ($L = \|\hat{u} - u\|_2 / \|u\|_2$, the canonical loss for FNO operator learning~\citep{li2021fourier}), the uninformed baseline is $\hat{u} \equiv 0$ (the conditional-mean prediction under no information about the input, which is zero on $z$-normalized targets), giving $L_0 = \|0 - u\|_2 / \|u\|_2 = 1$ exactly. This is the value used for the Darcy experiment (Appendix~\ref{app:our-experiments-darcy}). The same wrapper $h/(1+h)$ caps the predicted relative-$L_2$ at the uninformed baseline as $h \to \infty$.

\paragraph{Scale equivariance.} Because $L_0$ enters the form as a unit-bearing constant rather than a free parameter, the form is equivariant under uniform rescaling of the loss. If $L' = k L$ for any constant $k > 0$, then $E' = k E$ and $L_0' = k L_0$, and Equation~\eqref{eq:ours} substitutes consistently:
\[
L' \;=\; kL \;=\; k\!\left[E + (L_0 - E)\,\frac{h}{1+h}\right] \;=\; E' + (L_0' - E')\,\frac{h}{1+h}.
\]
Because $h$ is dimensionless, the fitted exponents and coefficients $(a, b, c, \alpha, \beta, \gamma, \delta)$ are invariant under $L \to kL$. Switching cross-entropy units from nats to bits ($k = 1/\ln 2$), or applying any uniform rescaling to the loss, leaves the calibrated form intact. Loss types without a natural uninformed-baseline ceiling (e.g., MSE on unnormalized targets, where the appropriate $L_0$ depends on $\mathrm{Var}(y)$ rather than the loss type alone) need normalization before fitting; the relative-$L_2$ convention above ($L_0 = 1$) is the canonical normalization for dense regression.

\subsection{Limit verification}
\label{app:limits}

Restated for convenience, Equation~\eqref{eq:ours} is
\begin{equation*}
L(N, D, T) = E + (L_0 - E) \cdot \frac{h(N,D,T)}{1 + h(N,D,T)},
\qquad
h = \frac{a}{N^\alpha} + \frac{b}{T^\beta} + c \, \frac{N^\gamma}{D^\delta},
\end{equation*}
with all parameters $(L_0, E, a, b, c, \alpha, \beta, \gamma, \delta)$ non-negative. The wrapper $w(h) = h/(1+h)$ maps $[0, \infty)$ monotonically to $[0, 1)$, so $L$ is automatically bounded in $[E, L_0]$ for any value of the inputs.

\paragraph{Rows 1--5 ($L \to L_0$).} The three terms in $h$ are non-negative, so $h \to \infty$ whenever any one term diverges, and the wrapper saturates: $L \to L_0$. Each of the first five rows of Table~\ref{tab:limits} sends exactly one term to infinity:
\begin{itemize}
\item Row 1 ($N \to 0$; $D, T$ finite): $a/N^\alpha \to \infty$.
\item Row 2 ($D \to 0$; $N, T$ finite): $c\,N^\gamma/D^\delta \to \infty$ for any $N > 0$.
\item Row 3 ($T \to 0$; $N, D$ finite): $b/T^\beta \to \infty$.
\item Row 4 ($N \to \infty$; $D, T$ finite): $c\,N^\gamma/D^\delta \to \infty$ since $D$ is finite.
\item Row 5 ($N, T \to \infty$; $D$ finite): same as row 4 via the overfitting term, irrespective of $T$.
\end{itemize}

\paragraph{Row 6 ($L \to E$).} This row asks for $h \to 0$, which requires \emph{all three} terms in $h$ to vanish simultaneously. The capacity term $a/N^\alpha$ vanishes as $N \to \infty$ and the undertraining term $b/T^\beta$ vanishes as $T \to \infty$, but the overfitting term $c\,N^\gamma/D^\delta$ vanishes only along paths in which $D^\delta$ outpaces $N^\gamma$. The form therefore satisfies row 6 as a \emph{path-dependent} limit: along any joint path with $N, T, D \to \infty$ such that $N^\gamma / D^\delta \to 0$, we have $h \to 0$ and $L \to E$. Paths that fail the third condition (e.g., $D \propto N$ with $\gamma > \delta$) instead drive $L \to L_0$ via the overfitting term. Row 6 thus asserts the existence of a valid path, not pass-anywhere behavior, matching the empirical observation that joint scaling approaches the irreducible loss only when data scales fast enough relative to model size.

\subsection{Chinchilla recovery}
\label{app:recovery}

\begin{proposition}
\label{prop:recovery}
Under three conditions, (i) the single-epoch convention $T = D$, (ii) small $h$ ($h \leq 0.1$), and (iii) negligible overfitting ($c\,N^\gamma/D^\delta \ll a/N^\alpha$), Equation~\eqref{eq:ours} reduces to Equation~\eqref{eq:chinchilla} up to $O(h^2)$ corrections, with the same exponents $(\alpha, \beta)$ and parameter map
\begin{equation}
A = (L_0 - E)\, a, \qquad B = (L_0 - E)\, b.
\label{eq:recovery-map}
\end{equation}
\end{proposition}

\paragraph{Derivation.} The single-epoch convention $T = D$ collapses our $T$ axis onto Chinchilla's $D$ axis: the undertraining term reduces to $b/T^\beta = b/D^\beta$, and the full $h$ becomes
\[
h = \frac{a}{N^\alpha} + \frac{b}{D^\beta} + c \, \frac{N^\gamma}{D^\delta}.
\]
At Chinchilla-optimal allocations $N/D \approx 1/20$. For the overfitting term to be negligible compared to the capacity term, we require
\[
c \, \frac{N^\gamma}{D^\delta} \ll \frac{a}{N^\alpha}, \quad\text{i.e.,}\quad c \ll \frac{a \cdot D^\delta}{N^{\alpha+\gamma}}.
\]
With Chinchilla-scale $N \approx 10^{10}$, $D \approx 2 \times 10^{11}$, $\alpha \approx 0.34$, and illustrative $(\gamma, \delta) = (0.5, 1.0)$, the bound evaluates to $c \ll 800\,a$.

Assuming negligibility and $h \ll 1$, Taylor expansion of $h/(1+h) = h - h^2 + h^3 - \cdots$ gives, to leading order,
\[
L \approx E + (L_0 - E) \Bigl[\frac{a}{N^\alpha} + \frac{b}{D^\beta}\Bigr] = E + \frac{(L_0 - E) a}{N^\alpha} + \frac{(L_0 - E) b}{D^\beta}.
\]
Comparing to Equation~\eqref{eq:chinchilla} gives the parameter map of Equation~\eqref{eq:recovery-map}: $A = (L_0 - E) a$ and $B = (L_0 - E) b$. Because $T$ enters Equation~\eqref{eq:ours} directly (not via $C/(kN)$), the recovery introduces no architecture-dependent $k$ factor. The next-order correction is $-h^2 \cdot (L_0 - E)$, which at $h \sim 0.05$ (our illustrative Chinchilla-scale value) is $\sim 0.02$ nats, smaller than Chinchilla's reported fitting residuals~\citep{hoffmann2022training,porian2024resolving}.

\subsection{Choice of saturating wrapper}
\label{app:wrapper}

The wrapper $h/(1+h)$ in Equation~\eqref{eq:ours} is one of several monotone bijections from $[0, \infty)$ to $[0, 1)$ that satisfy the design constraints in Section~\ref{sec:form}. The most natural alternative is $w(h) = 1 - e^{-h}$, which shares the endpoints $w(0) = 0$ and $w(\infty) = 1$, matches the small-$h$ slope $w'(0) = 1$ that recovers Chinchilla without rescaling, and introduces no new parameters. Other monotone bijections such as $\tanh(h)$ or sigmoidal forms either need a free slope or saturation-rate parameter; we exclude them on parsimony grounds.

The two differ in their approach to saturation. $h/(1+h)$ approaches 1 as $1 - 1/h + O(1/h^2)$, while $1 - e^{-h}$ approaches 1 exponentially. At small $h$, both expand as $h - \kappa h^2 + O(h^3)$ with $\kappa = 1$ for $h/(1+h)$ and $\kappa = 1/2$ for $1 - e^{-h}$, so the two have essentially the same Chinchilla-scale behavior; the difference becomes visible only in the over-trained or overfit corner where $h \gtrsim 1$.

We default to $h/(1+h)$ for closed-form simplicity. Its inverse is rational: writing $L_{\text{rel}} = (L - E)/(L_0 - E)$, the inverse wrapper is $w^{-1}(L_{\text{rel}}) = L_{\text{rel}}/(1 - L_{\text{rel}})$, used in Equation~\eqref{eq:h-target} of Appendix~\ref{app:cost} to map a target loss to the difficulty $h^*$ that achieves it during cost-allocation analysis. The analogous inverse for $1 - e^{-h}$ is the transcendental $-\ln(1 - L_{\text{rel}})$. The Chinchilla-recovery expansion of Proposition~\ref{prop:recovery} likewise has the clean second-order correction $-h^2$, straightforward to bound. Empirically, the two wrappers perform comparably in our evaluations.

\subsection{Remarks on capacity and overfitting}
\label{app:terms}

\paragraph{Population versus sample.} The undercapacity and overfitting terms in $h$ are not mutually exclusive. They separate two levels of the hypothesis-data hierarchy: $a/N^\alpha$ measures how well a hypothesis class of size $N$ can represent the \emph{true population distribution} (a property independent of any particular sample), while $c\,N^\gamma/D^\delta$ measures the variance of the empirical-risk minimizer around the population minimizer when only $D$ \emph{samples} are available, scaled by the model's capacity to express deviations. A small model on small data can suffer both undercapacity (too little capacity to represent the population) and overfitting (still enough capacity to memorize the sparse samples, since memorizing $D$ specific points is an easier hypothesis class than approximating the underlying distribution). In the large-$N/D$ limit the overfitting term grows toward memorization in the strict sense: the model fits the training set exactly while failing to generalize.

\paragraph{Overfitting at a single epoch.} At $T = D$, each example has been seen only once, and in classical statistics, overfitting requires repeated exposure. In modern deep learning, sufficiently large models memorize even single-epoch training data~\citep{carlini2022quantifying,tirumala2022memorization}, and the memorization rate scales with model capacity. A plausible mechanism: at large $N$, a single gradient step has enough degrees of freedom to encode an example into a subset of weights without perturbing the rest of the model, so excess capacity may flow into example-specific encoding rather than population-level features, with validation loss reflecting the resulting gap. On this reading, the $c\,N^\gamma/D^\delta$ term tracks the capacity allocated to training-set-specific encoding. The term itself does not depend on $T$, but its contribution to the predicted loss varies smoothly with $T$ via the saturating wrapper: at small $T$, it is masked by the dominant undertraining term $b/T^\beta$, and at large $T$ (where undertraining decays), it becomes the floor that the loss approaches. The transition is gradual and not sharp at $T = D$.

This same mechanism manifests differently across the training trajectory: \citet{tirumala2022memorization} observe verbatim memorization during the first epoch (well before convergence), while \citet{hernandez2022repeated} and \citet{muennighoff2023scaling} document classical train-val divergence at $T > D$. Both observations are consistent with a single capacity-allocation effect, observed under different criteria. Sub-epoch sample-size effects ($T < D$) are dominated by the undertraining term $b/T^\beta$, since the relevant cost in that regime is the loss decay with each fresh example seen.

\section{Baseline forms}
\label{app:forms}

We detail the prior parametric scaling laws surveyed in Section~\ref{sec:related}. For each form, we list its equation (or structural commitment), the motivation, and the free parameters.

\paragraph{Kaplan Eq.~1.5~\citep{kaplan2020scaling}.} $L(N, D) = [(N_c/N)^{\alpha_N/\alpha_D} + D_c/D]^{\alpha_D}$. The first joint $L(N, D)$ form, derived by combining a power-law fit on each axis. The composite-power structure implies an overfitting ratio of $N^{0.74}/D$ structurally close to our $N^\gamma/D^\delta$ term, but bundled inside a non-separable power, so the $N$- and $D$-axis exponents cannot be fit independently. No explicit $L_0$ ceiling; $L \to \infty$ as $D \to 0$ or $N \to 0$. 4 free parameters: $N_c, D_c, \alpha_N, \alpha_D$.

\paragraph{Chinchilla~\citep{hoffmann2022training}.} $L(N, D) = E + A/N^\alpha + B/D^\beta$ (Equation~\eqref{eq:chinchilla}). Replaces Kaplan's composite power with a strictly additive decomposition and an explicit irreducible-loss floor $E$. The canonical compute-optimal reference form, calibrated in the data-rich single-epoch regime; we recover it as the small-$h$ limit of our form (Appendix~\ref{app:recovery}). 5 free parameters: $E, A, B, \alpha, \beta$.

\paragraph{Muennighoff~\citep{muennighoff2023scaling}.} $L = E + A/(N')^\alpha + B/(D')^\beta$ where $N'$ and $D'$ are exponentially-saturating effective quantities with fitted saturation scales $R_D^*$ and $R_N^*$. The two saturate in different excesses: as repetitions $T/D$ grow, $D' \to D \cdot (1 + R_D^*)$; as parameters exceed $U_N$, the size compute-optimal for $D$ unique tokens, $N' \to U_N \cdot (1 + R_N^*)$. Extends Chinchilla to multi-epoch training by treating each repetition as adding diminishing effective data rather than a fresh example. Multi-epoch behavior in the data axis only; no overfitting term and no independent $T$ axis. 7 free parameters: $E, A, B, \alpha, \beta, R_D^*, R_N^*$.

\paragraph{M4~\citep{alabdulmohsin2022revisiting}.} $L = E + (L_0 - E) / (1 + (x/x_c)^{c})$, a single-axis form bounding $L \in [E, L_0]$, where $x_c$ sets the scale at which the curve turns over and $c$ its curvature in an input scalar $x \in \{N, D, C\}$. The closest structural analog of our wrapper, but single-axis (one scalar $x$ rather than a joint $h(N, D, T)$), so it cannot decompose loss into separate $N$, $D$, $T$ contributions. Our implementation fits 3 free parameters: $E$ and the two coefficients $x_c, c_\text{M4}$.

\paragraph{BNSL~\citep{caballero2023broken}.} A smoothly-broken power law in a composite scalar $x(N, D, C)$ with $k$ empirical breakpoints; each breakpoint adds three free parameters (location, slope change, smoothness). The only baseline able to fit non-monotone $L(N)$, but the breakpoint locations and shapes are fit, not predicted from $(N, D)$ structure; bounds $L$ in $[E, L_0]$ only approximately. We evaluate $k = 1$ and $k = 2$ in our refits. $3 + 3k$ free parameters: 3 base ($a, b, c_0$) + 3 per breakpoint.

\paragraph{Farseer~\citep{li2025farseer}.} $L(N, T) = \exp(a_1 N^{a_2} + a_3) + \exp(b_1 N^{b_2} + b_3) \cdot T^{-\exp(c_1 N^{c_2} + c_3)}$ (Eq.~15 of the Farseer paper; their $D$ axis is total tokens trained on, which corresponds to our $T$ under their compute-relation convention $C \approx 6 N T$). The three exponentials parameterize an $N$-dependent irreducible-loss floor, an $N$-dependent coefficient on the data term, and an $N$-dependent data-axis exponent. Reports state-of-the-art far-compute extrapolation. Single-epoch ($T = D$). 9 free parameters (all signed): $a_1, a_2, a_3, b_1, b_2, b_3, c_1, c_2, c_3$.

\section{Fitting methodology}
\label{app:fitting}

\subsection{Optimization procedure}
\label{app:optimization}

\paragraph{Loss function.} For each form, we minimize Huber loss on log-residuals $r = \log \hat L - \log L$ with transition at $\tau = 0.05$ nats: $\ell(r) = \tfrac{1}{2}r^2$ for $|r| \le \tau$ and $\ell(r) = \tau(|r| - \tau/2)$ otherwise. Huber is robust against the heavy-tailed residuals near saturation cells where small absolute errors translate to large fractional errors in log space, and degenerates to ordinary least-squares in the small-residual limit. On our form, the fit recovers the eight free parameters $(E, a, b, c, \alpha, \beta, \gamma, \delta)$; $L_0$ is fixed by the loss type and not optimized.

\paragraph{Reparameterization.}

Rather than fitting our form directly, we apply two reparameterizations to numerically stabilize the optimization.

\emph{First}, since $N$, $D$, and $T$ are typically large numbers, distributed across many orders of magnitude, their exponentiation tightly couples their coefficient to their exponent. Before fitting, we apply a change of coordinates that ``centers'' each variable by dividing it by the variable's geometric mean within the training set. This change of variables leaves the functional form, the exponents, and $E$ untouched, and rescales the three coefficients $(a, b, c)$ such that:
\begin{equation}
a' = a\,/\,N_0^{\alpha}, \qquad b' = b\,/\,T_0^{\beta}, \qquad c' = c\,N_0^{\gamma}\,/\,D_0^{\delta},
\label{eq:anchoring}
\end{equation}
where $(N_0, T_0, D_0)$ are the geometric means of $N$, $T$, and $D$ within the training set.

\emph{Second}, since $h/(1+h)$ is the logistic function of $\log h$, we never form $h$ itself: we accumulate $\log h$ from its three log-terms by a log-sum-exp (following \citet{hoffmann2022training}), shifted by the largest of them so that no exponential is taken of a positive number, and apply the wrapper to that.

\paragraph{Optimizer and restarts.} The Huber objective is minimized with BFGS on a JAX-jitted gradient in a log-reparameterized parameter space (positive parameters in log; $E$ through softplus to enforce $E \ge 0$). Each \emph{point-estimate} fit uses 30 random restarts to 
mitigate local optima (raised to 200 for Farseer, since its non-convex coupled-exponent landscape is more prone to local minima), and reports the best-of-restarts solution. For most forms (Chinchilla, Muennighoff, BNSL, M4, and ours), each restart independently samples a starting point from a per-parameter distribution: $E \sim \mathcal{U}(0.5, 3.0)$, power-law exponents $\sim \mathcal{U}(0.1, 0.7)$, multiplicative coefficients log-uniform on $[0.01, 1000]$ (with break-scale parameters log-uniform on $[10, 10^6]$ for BNSL). These heuristic ranges are chosen to cover the published point estimates of each form with generous margin. For Farseer, we sample coefficient ranges wide enough to cover the magnitudes its exponent-nested coefficients take ($\pm 100$ on the coefficient inside the data-term exponential). Restarts that fail to converge are discarded.

\paragraph{Bootstrap protocol.}\phantomsection\label{par:bootstrap-protocol} When deriving confidence intervals, we bootstrap by resampling the training set with replacement 200 times, and refit the form on each resample independently from scratch. We report the lower and upper bounds of the 95\% confidence interval using the 2.5\% and 97.5\% quantiles of the bootstrap resamples. To bootstrap efficiently, we use 8 restarts per resample for all forms, and 50 for Farseer (instead of 30 and 200 for the point estimate).

\subsection{Fitted parameters}
\label{app:fitted-params}

Applying the above reparameterization and bootstrap protocol for our form produces the fitted parameters reported in Table~\ref{tab:fitted-params}.

\begin{table}[ht]
\centering
\caption{\textbf{Fitted parameters of \textbf{Ours} (Eq.~\eqref{eq:ours}) on each dataset's full-data fit.} The primed coefficients are in the centered coordinates of Equation~\eqref{eq:anchoring}, so $a'$ is the capacity term's magnitude at $N = N_0$. Values are point estimates; parentheses give their 2.5\% and 97.5\% quantiles. Values are fixed-point to three decimals.}
\label{tab:fitted-params}
\resizebox{\linewidth}{!}{%
\begin{tabular}{@{}>{\rowstrut}l|cccccccc|ccc@{}}
\toprule
Domain $\downarrow$ & $E$ & $a'$ & $\alpha$ & $b'$ & $\beta$ & $c'$ & $\gamma$ & $\delta$ & $N_0$ & $T_0$ & $D_0$ \\
\midrule
MNIST & \msci{0.070}{0.069}{0.072} & \msci{0.000}{0.000}{0.000} & \msci{1.448}{1.383}{1.489} & \msci{0.003}{0.000}{0.004} & \msci{1.226}{1.106}{1.736} & \msci{0.383}{0.380}{0.387} & \msci{0.109}{0.107}{0.111} & \msci{0.584}{0.579}{0.588} & 6.04e5 & 5.98e4 & 5.15e2 \\
CIFAR-100 & \msci{0.563}{0.453}{0.681} & \msci{0.058}{0.033}{0.077} & \msci{0.668}{0.581}{0.861} & \msci{0.024}{0.011}{0.041} & \msci{1.312}{0.902}{1.788} & \msci{0.623}{0.586}{0.666} & \msci{0.107}{0.025}{0.147} & \msci{1.223}{1.099}{1.371} & 9.82e5 & 1.68e6 & 6.31e3 \\
Darcy & \msci{0.005}{0.000}{0.015} & \msci{0.021}{0.012}{0.027} & \msci{0.384}{0.290}{0.665} & \msci{0.029}{0.025}{0.033} & \msci{0.667}{0.564}{0.784} & \msci{0.060}{0.056}{0.062} & \msci{0.000}{0.000}{0.000} & \msci{0.679}{0.661}{0.698} & 5.36e5 & 4.20e4 & 3.10e2 \\
TinyStories & \msci{0.873}{0.744}{0.987} & \msci{0.094}{0.073}{0.116} & \msci{0.483}{0.407}{0.629} & \msci{0.100}{0.088}{0.117} & \msci{0.578}{0.516}{0.703} & \msci{0.112}{0.095}{0.125} & \msci{0.032}{0.007}{0.053} & \msci{0.480}{0.459}{0.505} & 3.23e6 & 5.48e7 & 2.19e6 \\
\midrule
Chinchilla & \msci{1.459}{0.422}{1.879} & \msci{0.050}{0.042}{0.092} & \msci{0.402}{0.319}{0.436} & \msci{0.002}{0.000}{0.142} & \msci{1.295}{0.112}{5.112} & \msci{0.091}{0.000}{0.184} & \msci{0.013}{0.000}{2.498} & \msci{0.182}{0.081}{5.100} & 8.71e8 & 1.58e10 & 1.58e10 \\
Muennighoff & \msci{2.114}{1.898}{2.285} & \msci{0.090}{0.071}{0.132} & \msci{0.391}{0.307}{0.443} & \msci{0.011}{0.006}{0.017} & \msci{0.929}{0.823}{1.094} & \msci{0.032}{0.008}{0.042} & \msci{0.793}{0.613}{1.700} & \msci{0.852}{0.686}{1.610} & 4.50e8 & 1.74e10 & 7.95e8 \\
Gadre & \msci{2.064}{1.038}{2.127} & \msci{0.155}{0.141}{0.212} & \msci{0.334}{0.282}{0.367} & \msci{0.065}{0.000}{0.087} & \msci{0.420}{0.022}{0.497} & \msci{0.000}{0.000}{0.058} & \msci{2.678}{0.080}{6.675} & \msci{2.933}{0.364}{6.702} & 1.14e8 & 6.36e9 & 6.36e9 \\
Porian & \msci{0.000}{0.000}{2.169} & \msci{0.462}{0.152}{0.470} & \msci{0.115}{0.108}{0.326} & \msci{0.252}{0.000}{0.266} & \msci{0.830}{0.131}{0.877} & \msci{0.000}{0.000}{0.266} & \msci{0.000}{0.000}{0.373} & \msci{13.423}{0.000}{14.483} & 8.64e7 & 6.18e8 & 6.18e8 \\
Farseer & \msci{0.103}{0.015}{0.271} & \msci{0.096}{0.055}{0.119} & \msci{0.138}{0.119}{0.204} & \msci{0.012}{0.007}{0.014} & \msci{0.432}{0.384}{0.491} & \msci{0.021}{0.014}{0.024} & \msci{0.195}{0.126}{0.224} & \msci{0.164}{0.147}{0.258} & 7.18e8 & 2.02e10 & 2.02e10 \\
\bottomrule
\end{tabular}
}
\end{table}

\subsection{Parameter identification}
\label{app:identification}

Each of the parameters in our form is identified by a different empirical regime, as summarized in Table~\ref{tab:identification}.

\begin{table}[h]
\centering
\caption{Empirical regime that identifies each parameter of Equation~\eqref{eq:ours} within a single fit.}
\label{tab:identification}
\begin{tabular}{@{}ll@{}}
\toprule
Parameter & Identifying regime within a fit \\
\midrule
$E$ & resource-rich asymptote: large $N$, large $D$, large $T$ jointly \\
$a, \alpha$ & $N$-sweep at large $D, T$ (Chinchilla envelope) \\
$b, \beta$ & $T$-sweep at large $N, D$ (capacity and overfitting terms suppressed) \\
$c, \gamma, \delta$ & small-$D$ runs at varied $N$ together with multi-epoch runs at varied $T/D$ \\
$L_0$ & not fitted: vocabulary size or loss-type ceiling \\
\bottomrule
\end{tabular}
\end{table}

\paragraph{Regime decomposition.} Each of the three terms in our form has a regime where the other two are suppressed, so a value sweep of the dominant term isolates the parameter of interest.

The \emph{capacity term} $a/N^\alpha$ is the only term that does not vanish as $D, T \to \infty$, so the Chinchilla envelope (large $D, T$) isolates $a, \alpha$ by sweeping $N$.

The \emph{undertraining term} $b/T^\beta$ is the only term that depends on $T$ at all, so the well-resourced corner (large $N$, which kills the capacity term, and $D \gg N^{\gamma/\delta}$, which kills the overfitting term) isolates $b, \beta$ by sweeping $T$.

The \emph{overfitting term} $c\,N^\gamma/D^\delta$ is the only term that grows as $D$ shrinks, so the data-constrained corner (large $T$, which kills the undertraining term, and large $N$, which kills the capacity term and boosts the overfitting term) pushes it into dominance at small $D$. Isolating it takes two sweeps rather than one: sweeping $D$ at fixed $N$ pins $\delta$, and sweeping $N$ at fixed $D$ pins $\gamma$. Either sweep alone leaves $c$ entangled with the exponent it did not vary, fixing only the prefactor $c\,N^\gamma$ or $c/D^\delta$.

Practically, the \emph{irreducible loss floor} $E$ is the most challenging parameter to identify, since it requires that $N$, $D$, and $T$ are all large simultaneously (the most expensive regime to sample).

The wrapper is strictly monotonic, so $h$ can be recovered from an observed loss as $h = (L - E)/(L_0 - L)$. That recovery is noise-sensitive at both ends: near the ceiling $L_0$ and near the floor $E$, a wide range of $h$ produces losses differing by less than measurement noise. Cells at either extreme therefore constrain $h$ weakly however densely they are sampled, which bootstrap CIs diagnose as wide marginal intervals.

\paragraph{Chinchilla-grid sub-regime.} For a fit restricted to a Chinchilla-style grid ($T \approx D$), the recovery map of Equation~\eqref{eq:recovery-map} identifies $a, b, \alpha, \beta$ within that fit, while $c, \gamma, \delta$ are resolved only weakly: at $T = D$, the three terms of $h$ occupy two axes rather than three, so the overfitting term trades against the other two instead of being pinned on its own. When fitting our form to the Chinchilla dataset, the overfitting term survives as a second power law on the data axis with an exponent of its own; it supplies $64\%$ of $h$ at the grid's geometric center, at $\delta = 0.182$ against $\beta = 1.295$.

Identifying $c, \gamma, \delta$ individually requires a grid that additionally exercises the overfitting regime: small-$D$ runs at varied $N$, multi-epoch runs at varied $T/D$, or both. When the fitted grid lacks these regimes, the marginal CI on $(c, \gamma, \delta)$ is wide.

\FloatBarrier
\section{Our training experiments}
\label{app:our-experiments}

This appendix documents the four experimental domains we ran ourselves: MNIST, CIFAR-100, Darcy, and TinyStories.

\paragraph{Conventions shared across domains.} For each run, we measure total training FLOPs $C$ directly with a FLOPs counter; the form takes $T$ rather than $C$ and does not depend on the linear approximation $C \approx kNT$. The four domains differ in how loss measurements are collected. \emph{MNIST} uses constant LR throughout with no cooldown, so every logged checkpoint is a valid end-state observation: at each checkpoint we record $(N, D, T_\text{checkpoint}, L_\text{val}^*)$ with $L_\text{val}^*$ the running minimum validation loss, giving one \emph{trajectory} of points per $(N, D)$ cell. \emph{CIFAR-100, Darcy, and TinyStories} use a Warmup-Stable-Decay schedule~\citep{hagele2024scaling}: each $(N, D)$ cell consists of one constant-LR stable run that feeds multiple cooldown forks at log-spaced $T_\text{total}$ targets, and we record $(N, D, T_\text{total}, L_\text{final})$ once per cooldown. \citet{hagele2024scaling} show that constant LR plus a short late cooldown matches cosine performance and lets a single stable run feed multiple cooldown forks at different total-compute budgets. Stable-phase intermediate checkpoints from these three domains are not used as data points, because the cooldown's downstream effect makes the intermediate $L_\text{val}^*$ inequivalent to a final-loss observation at the same $T$; MNIST's no-cooldown protocol makes intermediate-checkpoint and end-of-training observations equivalent by construction (so unlike in other grids, all our MNIST trajectory points are valid). We run a single seed per $(N, D)$ stable run for CIFAR-100, Darcy, and TinyStories, and 10 seeds per cell for MNIST (per-cell losses averaged before fitting), following standard practice in the scaling-law literature~\citep{kaplan2020scaling,hoffmann2022training,muennighoff2023scaling,hagele2024scaling}; the bootstrap procedure in Appendix~\ref{app:optimization} resamples $(N, D, T; L)$ training-set points to produce confidence intervals on the fitted parameters.

\paragraph{Effective unique-data exposure.} For trace points or cooldown forks where $T < D_\text{budget}$, the model has only been exposed to at most $T$ unique training points regardless of how many were prepared in the corpus. We cap $D = \min(D_\text{budget}, T)$ before fitting so that the form's overfitting term $c\,N^\gamma/D^\delta$ is fed the actually-seen sample count, not the budgeted one; without this cap, the form would treat sub-1-epoch rows as if the model had full access to the prepared corpus, mis-attributing the high observed loss to undertraining alone instead of also to limited data exposure. The cap is engaged on $\sim$6\% of MNIST trace rows (earliest checkpoints on large-$D$ cells, where $T_\text{checkpoint}$ has not yet reached $D_\text{budget}$) and $\sim$17\% of TinyStories cooldown forks (smallest-$T_\text{total}$ forks on large-$D$ cells); CIFAR-100 and Darcy have no sub-1-epoch rows by construction, so the cap has no effect there.

\subsection{MNIST}
\label{app:our-experiments-mnist}

\paragraph{Description.} MNIST 10-way digit classification~\citep{lecun1998gradient}. The grid covers all four probing regimes (undercapacity, compute-optimal, overfitting, and random initialization); with $L_0 = \ln 10 \approx 2.30$ nats setting an unambiguous saturation ceiling and a cheap constant-LR sweep that produces a continuous $L(N, D, T)$ trajectory, every limiting behavior of the form is accessible at modest compute.

\paragraph{Architecture and grid.} 2-hidden-layer ReLU MLP, widths $w \in \{16,\allowbreak 32,\allowbreak 64,\allowbreak 128,\allowbreak 256,\allowbreak 512,\allowbreak 1024,\allowbreak 2048,\allowbreak 4096,\allowbreak 8192\}$, giving $N \in \{13\text{K},\allowbreak 26\text{K},\allowbreak 55\text{K},\allowbreak 118\text{K},\allowbreak 269\text{K},\allowbreak 669\text{K},\allowbreak 1.9\text{M},\allowbreak 5.8\text{M},\allowbreak 20\text{M},\allowbreak 74\text{M}\}$ params (since $N = 784w + w^2 + 10w$). Unique-data sweep $D \in \{10,\allowbreak 20,\allowbreak 50,\allowbreak 100,\allowbreak 320,\allowbreak 1000,\allowbreak 3160,\allowbreak 10000,\allowbreak 31600,\allowbreak 60000\}$ training images (10 log-spaced values from 10 to 60\,000). Total-exposure sweep $T \in [64,\, 10^6]$ images seen, from one batch up to a per-cell $T_\text{max}$ of $10^5$ for $D \le 100$ and $10^6$ for $D > 100$. That spans from $\sim$17 epochs at the largest cell ($D = 60{,}000$) to $\sim$10{,}000 at the smallest ($D = 10$). $5\%$ of rows are sub-epoch in the prepared corpus, down to $T/D_\text{budget} \approx 10^{-3}$; under the cap above they enter the fit at $D = T$, so the fitted $T/D$ bottoms out at exactly 1. Validation loss measured on the standard 10{,}000-image test set.

\paragraph{Training recipe.} Adam ($\beta_1 = 0.9, \beta_2 = 0.999$) with constant LR $3 \times 10^{-3}$ from step 0, no warmup, no weight decay, batch size 64. The trace-point convention samples $\sim$100 log-uniform checkpoints between step 0 and $T_\text{max}/\text{batch}$ per cell. No cooldown forks: under the constant-LR protocol every logged checkpoint is a valid end-state observation, and the per-$(N, D)$ trajectory is collected directly. We run 10 independent seeds per $(N, D)$ cell and average per-cell losses before fitting.

\subsection{CIFAR-100}
\label{app:our-experiments-cifar}

\paragraph{Description.} CIFAR-100 100-way image classification~\citep{krizhevsky2009learning}. The grid tests whether the form's exponents and saturating wrapper transfer to vision with a different architecture and a different $L_0$ ($\ln 100 \approx 4.61$ nats).

\paragraph{Architecture and grid.} PreActResNet~\citep{he2016identity}, 5 configurations $(n, w) \in \{(1,16),\allowbreak (2,24),\allowbreak (3,32),\allowbreak (4,48),\allowbreak (5,64)\}$ where $n$ is PreActBlocks per stage (depths $6n+2 \in \{8, 14, 20, 26, 32\}$) and $w$ the base channel width, giving $N \in \{84\text{K},\allowbreak 400\text{K},\allowbreak 1.1\text{M},\allowbreak 3.3\text{M},\allowbreak 7.5\text{M}\}$ params. Unique-data sweep $D \in \{500, 2000, 8000, 25000, 50000\}$ training images. Total-exposure sweep $T \in [5.0 \times 10^5,\, 5.0 \times 10^6]$ images seen ($T_\text{max} = 5{\times}10^6$ examples, smallest fork at $T_\text{max}/10$), spanning $T/D$ from 10 epochs everywhere to $10^4$ at the largest fork on the smallest $D$. Validation loss measured on the standard 10{,}000-image test set.

\paragraph{Training recipe.} SGD with momentum 0.9, weight decay $5 \times 10^{-4}$, batch size 128. Peak LR scales with width as $0.1 \cdot (32/w)^{1/2}$ (anchor-tuned at $w = 32$), held constant after $W = 200$ linear-warmup steps. SGD+momentum is the canonical CIFAR ResNet optimizer~\citep{he2016deep}; we use constant LR for the stable phase rather than the conventional cosine decay so that the same stable run can feed multiple cooldown forks. We run 5 cooldown forks per stable run, log-spaced over $[T_\text{max}/10, T_\text{max}]$, with square-root decay over the final 20\% of fork compute. We apply standard CIFAR augmentation (random crop $32 \times 32$ with 4-pixel padding, random horizontal flip).

\subsection{Darcy}
\label{app:our-experiments-darcy}

\paragraph{Description.} PDEBench 2D Darcy flow at $\beta = 1.0$~\citep{takamoto2022pdebench}, the canonical operator-learning benchmark for parametric elliptic PDEs. The task is to map a scalar permeability field $a(x, y)$ on a $128 \times 128$ grid to the steady-state pressure field $u(x, y)$ that satisfies $-\nabla \cdot (a(x, y) \nabla u(x, y)) = f(x, y)$ with prescribed boundary conditions. Loss is the relative-$L_2$ residual ($L = \|\hat{u} - u\|_2 / \|u\|_2$, the canonical Li 2021 \texttt{LpLoss}\,$(p=2)$), which gives $L_0 = 1$ exactly under zero predictions on $z$-normalized targets.

\paragraph{Architecture and grid.} Fourier neural operator (FNO; \citealp{li2021fourier}) with fixed depth ($n_\text{layer} = 4$) and Fourier modes ($n_\text{modes} = 12$), sweeping the hidden channel width $w \in \{16, 24, 32, 48, 64, 96\}$ for 6 model configurations spanning $\sim\!90$K to $\sim\!3.2$M params. Unique-data sweep $D \in \{10, 30, 100, 320, 1000, 3160, 9000\}$ $(a, u)$ pairs. Total-exposure sweep $T \in [3.0 \times 10^3,\, 2.0 \times 10^6]$ pairs seen, with $T_\text{max}$ scaling with $D$ ($5{\times}10^4$ at $D = 30$, $2{\times}10^5$ at $D = 100$, $5{\times}10^5$ at $D = 320$, $10^6$ at $D \in \{1000, 3160\}$, $2{\times}10^6$ at $D = 9000$), spanning $T/D$ from 2.2 epochs to $\sim$2{,}000 at the largest fork. Validation loss measured on a fixed 1{,}000-pair holdout.

\paragraph{Training recipe.} AdamW with $\beta = (0.9, 0.999)$ and weight decay $10^{-4}$ (FNO defaults), peak LR $10^{-3}$ held flat after $W = 100$ linear-warmup steps, batch size 20 $(a, u)$ pairs per step (Li 2021 default), and 10 cooldown forks per stable run (square-root decay over the final 20\% of fork compute, log-spaced over $[\max(T_\text{fork}, T_\text{max}/100), T_\text{max}]$).

\subsection{TinyStories}
\label{app:our-experiments-tinystories}

\paragraph{Description.} TinyStories~\citep{eldan2023tinystories} next-token language modeling with a 2K BPE tokenizer trained on the TinyStories corpus (not Eldan \& Li's original 10K GPT-Neo vocabulary; $L_0 = \ln 2000 \approx 7.60$ nats). TinyStories is the natural setting for fitting the multi-epoch behavior of the form: every cell on the grid admits $T \gg D$, so the $b/T^\beta$ undertraining term and the $c\,N^\gamma/D^\delta$ overfitting term are jointly identifiable.

\paragraph{Architecture and grid.} GPT-style decoder-only transformer in the spirit of \citet{eldan2023tinystories}. 5 model sizes with $(d_\text{model}, n_\text{layer}, n_\text{head}, d_\text{ff}) \in \{(64, 4, 4, 256),\allowbreak (128, 4, 4, 512),\allowbreak (192, 6, 6, 768),\allowbreak (320, 8, 8, 1280),\allowbreak (480, 10, 8, 1920)\}$, giving $N \in \{344\text{K},\allowbreak 1.1\text{M},\allowbreak 3.1\text{M},\allowbreak 10.6\text{M},\allowbreak 28.8\text{M}\}$ params. Context length 256 tokens, fixed across runs. Unique-data sweep $D \in \{10^4, 10^5, 10^6, 10^7, 10^8, 5{\times}10^8\}$ unique tokens, giving 30 stable $(N, D)$ runs and 300 cooldown-fork final-loss measurements. Total-exposure sweep $T \in [2.0 \times 10^6,\, 2.5 \times 10^9]$ tokens seen, with $T_\text{max}$ scaling per cell from $5 \times 10^7$ tokens at the smallest $D$ to $2.5 \times 10^9$ at the largest, spanning $T/D$ up to $\sim$5{,}000 epochs at the largest fork. $17\%$ of rows are sub-epoch in the prepared corpus, reaching $T/D_\text{budget} = 0.05$ at the smallest forks of the large-$D$ cells. Those rows enter the fit at $D = T$ under the cap above. Validation loss measured on a fixed 2{,}000-document holdout from the TinyStories validation split.

\paragraph{Training recipe.} AdamW with $\beta = (0.9, 0.95)$ and weight decay $0.1$ (GPT-style defaults), peak LR $3 \times 10^{-4}$ held flat after $W = 500$ linear-warmup steps, batch size 64 sequences of 256-token context $= 16{,}384$ tokens per step (fixed across configs), and 10 cooldown forks per stable run (square-root decay over the final 20\% of fork compute, log-spaced over $[\max(T_\text{fork}, T_\text{max}/100), T_\text{max}]$). \emph{Note:} a config drift on the later 30M-size submission set $W = 100$ for those runs instead of 500; recorded loss is the post-cooldown final-loss, dominated by the late-training trajectory rather than the warmup duration, so $L$ at any shared $T_\text{total}$ is unaffected.

\section{External refits}
\label{app:external-refits}

This appendix documents the five published LLM scaling-law grids we refit our form to: Chinchilla, Muennighoff, Gadre, Porian, and Farseer. In each case, the grid covers only a subset of the $(N, D, T)$ space that our form parameterizes.

\subsection{Chinchilla}
\label{app:external-refits-chinchilla}

\paragraph{Description.} \citet{hoffmann2022training}, transcribed by~\citet{besiroglu2024chinchilla} from DeepMind's published Approach-3 isoFLOP figures (raw isoFLOP sweep was never released). The canonical compute-optimal reference grid for the Chinchilla parametric form. Single-epoch training ($T = D$ throughout).

\paragraph{Architecture and grid.} Decoder-only transformers; tokenizer is a 32K SentencePiece vocabulary on MassiveText ($L_0 = 10.373$). 245 rows spanning $N \in [5.73 \times 10^7, 1.62 \times 10^{10}]$ params and $D \in [2.45 \times 10^8, 3.18 \times 10^{11}]$ tokens at Chinchilla-optimal and near-optimal compute.

\paragraph{Training recipe.} As reported by~\citet{hoffmann2022training}: cosine LR schedule with warmup, AdamW, single epoch over MassiveText. Loss is cross-entropy nats per token.

\subsection{Muennighoff}
\label{app:external-refits-muennighoff}

\paragraph{Description.} \citet{muennighoff2023scaling}, source data is the \texttt{NAMES\_TO\_VAL\_LOSSES} dict in HuggingFace's \texttt{datablations} repository (\texttt{lm1-misc/parametric\_fit.ipynb}). The only public grid with $T > D$ (multi-epoch repetition).

\paragraph{Architecture and grid.} Decoder-only transformers on C4; 50,257-token GPT-NeoX/GPT-2 vocabulary ($L_0 = 10.825$). The source data holds 324 rows; averaging the replicate runs that share an $(N, D, T)$ cell leaves 232 rows, spanning $N \in [7.10 \times 10^6, 8.67 \times 10^9]$ params and $D \in [1.00 \times 10^8, 1.78 \times 10^{11}]$ tokens. Most cells are single runs (214 of the 232) and the 18 replicated cells carry 2 to 17 rows each. Replicates agree closely in most of those cells (median spread $0.027$ nats), but not in all: three spread by $0.17$ to $0.36$ nats, which is the same size as the held-out RMSE any form attains on this grid, so the averages in those three cells should not be read as precise measurements. The grid runs $T/D > 1$ in 198 of its 232 rows, at 23 distinct values of $D$ from $1.00 \times 10^8$ to $8.80 \times 10^{10}$, reaching $T/D = 9000$ at the most repeated cells.

\paragraph{Training recipe.} As reported by~\citet{muennighoff2023scaling}: cosine LR schedule, AdamW, multi-epoch passes over data-constrained subsets of C4. Loss is cross-entropy nats per token.

\subsection{Gadre}
\label{app:external-refits-gadre}

\paragraph{Description.} \citet{gadre2025language}, published CSV at \texttt{mlfoundations/scaling}. The over-trained-regime grid: single-epoch models sweeping the token multiplier $M = D/N$ from $\sim\!5$, below Chinchilla-optimal, up to $\sim\!640$. The published CSV mixes three corpora at the same $(N, D)$; we filter to \texttt{rw\_original} (RefinedWeb) to match Porian's corpus and enable direct over-trained-vs-Kaplan comparison. Single epoch ($T = D$).

\paragraph{Architecture and grid.} Decoder-only transformers on RefinedWeb; 50,432-token vocabulary ($L_0 = 10.828$; recoverable from their released \texttt{params} and \texttt{params\_no\_embed} fields, whose difference divided by the model width is exactly 50{,}432 at every width). The published CSV holds 104 rows across three corpora (35 \texttt{rw\_original}, 35 \texttt{rpj}, 34 \texttt{c4\_original}); the corpus filter leaves only 35 rows, spanning $N \in [1.06 \times 10^7, 6.89 \times 10^9]$ params and $D \in [5.28 \times 10^7, 4.61 \times 10^{11}]$ tokens. This is the smallest of our nine grids by a wide margin, and the high-$C$ and high-$D$ protocols hold out 4 cells each, so its held-out RMSE carries correspondingly more sampling noise than the other columns, and we do not read a narrow margin on this grid as meaningful.

\paragraph{Training recipe.} As reported by~\citet{gadre2025language}: cosine LR schedule with linear warmup, AdamW, single-epoch over RefinedWeb. Loss is cross-entropy nats per token.

\subsection{Porian}
\label{app:external-refits-porian}

\paragraph{Description.} \citet{porian2024resolving}, used as a Kaplan stand-in (Kaplan's 2020 raw data is not public). Porian re-runs Chinchilla-style training under several LR schedules including a Kaplan-style decay-to-fixed-$T_\text{max}$.

\paragraph{Architecture and grid.} Decoder-only transformers on RefinedWeb; 32K SentencePiece tokenizer ($L_0 = 10.373$). 238 rows after filtering to the \texttt{rw|base|long\_warmup|kaplan\_decay} subset (the closest reproduction of Kaplan-style training; other protocols dropped to avoid mixing schedules in one fit), spanning $N \in [3.32 \times 10^5, 9.02 \times 10^8]$ params and $D \in [5.05 \times 10^6, 2.86 \times 10^{10}]$ tokens, with a denser experimental grid near the compute-optimal frontier. Single epoch ($T = D$).

\paragraph{Training recipe.} As reported by~\citet{porian2024resolving}: long-warmup cosine schedule with optional Kaplan-style decay-to-fixed-$T_\text{max}$, AdamW with carefully tuned batch size and LR, single epoch over RefinedWeb. Loss is cross-entropy nats per token.

\subsection{Farseer}
\label{app:external-refits-farseer}

\paragraph{Description.} \citet{li2025farseer}: the largest single-recipe scaling-law dataset publicly available (404 dense-LM runs, single-epoch, cosine-to-zero LR schedule). We use the authors' \texttt{1222\_full.csv} release, scored on their held-out \texttt{IntelliValSet\_Raw} bits-per-character measure. Four $(N, D, T)$ cells hold two rows each, and we average them, resulting in a 400-cell cohort.

\paragraph{Architecture and grid.} Dense decoder-only transformers; 65,536-token tokenizer on a web + math + code mixture. We fit the bits-per-character loss their paper reports rather than cross-entropy nats, which sets the saturating ceiling at $L_0 = \log_2(65{,}536)/4 = 4.0$ bits/char, assuming a mixture averaging 4 characters per token. 400 cells, spanning $N \in [9.96 \times 10^7, 6.37 \times 10^9]$ params and $D \in [1.00 \times 10^9, 5.12 \times 10^{11}]$ tokens. Single-epoch ($T = D$).

\paragraph{Training recipe.} As reported by~\citet{li2025farseer}: cosine-to-zero LR schedule with warmup, AdamW, single-epoch over the web/math/code mixture. Loss is bits per character.

\section{Cost-aware allocation}
\label{app:cost-aware}

\subsection{Compute-optimal model size at fixed unique data}
\label{app:nopt}

We derive the optimal model size $N^*$ at fixed unique data $D$ under our form, in two regimes: the asymptotic limit $T \to \infty$, and the finite-compute case using the planning approximation $C \approx kNT$. We then verify that no other parametric scaling law referenced in this paper admits an analogous interior minimum in $N$ at fixed $D$.

\paragraph{Asymptotic optimum ($T \to \infty$).} At fixed $D$ as $T \to \infty$, the undertraining term $b/T^\beta$ vanishes and
\[
h_\infty(N, D) = \frac{a}{N^\alpha} + \frac{c\,N^\gamma}{D^\delta}.
\]
The stationarity condition $\partial h_\infty / \partial N = 0$ gives
\begin{equation}
-\frac{\alpha a}{N^{\alpha+1}} + \frac{\gamma c\,N^{\gamma-1}}{D^\delta} = 0
\quad\Longrightarrow\quad
N^*_\infty(D) = \left(\frac{\alpha a\, D^\delta}{\gamma c}\right)^{1/(\alpha+\gamma)}.
\label{eq:nopt-asymptotic}
\end{equation}
This is an interior minimum:
$$\frac{\partial^2 h_\infty}{\partial N^2} = \alpha(\alpha+1) \frac{a}{N^{\alpha+2}} + \gamma(\gamma-1) \frac{c\,N^{\gamma-2}}{D^\delta},$$
which is positive at $N^*_\infty$ for $\alpha, \gamma > 0$ and $a, c > 0$: substituting the stationarity condition gives $\partial^2 h_\infty / \partial N^2 = \alpha(\alpha + \gamma)\,a / N^{\alpha+2} > 0$ at $N = N^*_\infty$. Below $N^*_\infty$, the undercapacity term dominates and increasing $N$ helps; above $N^*_\infty$, the overfitting term dominates and increasing $N$ raises the asymptotic loss. Because the wrapper $h/(1+h)$ is monotone, the same $N^*_\infty$ minimizes $L$ itself.

\paragraph{Finite-compute optimum.} Using $T = C/(kN)$, at fixed $C, D$ we minimize
\[
h(N, D, C) = \frac{a}{N^\alpha} + \frac{b\,k^\beta\,N^\beta}{C^\beta} + \frac{c\,N^\gamma}{D^\delta}.
\]
Setting $\partial h / \partial N = 0$ gives the implicit relation
\begin{equation}
\alpha a \;=\; \frac{\beta\,b\,k^\beta\,N^{*\,\alpha+\beta}}{C^\beta} \;+\; \frac{\gamma c\,N^{*\,\alpha+\gamma}}{D^\delta},
\label{eq:nopt-finite}
\end{equation}
which has a unique solution since the right-hand side is monotone increasing in $N$ from $0$ at $N = 0$ to $\infty$. Two limits recover the boundary behavior:

\emph{Limit $D \to \infty$ (data-rich, Chinchilla regime).} The third term in Equation~\eqref{eq:nopt-finite} vanishes, giving
\begin{equation}
N^*(C) = \left(\frac{\alpha a\,C^\beta}{\beta\,b\,k^\beta}\right)^{1/(\alpha+\beta)} \;\propto\; C^{\beta/(\alpha+\beta)}.
\label{eq:nopt-chinchilla}
\end{equation}
With Chinchilla-like exponents $\alpha \approx \beta$, $N^*(C) \propto C^{1/2}$, recovering~\citet{hoffmann2022training}'s compute-optimal scaling.

\emph{Limit $C \to \infty$ at fixed $D$.} The second term in Equation~\eqref{eq:nopt-finite} vanishes, recovering $N^*_\infty(D)$ from Equation~\eqref{eq:nopt-asymptotic}. The compute-optimal model size therefore saturates at a finite value as compute grows past the point at which adding capacity costs more in overfitting than it gains in undercapacity reduction. This is the data-constrained shift qualitatively documented by~\citet{muennighoff2023scaling}: spending more compute on a fixed unique-data pool eventually requires shrinking the model rather than enlarging it.

\paragraph{Other parametric forms.} Among the parametric scaling laws of Section~\ref{sec:related}, none admits a non-trivial $N^*$ at fixed $D$ as $T \to \infty$:
\begin{itemize}
\itemsep -0.05em
\item \emph{Chinchilla.} $L = E + A/N^\alpha + B/D^\beta$ is monotone non-increasing in $N$.
\item \emph{Kaplan Eq.~1.5.} $L = [(N_c/N)^{\alpha_N/\alpha_D} + D_c/D]^{\alpha_D}$. As $N \to \infty$ the first bracket term vanishes and $L \to (D_c/D)^{\alpha_D}$; monotone non-increasing in $N$.
\item \emph{Muennighoff.} $L = E + A/(N')^\alpha + B/(D')^\beta$ with effective $N'$ saturating at $N(1 + R_N^*)$ as $N \to \infty$. Both terms are monotone non-increasing in $N$.
\item \emph{M4.} Single-axis form, monotone by construction.
\item \emph{Farseer.} $N$-dependent exponents but no overfitting term; inherits Chinchilla's monotonicity in $N$.
\item \emph{BNSL.} Can express non-monotone behavior via empirical breakpoints, but the breakpoint locations and shapes are fit, not predicted from $(N, D)$ structure.
\end{itemize}
The U-shape in $L(N)$ at fixed $D$ is therefore a structural feature unique to our form among the reference set, and matches the empirical observation of~\citet{hernandez2022repeated} that validation loss is non-monotone in capacity under heavy data repetition.

\subsection{Cost-optimal allocation}
\label{app:cost}

A calibrated scaling law lets us answer two reciprocal practitioner questions: \emph{(P1)} given a target validation loss $L^*$, what is the minimum dollar cost $\mathcal{B}^*$ to reach it, and how should that budget split across model size $N$, unique data $D$, and training duration $T$? \emph{(P2)} given a fixed dollar budget $\mathcal{B}_\text{max}$, what is the lowest validation loss achievable, and what allocation achieves it? When training compute is the sole cost (the Chinchilla setup), (P1) and (P2) collapse to dual readouts of the compute-optimal frontier. When unique data must be generated, labeled, or licensed at non-trivial per-example cost, both require joint optimization over $(N, D, T)$ with two independent cost terms.

\paragraph{Setup.} Let $\rho_D$ be the dollar cost per unique training example (label, simulation run, scraped token, etc.) and $\rho_C$ the dollar cost per FLOP. Using the standard approximation $C \approx kNT$, the dollar cost is
\begin{equation}
\mathcal{B}(N, D, T) = \rho_D \cdot D + \rho_C \cdot k N T.
\label{eq:cost}
\end{equation}
Both (P1) and (P2) are smooth constrained optimizations on $\mathbb{R}^3_{>0}$ under our form $L = E + (L_0 - E)\,h/(1+h)$ with $h = a/N^\alpha + b/T^\beta + c\,N^\gamma/D^\delta$. They are duals (the same Lagrangian governs both), so the optimal allocation $(N^*, D^*, T^*)$ that solves (P1) at target $L^*$ also solves (P2) at budget $\mathcal{B}^*(L^*)$.

\paragraph{(P1) Target-loss optimum.} Form the Lagrangian $\mathcal{L}_1 = \mathcal{B} + \mu (L - L^*)$ with $\mu \geq 0$. Setting $\partial \mathcal{L}_1 / \partial X = 0$ for $X \in \{N, D, T\}$ gives the stationarity conditions
\begin{equation}
-\frac{\rho_C k T}{\partial L / \partial N} \;=\; -\frac{\rho_D}{\partial L / \partial D} \;=\; -\frac{\rho_C k N}{\partial L / \partial T} \;=\; \mu,
\label{eq:foc-loss}
\end{equation}
i.e., \emph{equate the marginal dollar cost per unit of loss reduction across all three axes} (the minus signs render $\mu \geq 0$ since $\partial L/\partial X < 0$ at any interior optimum). Combined with the loss constraint $L(N, D, T) = L^*$, Equation~\eqref{eq:foc-loss} pins down a unique $(N^*, D^*, T^*)$. Substituting into Equation~\eqref{eq:cost} gives $\mathcal{B}^*(L^*) = \rho_D D^* + \rho_C k N^* T^*$, the minimum dollar spend to reach $L^*$.

\paragraph{(P2) Budget-constrained optimum.} Form $\mathcal{L}_2 = L + \lambda (\mathcal{B} - \mathcal{B}_\text{max})$ with $\lambda \geq 0$. The stationarity conditions are
\begin{equation}
\frac{\partial L / \partial N}{\rho_C k T} \;=\; \frac{\partial L / \partial D}{\rho_D} \;=\; \frac{\partial L / \partial T}{\rho_C k N} \;=\; -\lambda,
\label{eq:foc}
\end{equation}
i.e., \emph{equate the marginal loss reduction per dollar across all three axes}. Together with $\mathcal{B}(N, D, T) = \mathcal{B}_\text{max}$ this pins down $(N^*, D^*, T^*)$ and yields $L^* = L(N^*, D^*, T^*)$. The two sets of stationarity conditions are reciprocals (Equation~\eqref{eq:foc-loss} divides what Equation~\eqref{eq:foc} multiplies), so the two optima trace out the same Pareto frontier $\{(\mathcal{B}^*, L^*)\}$ and either problem can be read off the other.

\paragraph{Data already in hand.} A practitioner who already holds $D_0$ examples pays only for what they generate on top of them, so the budget becomes
\begin{equation}
\mathcal{B}(N, D, T) = \rho_D \cdot (D - D_0) + \rho_C \cdot k N T, \qquad D \ge D_0,
\label{eq:cost-existing}
\end{equation}
and $D_0 = 0$ recovers Equation~\eqref{eq:cost}. Restricting to $D \ge D_0$ rules out nothing worth considering: training on fewer examples than one already has refunds no money, since the budget pays only for generation beyond $D_0$, and raises the loss, since $\partial L / \partial D < 0$. Both problems' stationarity conditions are unchanged, because $\partial \mathcal{B} / \partial D = \rho_D$ either way. A sunk cost shifts the objective by a constant, leaving the marginal condition unchanged. So whenever the unconstrained optimum satisfies $D^* \ge D_0$, (P1) returns the same $(N^*, D^*, T^*)$ and reports a cost lower by $\rho_D D_0$, and (P2) under budget $\mathcal{B}_\text{max}$ is exactly (P2) with no existing data under budget $\mathcal{B}_\text{max} + \rho_D D_0$.

\paragraph{Feasibility and boundary behavior.} The form's $[E, L_0]$ ceiling makes infeasibility of (P1) detectable in closed form. Inverting the wrapper, the difficulty $h$ required to hit a target $L^*$ is
\begin{equation}
h^* \;=\; \frac{L^* - E}{L_0 - L^*},
\label{eq:h-target}
\end{equation}
which is positive and finite iff $L^* \in (E, L_0)$. Outside this open interval the constraint manifold $\{(N, D, T) : L(N, D, T) = L^*\}$ is empty:
\begin{itemize}
\itemsep -0.05em
  \item \emph{$L^* < E$}: asks for a model below the irreducible loss; (P1) has no solution. The Lagrangian's stationarity conditions become inconsistent and the Newton iteration diverges or stalls. To diagnose this, compute $h^*$ and check the sign.
  \item \emph{$L^* \geq L_0$}: trivially achievable by an untrained model. The optimizer pushes one of $a/N^\alpha$, $b/T^\beta$, or $c\,N^\gamma/D^\delta$ toward infinity (e.g., $T \to 0$), so $\mathcal{B}^* \to 0$. (P1) at this target is well-defined but uninformative.
  \item \emph{$L^* \to E^+$}: $h^* \to 0$, which requires $a/N^\alpha + b/T^\beta + c\,N^\gamma/D^\delta \to 0$ jointly. This forces $N \to \infty$ and $T \to \infty$, with $D \to \infty$ along a path where $D^\delta$ outpaces $N^\gamma$ in the overfitting term (see Appendix~\ref{app:limits}, Row 6). The cost $\mathcal{B}^* \to \infty$. So $L^* = E$ is a limit point of the feasible region.
\end{itemize}
Within the feasible interior $L^* \in (E, L_0)$, the optimum is attained at \emph{finite} $(N^*, D^*, T^*)$. At any fixed $D$, the form admits a finite interior minimum $N^*(D)$ from the U-shape competition between the capacity term and the overfitting term (Appendix~\ref{app:nopt} derives $N^*_\infty(D) = (\alpha a D^\delta / \gamma c)^{1/(\alpha+\gamma)}$ in the asymptotic $T \to \infty$ limit), achieving $L^*(D) > E$ at that finite $N^*(D)$. The family $\{L^*(D) : D > 0\}$ traces out an achievable frontier that approaches $E$ only as $D \to \infty$. (P1) at any target above this frontier has a finite-allocation solution; only $L^* = E$ requires the joint $\infty$ limit. The numerical solver should compute $h^*$ first via Equation~\eqref{eq:h-target} and reject targets outside $(E, L_0)$ before iterating.

\paragraph{Recipe.} The general system has no closed form, so we solve numerically. The procedure for either (P1) or (P2):
\begin{enumerate}
\itemsep -0.05em
  \item Calibrate the form on a domain-specific training grid to obtain $(E, a, b, c, \alpha, \beta, \gamma, \delta)$ and $L_0$ (Section~\ref{sec:experiments}).
  \item Specify the price pair $(\rho_D, \rho_C)$ for the deployment setting and an architecture-dependent FLOP-per-step constant $k$.
  \item Solve Equation~\eqref{eq:foc-loss} (for P1) or~\eqref{eq:foc} (for P2) jointly with the active constraint over $(N, D, T)$ until the stationarity conditions hold to tolerance.
  \item Read off the dual quantities: $\mathcal{B}^*$ and $(N^*, D^*, T^*, T^*/D^*)$ for P1, or $L^*$ and the same allocation for P2.
\end{enumerate}

\paragraph{Cost ratios and the data-constrained regime.} A useful summary statistic is the dollar-cost ratio $\eta = \rho_D / \rho_C$. For LLM pretraining on web-scraped data, $\eta$ is small (data is cheap, compute dominates) and the Chinchilla-style optimum approximates both (P1) and (P2). For physical-simulation surrogates, RL data generation, and expert label acquisition, $\eta$ is large (each example is expensive). As $\eta$ grows, the optimum shifts toward smaller $D$ (fewer expensive examples), more epochs, and more compute spent revisiting the same data.

\paragraph{Compute is not exactly proportional to dollars.} Equation~\eqref{eq:cost} treats compute as linear in dollars via a constant $\rho_C$, but in practice users pay for GPU-hours, not FLOPs. For instance, small models cannot fully utilize a multi-GPU node, and large models incur communication overhead. The apparent $\rho_C$ therefore depends on $N$ and on the cluster topology. Accurate cost-aware fits should treat $\rho_C(N)$ as a function calibrated to the actual hardware.

\paragraph{Stationarity conditions in terms of $\partial h$.} The partial derivatives of $h = a/N^\alpha + b/T^\beta + c\,N^\gamma/D^\delta$ are
\[
\frac{\partial h}{\partial N} = -\frac{\alpha a}{N^{\alpha+1}} + \frac{\gamma c \, N^{\gamma-1}}{D^\delta}, \quad
\frac{\partial h}{\partial D} = -\frac{\delta c \, N^\gamma}{D^{\delta+1}}, \quad
\frac{\partial h}{\partial T} = -\frac{\beta b}{T^{\beta+1}},
\]
and $\partial L / \partial X = (L_0 - E)(\partial h / \partial X)/(1+h)^2$ for $X \in \{N, D, T\}$. Substituting into Equations~\eqref{eq:foc-loss} and~\eqref{eq:foc} gives the stationarity conditions in terms of $\partial h$:
\begin{align}
\text{(P2)}: \quad &\frac{1}{\rho_C k T} \frac{\partial h}{\partial N} = \frac{1}{\rho_D} \frac{\partial h}{\partial D} = \frac{1}{\rho_C k N} \frac{\partial h}{\partial T} = -\frac{\lambda (1+h)^2}{(L_0 - E)}, \label{eq:foc2-app} \\
\text{(P1)}: \quad &\rho_C k T \cdot \left(\frac{\partial h}{\partial N}\right)^{-1} = \rho_D \cdot \left(\frac{\partial h}{\partial D}\right)^{-1} = \rho_C k N \cdot \left(\frac{\partial h}{\partial T}\right)^{-1} = -\frac{\mu (L_0 - E)}{(1+h)^2}, \label{eq:foc1-app}
\end{align}
each system four equations in four unknowns ($\{N^*, D^*, T^*, \lambda\}$ for P2, replacing $\lambda$ with $\mu$ for P1, with $\mu = 1/\lambda$ at the dual optimum).

\paragraph{Asymptotic regimes.} (i) When $\rho_D \gg \rho_C$ (data dominates cost): $D^*$ is small, multi-epoch training is preferred, $T^*$ scales freely until the overfitting term $c\,N^\gamma/D^\delta$ saturates the wrapper. (ii) When $\rho_D \ll \rho_C$ (compute dominates cost): the stationarity condition for $D$ pushes $D^* \to \infty$, the overfitting term vanishes, and $(N^*, T^*)$ approach the Chinchilla compute-optimal allocation in $(N, D{=}T)$ variables; the Chinchilla single-epoch convention $T = D$ is recovered by selecting $D$ at the smallest value already killing the overfitting term (once $cN^\gamma/D^\delta$ is negligible, the loss is invariant under further $D$ increases).

\paragraph{Log-convexity and uniqueness.} Minimizing $L$ is equivalent to minimizing $h(N, D, T) = a/N^\alpha + b/T^\beta + c\,N^\gamma/D^\delta$, since $L$ is monotone in $h$ (the wrapper $h/(1+h)$ is monotone non-decreasing). Substituting $u = \log N$, $v = \log D$, $w = \log T$, $h = a\,e^{-\alpha u} + b\,e^{-\beta w} + c\,e^{\gamma u - \delta v}$, a sum of exponentials of linear functions of $(u, v, w)$, which is convex (each $e^{\text{affine}}$ term is convex; sum of convex is convex). In fact $h$ is \emph{strictly} convex in $(u, v, w)$ when $a, b, c, \alpha, \beta, \gamma, \delta > 0$: its Hessian is block-diagonal in $\{(u, v), w\}$, with the $(w, w)$ entry $b\beta^2 e^{-\beta w} > 0$ and the $(u, v)$ block having determinant $a c\, \alpha^2 \delta^2 e^{(\gamma - \alpha) u - \delta v} > 0$ and positive trace, so both blocks are positive definite. The dollar cost $\mathcal{B} = \rho_D e^v + \rho_C k e^{u + w}$ is convex in $(u, v, w)$, so the feasible set $\{\mathcal{B} \leq \mathcal{B}_\text{max}\}$ is convex. Minimizing a strictly convex function on a convex feasible set yields a unique optimum, so (P1) and (P2) each have a unique solution $(N^*, D^*, T^*)$, and standard convex-optimization tools converge to it from any feasible starting point. (At the boundary, $c = 0$ or $b = 0$, the strict convexity fails and the optimum on the $D$ or $T$ axis becomes degenerate.)

\section{Conformal prediction}
\label{app:conformal-prediction}

When using a fitted form to make a new prediction, we would like to quantify our uncertainty in that prediction. A tempting approach would follow the pattern used when reporting confidence intervals throughout the paper; i.e., redraw bootstrap samples from the training set, refit the form on each, and report the spread of predictions across the ensemble. However, this approach would underestimate the uncertainty of the prediction because the bootstrap interval models the spread of the fit itself, not the spread of a new observation around that fit. The interval shrinks as the number of grid points increases, and it does not capture, for instance, a systematic offset caused by a misspecified form or the inherent variation of noisy observations. Instead, the standard approach is to use \emph{conformal prediction} to build a calibrated interval for the prediction~\citep{vovk2005algorithmic,lei2018distribution}.

\begin{figure}
  \centering
  \includegraphics[width=\linewidth]{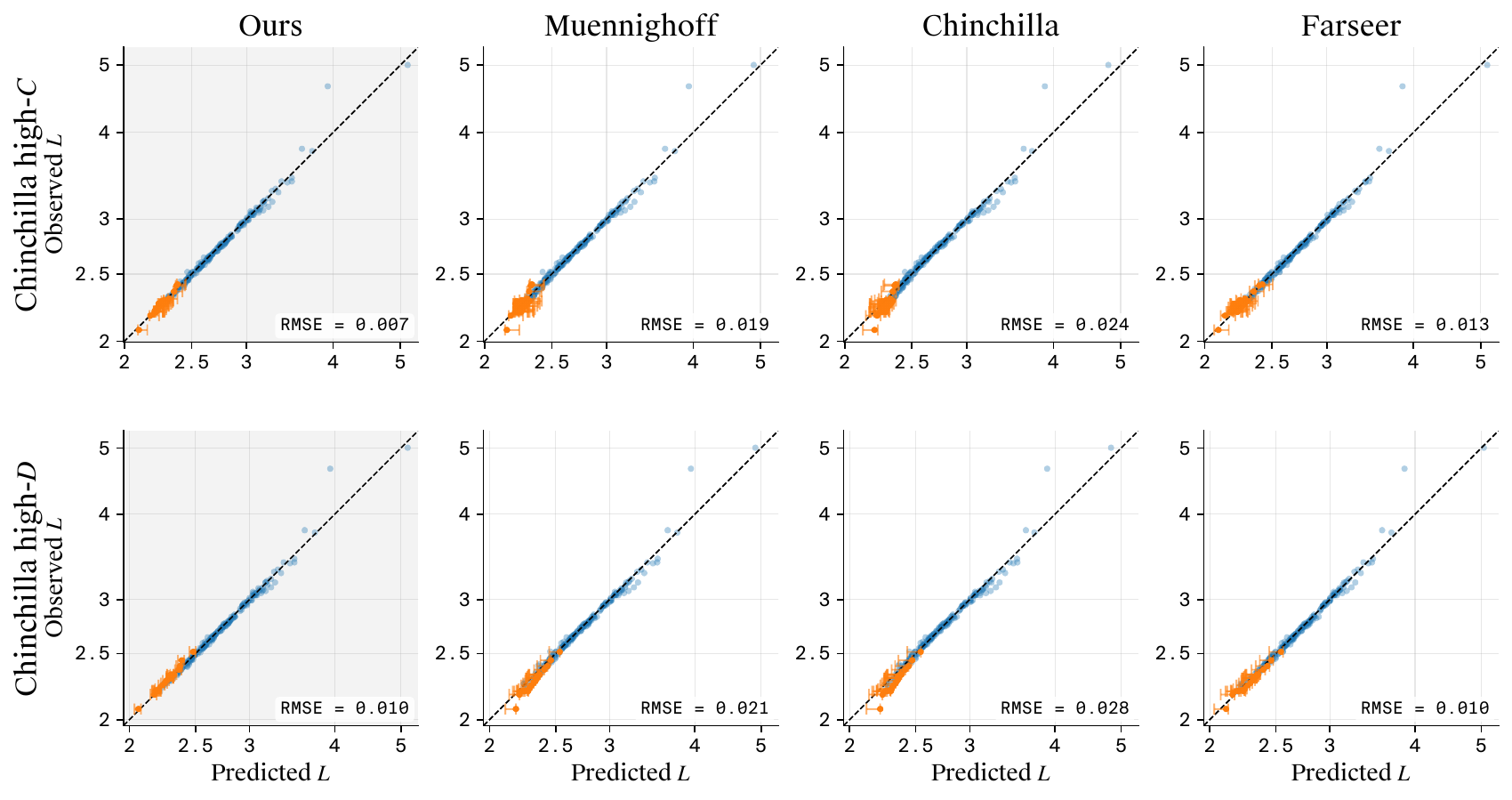}
  \caption{\textbf{Chinchilla high-$C$ and high-$D$ conformal predictions.} Our form compared against the Muennighoff, Chinchilla, and Farseer forms on the Chinchilla dataset. Blue: training cells. Orange: held-out cells, with the 90\% asymmetric conformal band of Eq.~\eqref{eq:conformal}.}
  \label{fig:extrap-grid-chinchilla}
\end{figure}

\begin{figure}
  \centering
  \includegraphics[width=\linewidth]{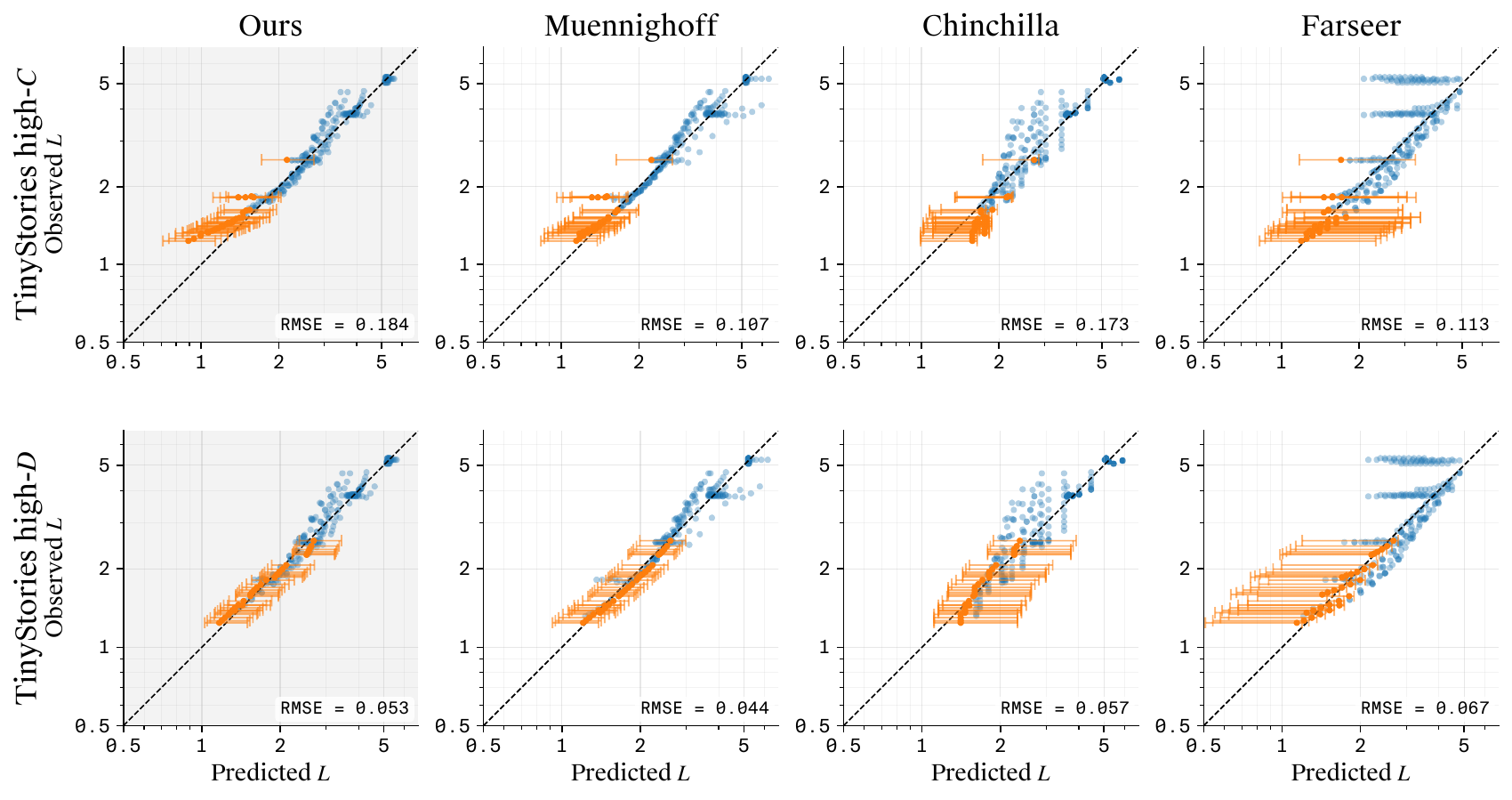}
  \caption{\textbf{TinyStories high-$C$ and high-$D$ conformal predictions.} Our form compared against the Muennighoff, Chinchilla, and Farseer forms on the TinyStories dataset. Blue: training cells. Orange: held-out cells, with the 90\% asymmetric conformal band of Eq.~\eqref{eq:conformal}. Notice that because the TinyStories grid is difficult to accurately model, the conformal bands across all forms are wider than the intervals built on the Chinchilla grid.}
  \label{fig:extrap-grid-tinystories}
\end{figure}

\paragraph{Calibration procedure.} Conformal prediction assumes \emph{exchangeability} of the new prediction and the training set, so to build a calibration set for extrapolation, we must choose a calibration set that lies along the axis we are extrapolating. Rather than creating a single held-out subset, we create many nested subsets for robustness according to the following procedure:
\begin{enumerate}
\itemsep -0.05em
  \item Sort the training rows along the axis to be extrapolated.
  \item Cut the sorted rows at $40\%$.
  \item Fit the form on the rows below the cut.
  \item Divide the rows above the cut into slices of $10\%$.
  \item Predict the loss of every row in every slice above the cut.
  \item Record the error $\log \hat{L} - \log L$ of each prediction.
  \item Move the cut up by $5\%$. Do steps 3 to 6 again. Stop when the cut passes $85\%$.
\end{enumerate}

Once we have gathered all the errors, we read off the quantiles that leave $\alpha$ of them outside. Writing the sorted errors as $r_1 \le \cdots \le r_n$, the $1 - \alpha$ prediction interval for a new point whose predicted loss is $\hat{L}$ is
\begin{equation}
  \Big[\, \hat{L}\,e^{-\hat{q}_{+}}, \;\; \hat{L}\,e^{-\hat{q}_{-}} \,\Big],
  \qquad
  \hat{q}_{+} = r_{(\lceil (n+1)(1-\alpha/2)\rceil)}, \quad
  \hat{q}_{-} = -\big(-r\big)_{(\lceil (n+1)(1-\alpha/2)\rceil)},
  \label{eq:conformal}
\end{equation}
where $r_{(k)}$ is the $k$-th smallest error and $\alpha = 0.1$ is the target miss rate. We take $\alpha/2$ from each tail, so this value of $\alpha$ gives an asymmetric 90\% interval. In our case, to obtain the prediction interval for a new $(N, D, T)$ cell, compute $\hat{L}$ from the fitted form, and multiply it by $e^{-\hat{q}_{+}}$ and $e^{-\hat{q}_{-}}$ for the lower and upper limit, respectively. In Figures~\ref{fig:extrap-grid-chinchilla} and~\ref{fig:extrap-grid-tinystories} we show the conformal predictions for the Chinchilla and TinyStories grids, respectively, comparing our form to the Muennighoff, Chinchilla, and Farseer forms.

\section{Far extrapolation}
\label{app:far-extrapolation}

While our form is designed to make predictions in the data-constrained regime, one might ask how well it performs in a realistic data-rich LLM scaling prediction regime. \citet{li2025farseer} introduce a dataset of 7 large-model runs held back from the 400 they fit, which are beyond the scope of our main-paper extrapolations. These 7 runs sit $1.7$ to $5.4\times$ beyond the cohort in compute and form the basis of the Farseer ``far extrapolation'' benchmark, which we reproduce here. We find that when asked to predict these far-compute targets, our form performs as well as Farseer's own form on the grid it was built specifically for (substantially better than Chinchilla's form). We report the results in Figure~\ref{fig:far-extrap} and show conformal intervals (as described in Appendix~\ref{app:conformal-prediction}) in Figure~\ref{fig:extrap-grid-farseer}.

\begin{figure}[ht]
  \centering
  \includegraphics[width=\linewidth]{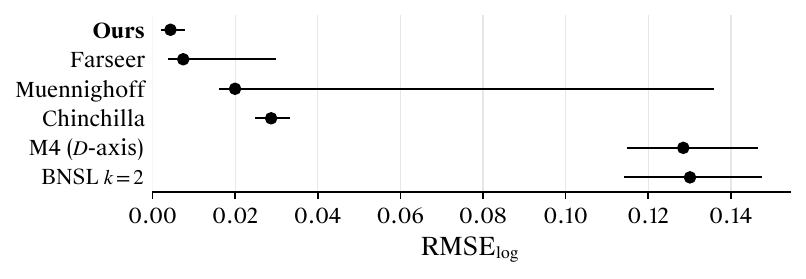}
  \caption{\textbf{Farseer far-compute extrapolation $\text{RMSE}_{\log}$.} Each form is fitted on the 400-cell cohort and asked to predict the 7 held-out large-model runs. Dots are the median $\text{RMSE}_{\log}$ over 200 bootstrap resamples, and whiskers span the 2.5\% and 97.5\% quantiles of the same draws. Our form achieves both the lowest error and the narrowest interval of all forms.}
  \label{fig:far-extrap}
\end{figure}

\begin{figure}[t]
  \centering
  \includegraphics[width=\linewidth]{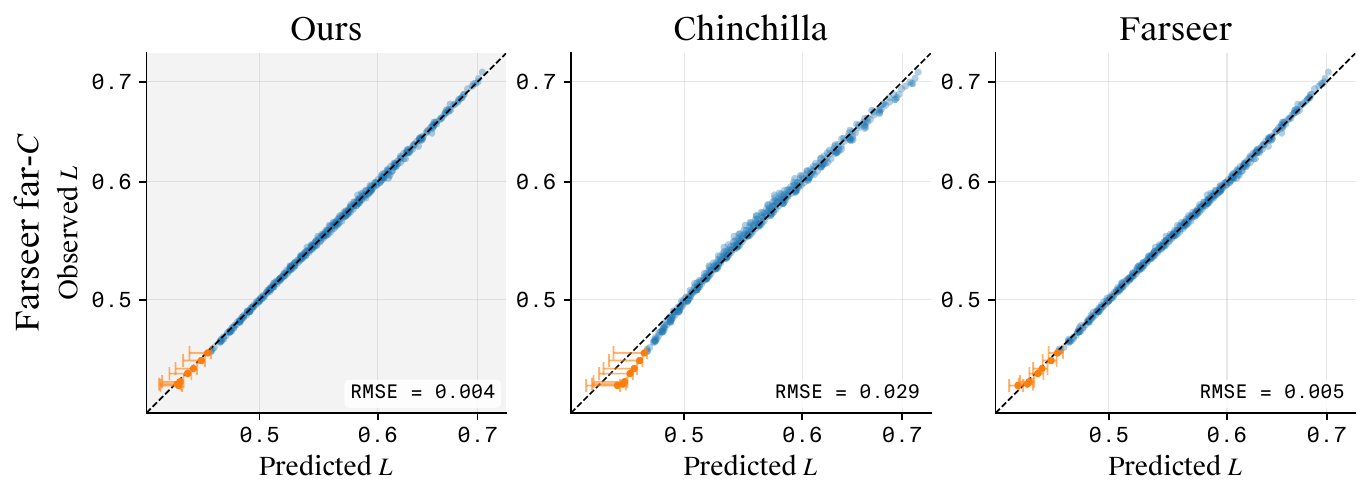}
  \caption{\textbf{Farseer far-compute extrapolation conformal intervals.} Comparison of far-$C$ predictions between our form, the Chinchilla form, and the Farseer form. Blue: training cells. Orange: held-out cells, with the 90\% asymmetric conformal band of Eq.~\eqref{eq:conformal} around the prediction point estimates from each form. Both ours and Farseer's predictions are close to the observed values with tight conformal intervals, while Chinchilla overestimates the loss and has a wider interval.}
  \label{fig:extrap-grid-farseer}
\end{figure}

\section{Full results}
\label{app:results}

\subsection{Full experiment results and ablations}
\label{app:full-results}

The three tables below report each form's RMSE on each dataset under four protocols: in-sample, the full-dataset fit with no holdout (Table~\ref{tab:in-sample}); 5-fold cross-validation (Table~\ref{tab:holdout-random}); and high-$C$ and high-$D$ holdouts (Table~\ref{tab:holdout-axis}, both defined in \S\ref{sec:empirical-setup}). Forms above the horizontal rule are external baselines fit to the loss surface; below the rule, our form (Eq.~\eqref{eq:ours}) and four ablations of it. Datasets to the left of the vertical rule are from our experiments (Section~\ref{sec:experiments}); to the right are published grids on which we refit each form.

\paragraph{In-sample fit.} Our form (or one of its wrapper-choice and single-exponent siblings) attains the lowest in-sample RMSE on every column of Table~\ref{tab:in-sample}. We report this metric for completeness because it is the metric that much of the scaling-law literature focuses on, but it is deliberately not the focus of this paper. A form with three additional free parameters mechanically fits its training set better than a baseline, so its strong performance on this metric is not surprising. The practical use of a fitted scaling law comes instead from extrapolation, not interpolation.

\paragraph{Random 5-fold cross validation.} Random 5-fold tests near-interpolation: each held-out cell is bracketed by training cells on every axis. Our form family wins eight of nine columns (Table~\ref{tab:holdout-random}); the only loss is Chinchilla, where the Muennighoff and Farseer forms have a lower median RMSE.

\paragraph{High-$C$ and high-$D$ extrapolation ablations.} Here, we perform ablations of our form to investigate the impact of each component of our form on its prediction performance on our main metrics of interest: high-$C$ and high-$D$ extrapolation. In our own training grids that exercise the high $T/D$ regime, we observe the largest degradation in performance from the removal of the overfitting term $c\,N^\gamma/D^\delta$. The next largest performance degradation comes from tying the $N/D$ term to a single exponent $\gamma$ instead of one exponent per parameter. Next, we see slight degradation from removing the saturating wrapper. There is no significant impact from using an exponential wrapper $1-e^{-h}$ instead of our $h/(1+h)$ wrapper. In the large-scale LLM refits, the difference between our form variants is less pronounced. We observe slight degradation when we remove the overfitting term, but actually observe \emph{improvement} when we use a single exponent for the ${N/D}$ term. Adding the overfitting term provides extra flexibility that helps the fit surface conform to the true scaling shape, while the shared exponent provides regularization that aids in the identification of the fit parameters when $T$ and $D$ are tightly coupled in the training grids.

\begin{table}[ht]
\centering
\caption{\textbf{In-Sample.} In-sample training RMSE (log space) under the full-data fit. Each form is fit on every observed cell. Lower is better. Per column, \textbf{best} and \underline{second best} mark the median. Values are the median over 200 bootstrap resamples of the training set, each refit from scratch; parentheses give their 2.5\% and 97.5\% quantiles.}
\label{tab:in-sample}
\resizebox{\linewidth}{!}{%
\setlength{\aboverulesep}{0pt}\setlength{\belowrulesep}{0pt}%
\begin{tabular}{@{}>{\rowstrut}l|cccc|ccccc@{}}
\toprule
\rowcolor{black!10} \multicolumn{10}{c}{\rowstrut \textbf{In-Sample}} \\
\midrule
Form $\downarrow$ & MNIST & CIFAR-100 & Darcy & TinyStories & Chinchilla & Muennighoff & Gadre & Porian & Farseer \\
\midrule
Chinchilla~\citep{hoffmann2022training}      & \msci{0.230}{0.229}{0.231} & \msci{0.125}{0.121}{0.134} & \msci{0.272}{0.264}{0.283} & \msci{0.137}{0.133}{0.142} & \msci{0.018}{0.017}{0.024} & \msci{0.208}{0.203}{0.213} & \msci{0.024}{0.023}{0.041} & \msci{0.062}{0.058}{0.071} & \msci{0.009}{0.009}{0.009} \\
Muennighoff~\citep{muennighoff2023scaling}   & \msci{0.202}{0.190}{0.233} & \underline{\msci{0.064}{0.059}{0.134}} & \msci{0.130}{0.118}{0.185} & \msci{0.085}{0.081}{0.164} & \msci{0.015}{0.015}{0.061} & \msci{0.181}{0.175}{0.188} & \msci{0.020}{0.018}{0.043} & \msci{0.057}{0.055}{0.074} & \msci{0.007}{0.007}{0.060} \\
M4 ($N$-axis)~\citep{alabdulmohsin2022revisiting} & \msci{0.970}{0.965}{0.977} & \msci{0.535}{0.530}{0.631} & \msci{0.824}{0.816}{0.845} & \msci{0.398}{0.393}{0.425} & \msci{0.104}{0.103}{0.107} & \msci{0.280}{0.269}{0.288} & \msci{0.093}{0.088}{0.104} & \msci{0.314}{0.299}{0.337} & \msci{0.086}{0.084}{0.088} \\
M4 ($D$-axis)~\citep{alabdulmohsin2022revisiting} & \msci{0.224}{0.223}{0.240} & \msci{0.146}{0.139}{0.161} & \msci{0.301}{0.296}{0.312} & \msci{0.200}{0.196}{0.212} & \msci{0.064}{0.063}{0.066} & \msci{0.217}{0.204}{0.226} & \msci{0.090}{0.086}{0.101} & \msci{0.099}{0.098}{0.103} & \msci{0.060}{0.059}{0.061} \\
BNSL $k{=}1$~\citep{caballero2023broken}     & \msci{0.224}{0.223}{0.225} & \msci{0.146}{0.140}{0.163} & \msci{0.302}{0.296}{0.333} & \msci{0.196}{0.191}{0.212} & \msci{0.064}{0.063}{0.066} & \msci{0.217}{0.205}{0.226} & \msci{0.090}{0.086}{0.103} & \msci{0.100}{0.098}{0.104} & \msci{0.060}{0.059}{0.061} \\
BNSL $k{=}2$~\citep{caballero2023broken}     & \msci{0.224}{0.223}{0.225} & \msci{0.146}{0.140}{0.160} & \msci{0.302}{0.296}{0.318} & \msci{0.196}{0.189}{0.211} & \msci{0.064}{0.063}{0.066} & \msci{0.217}{0.204}{0.225} & \msci{0.090}{0.086}{0.123} & \msci{0.100}{0.098}{0.104} & \msci{0.060}{0.059}{0.061} \\
Farseer~\citep{li2025farseer}                & \msci{0.778}{0.774}{0.784} & \msci{0.568}{0.531}{0.643} & \msci{0.484}{0.469}{0.513} & \msci{0.261}{0.241}{0.284} & \msci{0.015}{0.014}{0.024} & \msci{0.245}{0.239}{0.250} & \msci{0.019}{0.015}{0.037} & \msci{0.055}{0.052}{0.062} & \msci{0.004}{0.003}{0.009} \\
\midrule
\textbf{Ours} (Eq.~\eqref{eq:ours})          & \msci{0.130}{0.129}{0.133} & \textbf{\msci{0.062}{0.060}{0.067}} & \textbf{\msci{0.117}{0.113}{0.124}} & \underline{\msci{0.076}{0.074}{0.080}} & \textbf{\msci{0.014}{0.013}{0.047}} & \underline{\msci{0.128}{0.117}{0.143}} & \underline{\msci{0.015}{0.012}{0.094}} & \underline{\msci{0.047}{0.046}{0.052}} & \textbf{\msci{0.003}{0.003}{0.003}} \\
Ours, no wrapper                             & \underline{\msci{0.128}{0.127}{0.164}} & \msci{0.114}{0.109}{0.133} & \msci{0.128}{0.124}{0.133} & \msci{0.077}{0.076}{0.080} & \textbf{\msci{0.014}{0.013}{0.193}} & \msci{0.143}{0.131}{0.178} & \msci{0.018}{0.015}{0.128} & \msci{0.062}{0.058}{0.071} & \textbf{\msci{0.003}{0.003}{0.004}} \\
Ours, no overfitting term                    & \msci{0.750}{0.746}{0.755} & \msci{0.535}{0.529}{0.632} & \msci{0.485}{0.474}{0.507} & \msci{0.247}{0.241}{0.261} & \msci{0.020}{0.018}{0.025} & \msci{0.247}{0.239}{0.251} & \msci{0.018}{0.018}{0.029} & \underline{\msci{0.047}{0.046}{0.051}} & \msci{0.007}{0.007}{0.007} \\
Ours, $1{-}e^{-h}$ wrapper                   & \textbf{\msci{0.110}{0.109}{0.112}} & \msci{0.067}{0.064}{0.074} & \underline{\msci{0.121}{0.117}{0.128}} & \textbf{\msci{0.075}{0.073}{0.078}} & \textbf{\msci{0.014}{0.013}{0.048}} & \msci{0.134}{0.120}{0.148} & \msci{0.016}{0.014}{0.098} & \textbf{\msci{0.046}{0.046}{0.053}} & \textbf{\msci{0.003}{0.003}{0.003}} \\
Ours, single-exp $(N/D)^\gamma$              & \msci{0.481}{0.479}{0.489} & \msci{0.278}{0.265}{0.328} & \msci{0.280}{0.276}{0.286} & \msci{0.123}{0.119}{0.134} & \textbf{\msci{0.014}{0.013}{0.047}} & \textbf{\msci{0.125}{0.117}{0.143}} & \textbf{\msci{0.014}{0.013}{0.038}} & \underline{\msci{0.047}{0.046}{0.052}} & \textbf{\msci{0.003}{0.003}{0.003}} \\
\bottomrule
\end{tabular}
}
\end{table}

\begin{table}[ht]
\centering
\caption{\textbf{Random 5-fold Cross Validation.} Held-out RMSE (log space) under 5-fold random cross-validation. Each fold holds out a uniform-random 20\% of the \emph{rows}. Lower is better. Per column, \textbf{best} and \underline{second best} mark the mean. Values are the mean across the 5 folds; parentheses give their range.}
\label{tab:holdout-random}
\resizebox{\linewidth}{!}{%
\setlength{\aboverulesep}{0pt}\setlength{\belowrulesep}{0pt}%
\begin{tabular}{@{}>{\rowstrut}l|cccc|ccccc@{}}
\toprule
\rowcolor{black!10} \multicolumn{10}{c}{\rowstrut \textbf{Random 5-fold Cross Validation}} \\
\midrule
Form $\downarrow$ & MNIST & CIFAR-100 & Darcy & TinyStories & Chinchilla & Muennighoff & Gadre & Porian & Farseer \\
\midrule
Chinchilla~\citep{hoffmann2022training}      & \msci{0.230}{0.225}{0.238} & \msci{0.129}{0.121}{0.149} & \msci{0.273}{0.236}{0.314} & \msci{0.138}{0.109}{0.156} & \msci{0.017}{0.009}{0.031} & \msci{0.204}{0.134}{0.269} & \msci{0.027}{0.013}{0.048} & \msci{0.064}{0.050}{0.079} & \msci{0.009}{0.008}{0.009} \\
Muennighoff~\citep{muennighoff2023scaling}   & \msci{0.202}{0.199}{0.206} & \msci{0.089}{0.059}{0.146} & \msci{0.125}{0.099}{0.146} & \msci{0.084}{0.061}{0.099} & \textbf{\msci{0.014}{0.008}{0.027}} & \msci{0.180}{0.122}{0.235} & \msci{0.024}{0.013}{0.039} & \msci{0.057}{0.049}{0.066} & \msci{0.007}{0.006}{0.007} \\
M4 ($N$-axis)~\citep{alabdulmohsin2022revisiting} & \msci{0.971}{0.946}{0.987} & \msci{0.538}{0.514}{0.566} & \msci{0.826}{0.781}{0.854} & \msci{0.398}{0.358}{0.420} & \msci{0.100}{0.062}{0.137} & \msci{0.278}{0.212}{0.340} & \msci{0.093}{0.057}{0.137} & \msci{0.315}{0.266}{0.335} & \msci{0.086}{0.080}{0.091} \\
M4 ($D$-axis)~\citep{alabdulmohsin2022revisiting} & \msci{0.224}{0.220}{0.229} & \msci{0.144}{0.102}{0.218} & \msci{0.302}{0.258}{0.345} & \msci{0.200}{0.156}{0.220} & \msci{0.064}{0.061}{0.072} & \msci{0.215}{0.166}{0.269} & \msci{0.094}{0.069}{0.167} & \msci{0.099}{0.083}{0.121} & \msci{0.060}{0.058}{0.062} \\
BNSL $k{=}1$~\citep{caballero2023broken}     & \msci{0.224}{0.220}{0.229} & \msci{0.145}{0.100}{0.218} & \msci{0.302}{0.255}{0.346} & \msci{0.198}{0.165}{0.217} & \msci{0.064}{0.060}{0.072} & \msci{0.214}{0.166}{0.269} & \msci{0.095}{0.071}{0.172} & \msci{0.100}{0.085}{0.122} & \msci{0.060}{0.058}{0.063} \\
BNSL $k{=}2$~\citep{caballero2023broken}     & \msci{0.224}{0.220}{0.229} & \msci{0.145}{0.100}{0.218} & \msci{0.304}{0.255}{0.347} & \msci{0.197}{0.160}{0.217} & \msci{0.064}{0.060}{0.072} & \msci{0.214}{0.166}{0.269} & \msci{0.095}{0.069}{0.172} & \msci{0.101}{0.087}{0.122} & \msci{0.060}{0.058}{0.063} \\
Farseer~\citep{li2025farseer}                & \msci{0.778}{0.767}{0.789} & \msci{0.564}{0.526}{0.624} & \msci{0.485}{0.451}{0.523} & \msci{0.258}{0.219}{0.276} & \underline{\msci{0.015}{0.006}{0.029}} & \msci{0.238}{0.167}{0.305} & \msci{0.021}{0.014}{0.035} & \msci{0.057}{0.051}{0.067} & \textbf{\msci{0.003}{0.002}{0.004}} \\
\midrule
\textbf{Ours} (Eq.~\eqref{eq:ours})          & \msci{0.130}{0.127}{0.131} & \textbf{\msci{0.062}{0.052}{0.072}} & \textbf{\msci{0.118}{0.106}{0.132}} & \underline{\msci{0.076}{0.050}{0.090}} & \msci{0.028}{0.005}{0.090} & \textbf{\msci{0.130}{0.082}{0.158}} & \msci{0.024}{0.005}{0.064} & \textbf{\msci{0.047}{0.042}{0.057}} & \textbf{\msci{0.003}{0.003}{0.003}} \\
Ours, no wrapper                             & \underline{\msci{0.128}{0.127}{0.132}} & \msci{0.118}{0.106}{0.133} & \msci{0.129}{0.117}{0.141} & \msci{0.078}{0.060}{0.086} & \msci{0.058}{0.005}{0.240} & \msci{0.151}{0.109}{0.177} & \msci{0.035}{0.007}{0.096} & \msci{0.064}{0.050}{0.079} & \textbf{\msci{0.003}{0.003}{0.004}} \\
Ours, no overfitting term                    & \msci{0.751}{0.731}{0.771} & \msci{0.539}{0.518}{0.566} & \msci{0.485}{0.464}{0.508} & \msci{0.249}{0.238}{0.258} & \msci{0.018}{0.008}{0.034} & \msci{0.242}{0.176}{0.310} & \underline{\msci{0.020}{0.009}{0.035}} & \msci{0.048}{0.042}{0.057} & \msci{0.007}{0.007}{0.008} \\
Ours, $1{-}e^{-h}$ wrapper                   & \textbf{\msci{0.110}{0.107}{0.112}} & \underline{\msci{0.069}{0.055}{0.077}} & \underline{\msci{0.123}{0.110}{0.136}} & \textbf{\msci{0.075}{0.051}{0.087}} & \msci{0.031}{0.005}{0.103} & \msci{0.137}{0.097}{0.167} & \msci{0.028}{0.006}{0.075} & \textbf{\msci{0.047}{0.043}{0.053}} & \textbf{\msci{0.003}{0.003}{0.003}} \\
Ours, single-exp $(N/D)^\gamma$              & \msci{0.482}{0.454}{0.495} & \msci{0.279}{0.234}{0.312} & \msci{0.280}{0.250}{0.298} & \msci{0.125}{0.103}{0.137} & \msci{0.029}{0.006}{0.095} & \textbf{\msci{0.130}{0.083}{0.159}} & \textbf{\msci{0.015}{0.002}{0.027}} & \msci{0.048}{0.042}{0.057} & \textbf{\msci{0.003}{0.003}{0.003}} \\
\bottomrule
\end{tabular}
}
\end{table}

\begin{table}[ht]
\centering
\caption{\textbf{high-$C$ / high-$D$ extrapolation.} Held-out RMSE (log space) under the holdout protocols of \S\ref{sec:empirical-setup}. Lower is better. Per column within each pane, \textbf{best} and \underline{second best} mark the median. Values are the median over 200 bootstrap resamples of the training set, each refit from scratch; parentheses give their 2.5\% and 97.5\% quantiles.}
\label{tab:holdout-axis}
\resizebox{\linewidth}{!}{
\setlength{\aboverulesep}{0pt}\setlength{\belowrulesep}{0pt}%
\begin{tabular}{@{}>{\rowstrut}l|cccc|ccccc@{}}
\toprule
\rowcolor{black!10} \multicolumn{10}{c}{\rowstrut \textbf{high-$C$ holdout}} \\
\midrule
Form $\downarrow$ & MNIST & CIFAR-100 & Darcy & TinyStories & Chinchilla & Muennighoff & Gadre & Porian & Farseer \\
\midrule
Chinchilla~\citep{hoffmann2022training}      & \msci{0.331}{0.328}{0.334} & \msci{0.156}{0.137}{0.184} & \msci{0.322}{0.281}{0.380} & \msci{0.167}{0.134}{0.222} & \msci{0.023}{0.016}{0.027} & \msci{0.093}{0.072}{0.115} & \msci{0.037}{0.013}{0.060} & \msci{0.098}{0.071}{0.125} & \msci{0.022}{0.020}{0.025} \\
Muennighoff~\citep{muennighoff2023scaling}   & \msci{0.335}{0.316}{1.126} & \msci{0.091}{0.077}{0.142} & \msci{0.195}{0.123}{0.267} & \textbf{\msci{0.079}{0.057}{0.273}} & \msci{0.019}{0.012}{0.044} & \msci{0.083}{0.068}{0.102} & \msci{0.024}{0.008}{0.055} & \msci{0.083}{0.023}{0.125} & \msci{0.017}{0.014}{0.107} \\
M4 ($N$-axis)~\citep{alabdulmohsin2022revisiting} & \msci{1.117}{1.083}{1.136} & \msci{0.608}{0.607}{0.723} & \msci{1.303}{0.973}{1.369} & \msci{0.615}{0.542}{0.774} & \msci{0.070}{0.058}{0.088} & \msci{0.179}{0.121}{0.234} & \msci{0.075}{0.047}{0.163} & \msci{0.376}{0.301}{0.447} & \msci{0.126}{0.111}{0.143} \\
M4 ($D$-axis)~\citep{alabdulmohsin2022revisiting} & \msci{0.295}{0.267}{0.299} & \msci{0.125}{0.096}{0.180} & \msci{0.404}{0.364}{0.475} & \msci{0.331}{0.287}{0.412} & \msci{0.067}{0.058}{0.080} & \msci{0.094}{0.077}{0.128} & \msci{0.121}{0.071}{0.183} & \msci{0.112}{0.097}{0.145} & \msci{0.104}{0.091}{0.115} \\
BNSL $k{=}1$~\citep{caballero2023broken}     & \msci{0.308}{0.305}{0.312} & \msci{0.138}{0.093}{0.187} & \msci{0.408}{0.355}{0.484} & \msci{0.335}{0.287}{0.421} & \msci{0.068}{0.058}{0.079} & \msci{0.095}{0.077}{0.123} & \msci{0.128}{0.073}{0.195} & \msci{0.121}{0.100}{0.174} & \msci{0.103}{0.092}{0.115} \\
BNSL $k{=}2$~\citep{caballero2023broken}     & \msci{0.308}{0.305}{0.313} & \msci{0.138}{0.092}{0.191} & \msci{0.411}{0.367}{0.483} & \msci{0.336}{0.289}{0.420} & \msci{0.067}{0.058}{0.082} & \msci{0.095}{0.076}{0.126} & \msci{0.127}{0.070}{0.304} & \msci{0.122}{0.102}{0.181} & \msci{0.104}{0.091}{0.117} \\
Farseer~\citep{li2025farseer}                & \msci{1.363}{0.989}{1.497} & \msci{0.621}{0.606}{0.875} & \msci{0.343}{0.305}{0.535} & \msci{0.127}{0.110}{0.261} & \msci{0.015}{0.009}{0.032} & \msci{0.098}{0.092}{0.132} & \msci{0.037}{0.014}{0.114} & \msci{0.084}{0.050}{0.144} & \msci{0.007}{0.003}{0.021} \\
\midrule
\textbf{Ours} (Eq.~\eqref{eq:ours})          & \underline{\msci{0.127}{0.125}{0.129}} & \textbf{\msci{0.083}{0.067}{0.108}} & \msci{0.169}{0.140}{0.200} & \msci{0.183}{0.150}{0.232} & \textbf{\msci{0.008}{0.005}{0.012}} & \msci{0.060}{0.046}{0.097} & \textbf{\msci{0.014}{0.005}{0.091}} & \underline{\msci{0.063}{0.045}{0.092}} & \textbf{\msci{0.005}{0.004}{0.007}} \\
Ours, no wrapper                             & \msci{0.173}{0.168}{0.175} & \msci{0.153}{0.135}{0.178} & \textbf{\msci{0.148}{0.130}{0.173}} & \underline{\msci{0.123}{0.090}{0.155}} & \msci{0.009}{0.003}{0.013} & \underline{\msci{0.057}{0.043}{0.082}} & \msci{0.031}{0.017}{0.073} & \msci{0.098}{0.075}{0.124} & \msci{0.007}{0.004}{0.009} \\
Ours, no overfitting term                    & \msci{0.867}{0.848}{0.883} & \msci{0.609}{0.607}{0.887} & \msci{0.355}{0.308}{0.467} & \msci{0.235}{0.116}{0.312} & \msci{0.021}{0.014}{0.029} & \msci{0.114}{0.094}{0.151} & \msci{0.017}{0.006}{0.054} & \underline{\msci{0.063}{0.045}{0.090}} & \msci{0.020}{0.017}{0.022} \\
Ours, $1{-}e^{-h}$ wrapper                   & \textbf{\msci{0.123}{0.121}{0.153}} & \textbf{\msci{0.083}{0.063}{0.112}} & \underline{\msci{0.161}{0.138}{0.188}} & \msci{0.161}{0.128}{0.204} & \textbf{\msci{0.008}{0.004}{0.013}} & \msci{0.059}{0.044}{0.104} & \msci{0.023}{0.008}{0.084} & \msci{0.074}{0.053}{0.101} & \msci{0.006}{0.004}{0.008} \\
Ours, single-exp $(N/D)^\gamma$              & \msci{1.136}{1.127}{1.147} & \msci{0.290}{0.152}{0.470} & \msci{0.308}{0.262}{0.357} & \msci{0.200}{0.155}{0.262} & \msci{0.009}{0.003}{0.013} & \textbf{\msci{0.051}{0.043}{0.064}} & \textbf{\msci{0.014}{0.005}{0.063}} & \textbf{\msci{0.062}{0.044}{0.091}} & \textbf{\msci{0.005}{0.004}{0.006}} \\
\midrule[\heavyrulewidth]
\addlinespace[1em]
\midrule[\heavyrulewidth]
\rowcolor{black!10} \multicolumn{10}{c}{\rowstrut \textbf{high-$D$ holdout}} \\
\midrule
Form $\downarrow$ & MNIST & CIFAR-100 & Darcy & TinyStories & Chinchilla & Muennighoff & Gadre & Porian & Farseer \\
\midrule
Chinchilla~\citep{hoffmann2022training}      & \msci{0.124}{0.122}{0.167} & \msci{0.179}{0.148}{0.215} & \msci{0.385}{0.327}{0.455} & \msci{0.063}{0.041}{0.094} & \msci{0.027}{0.018}{0.032} & \msci{0.109}{0.095}{0.140} & \msci{0.037}{0.013}{0.060} & \msci{0.114}{0.084}{0.148} & \msci{0.015}{0.014}{0.018} \\
Muennighoff~\citep{muennighoff2023scaling}   & \msci{0.156}{0.115}{0.171} & \underline{\msci{0.121}{0.060}{0.283}} & \msci{0.209}{0.138}{0.369} & \msci{0.063}{0.034}{0.157} & \msci{0.021}{0.009}{0.026} & \msci{0.069}{0.037}{0.101} & \msci{0.024}{0.008}{0.055} & \msci{0.104}{0.047}{0.145} & \msci{0.014}{0.013}{0.065} \\
M4 ($N$-axis)~\citep{alabdulmohsin2022revisiting} & \msci{1.914}{1.884}{1.946} & \msci{0.833}{0.652}{1.094} & \msci{1.227}{1.032}{1.290} & \msci{0.544}{0.498}{0.648} & \msci{0.085}{0.075}{0.095} & \msci{0.191}{0.132}{0.234} & \msci{0.075}{0.047}{0.163} & \msci{0.317}{0.256}{0.405} & \msci{0.104}{0.091}{0.117} \\
M4 ($D$-axis)~\citep{alabdulmohsin2022revisiting} & \msci{0.193}{0.186}{0.313} & \msci{0.262}{0.244}{0.297} & \msci{0.361}{0.350}{0.414} & \msci{0.222}{0.210}{0.271} & \msci{0.038}{0.036}{0.054} & \msci{0.101}{0.097}{0.129} & \msci{0.121}{0.071}{0.183} & \msci{0.081}{0.078}{0.097} & \msci{0.064}{0.062}{0.075} \\
BNSL $k{=}1$~\citep{caballero2023broken}     & \msci{0.171}{0.168}{0.175} & \msci{0.294}{0.244}{0.351} & \msci{0.357}{0.349}{0.563} & \msci{0.220}{0.209}{0.279} & \msci{0.038}{0.036}{0.053} & \msci{0.102}{0.096}{0.135} & \msci{0.128}{0.073}{0.195} & \msci{0.091}{0.079}{0.131} & \msci{0.064}{0.062}{0.076} \\
BNSL $k{=}2$~\citep{caballero2023broken}     & \msci{0.173}{0.168}{0.444} & \msci{0.283}{0.244}{0.352} & \msci{0.368}{0.349}{0.610} & \msci{0.221}{0.209}{0.285} & \msci{0.038}{0.036}{0.056} & \msci{0.103}{0.097}{0.135} & \msci{0.127}{0.070}{0.304} & \msci{0.090}{0.079}{0.136} & \msci{0.064}{0.061}{0.076} \\
Farseer~\citep{li2025farseer}                & \msci{1.484}{1.449}{1.513} & \msci{0.883}{0.668}{1.073} & \msci{0.471}{0.371}{0.591} & \msci{0.062}{0.035}{0.146} & \msci{0.013}{0.007}{0.030} & \msci{0.092}{0.058}{0.149} & \msci{0.037}{0.014}{0.114} & \msci{0.101}{0.080}{0.136} & \textbf{\msci{0.007}{0.006}{0.021}} \\
\midrule
\textbf{Ours} (Eq.~\eqref{eq:ours})          & \msci{0.137}{0.128}{0.147} & \textbf{\msci{0.078}{0.052}{0.120}} & \textbf{\msci{0.171}{0.145}{0.200}} & \textbf{\msci{0.056}{0.035}{0.090}} & \underline{\msci{0.008}{0.006}{0.012}} & \msci{0.046}{0.039}{0.073} & \textbf{\msci{0.014}{0.005}{0.091}} & \textbf{\msci{0.033}{0.028}{0.043}} & \textbf{\msci{0.007}{0.006}{0.008}} \\
Ours, no wrapper                             & \textbf{\msci{0.097}{0.095}{0.100}} & \msci{0.142}{0.105}{0.193} & \msci{0.187}{0.169}{0.211} & \msci{0.071}{0.055}{0.095} & \textbf{\msci{0.007}{0.004}{0.012}} & \textbf{\msci{0.043}{0.036}{0.065}} & \msci{0.031}{0.017}{0.073} & \msci{0.115}{0.086}{0.151} & \textbf{\msci{0.007}{0.006}{0.009}} \\
Ours, no overfitting term                    & \msci{1.272}{1.245}{1.298} & \msci{0.859}{0.650}{1.093} & \msci{0.461}{0.384}{0.574} & \msci{0.070}{0.048}{0.112} & \msci{0.026}{0.016}{0.035} & \msci{0.120}{0.073}{0.163} & \msci{0.017}{0.006}{0.054} & \textbf{\msci{0.033}{0.029}{0.043}} & \msci{0.013}{0.012}{0.016} \\
Ours, $1{-}e^{-h}$ wrapper                   & \underline{\msci{0.104}{0.099}{0.109}} & \msci{0.126}{0.085}{0.186} & \underline{\msci{0.183}{0.163}{0.211}} & \underline{\msci{0.059}{0.044}{0.096}} & \underline{\msci{0.008}{0.005}{0.013}} & \msci{0.044}{0.037}{0.079} & \msci{0.023}{0.008}{0.084} & \msci{0.057}{0.044}{0.078} & \textbf{\msci{0.007}{0.006}{0.008}} \\
Ours, single-exp $(N/D)^\gamma$              & \msci{0.756}{0.739}{0.783} & \msci{0.392}{0.180}{0.608} & \msci{0.282}{0.238}{0.418} & \msci{0.152}{0.112}{0.210} & \underline{\msci{0.008}{0.004}{0.014}} & \textbf{\msci{0.043}{0.038}{0.061}} & \textbf{\msci{0.014}{0.005}{0.063}} & \textbf{\msci{0.033}{0.029}{0.044}} & \textbf{\msci{0.007}{0.006}{0.008}} \\
\bottomrule
\end{tabular}
}
\end{table}

\clearpage

\captionsetup{hypcap=false}

\subsection{Training trajectory plots}
\label{app:plots-trajectories}

The figures below show the empirical running-min validation-loss trajectories across $(N, D, C)$ cells for each of our four training experiments, decomposed into per-$D$ and per-$N$ slices. Figure~\ref{fig:mnist-facets} in the main paper showed a six-cell selection of the MNIST surface; Figure~\ref{fig:mnist-facets-full} below shows all cells. For CIFAR-100, Darcy, and TinyStories, which use the Warmup-Stable-Decay schedule of \S\ref{sec:empirical-setup}, each cell decomposes into three components: solid lines are the running-min along the stable run, dashed branches are cooldown forks departing from the stable run at log-spaced compute targets, and points mark the final running-min loss at the end of each fork. MNIST uses no LR schedule, so each cell shows a single solid trajectory.

\vspace*{\fill}
\begin{minipage}{\linewidth}
  \centering
  \includegraphics[width=\linewidth]{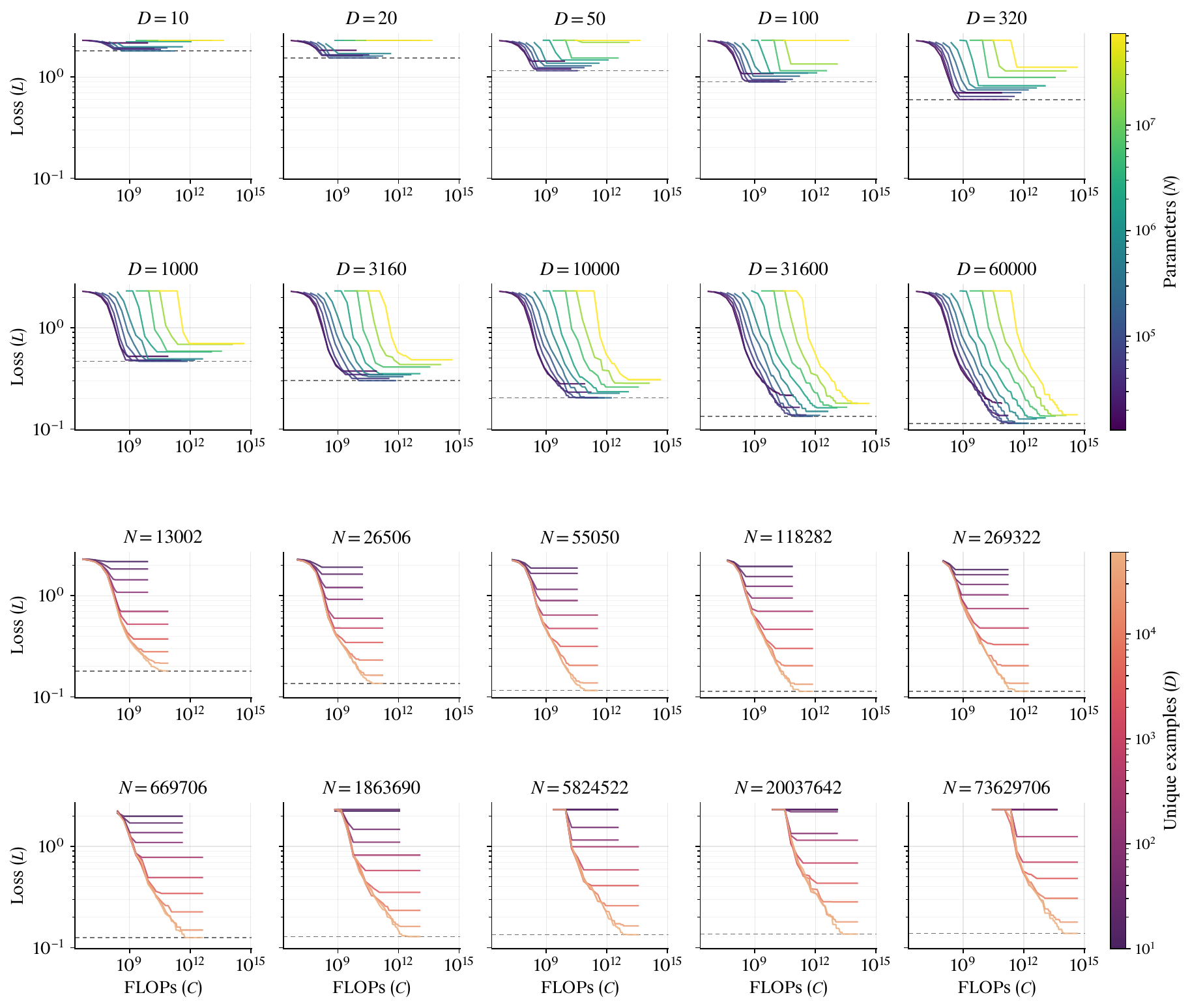}
  \captionof{figure}{MNIST empirical running-min validation-loss surface across all $(N, D)$ cells. \emph{Top}: loss vs FLOPs at fixed $D$, colored by $N$. \emph{Bottom}: same surface at fixed $N$, colored by $D$.}
  \label{fig:mnist-facets-full}
\end{minipage}
\vspace*{\fill}

\clearpage

\vspace*{\fill}
\begin{minipage}{\linewidth}
  \centering
  \includegraphics[width=\linewidth]{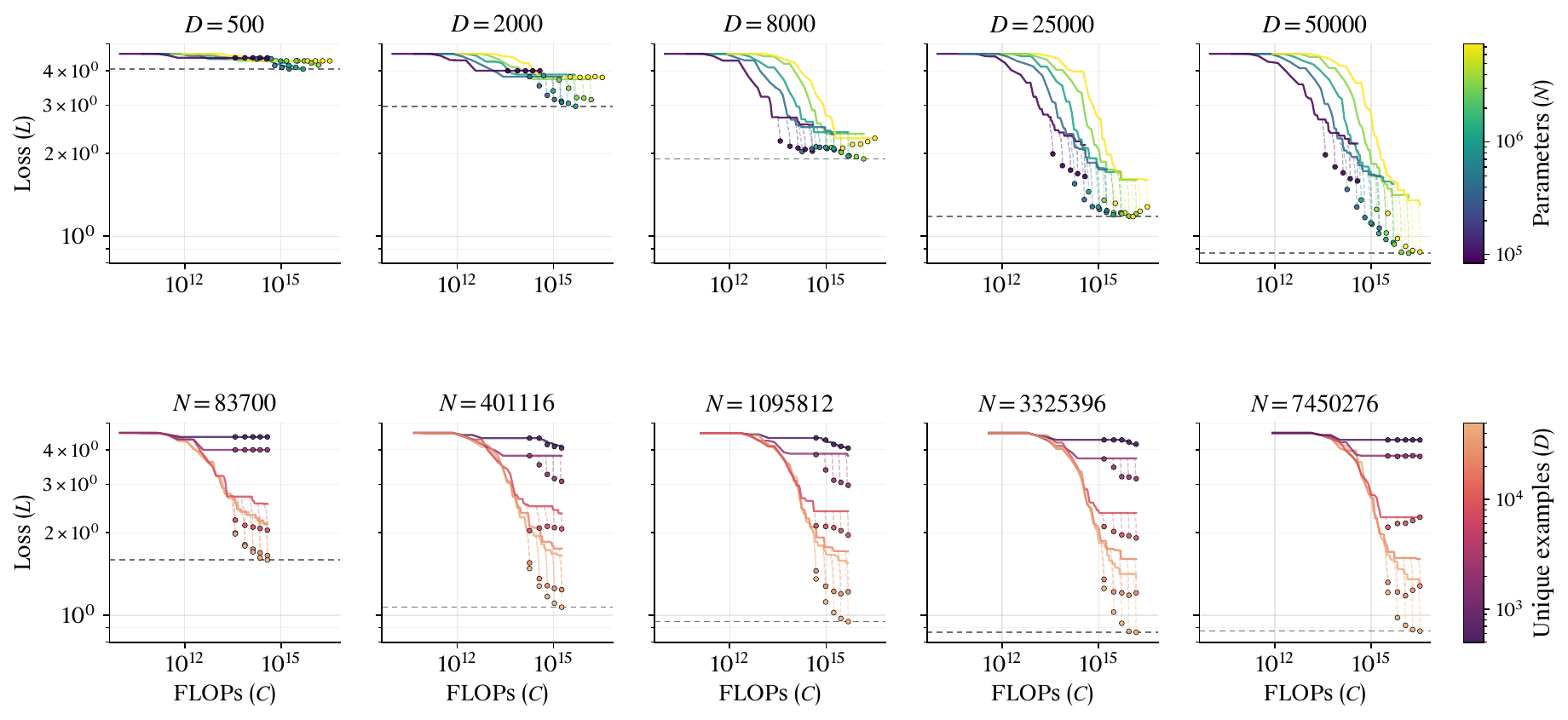}
  \captionof{figure}{CIFAR-100 empirical running-min validation-loss surface. \emph{Top}: loss vs FLOPs at fixed $D$, colored by $N$. \emph{Bottom}: same surface at fixed $N$, colored by $D$.}
  \label{fig:cifar-facets}
\end{minipage}
\vspace*{\fill}

\clearpage

\vspace*{\fill}
\begin{minipage}{\linewidth}
  \centering
  \includegraphics[width=\linewidth]{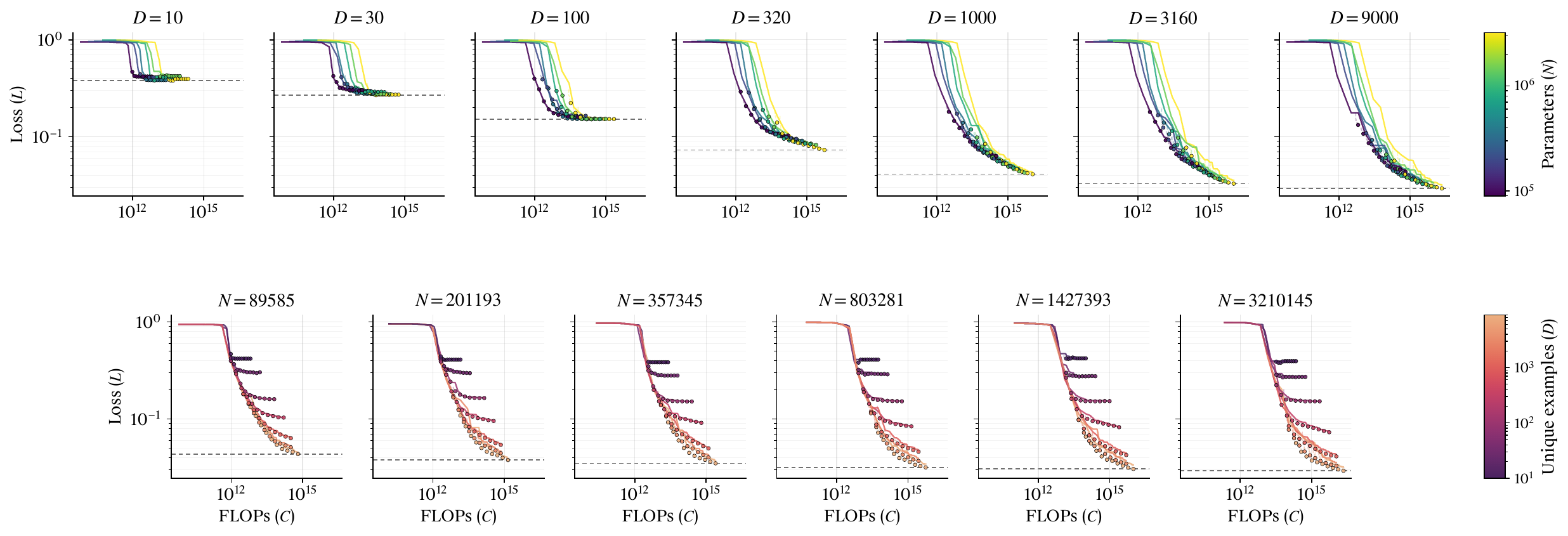}
  \captionof{figure}{Darcy empirical running-min validation-loss surface. \emph{Top}: loss vs FLOPs at fixed $D$, colored by $N$. \emph{Bottom}: same surface at fixed $N$, colored by $D$.}
  \label{fig:darcy-facets}
\end{minipage}
\vspace*{\fill}

\begin{minipage}{\linewidth}
  \centering
  \includegraphics[width=\linewidth]{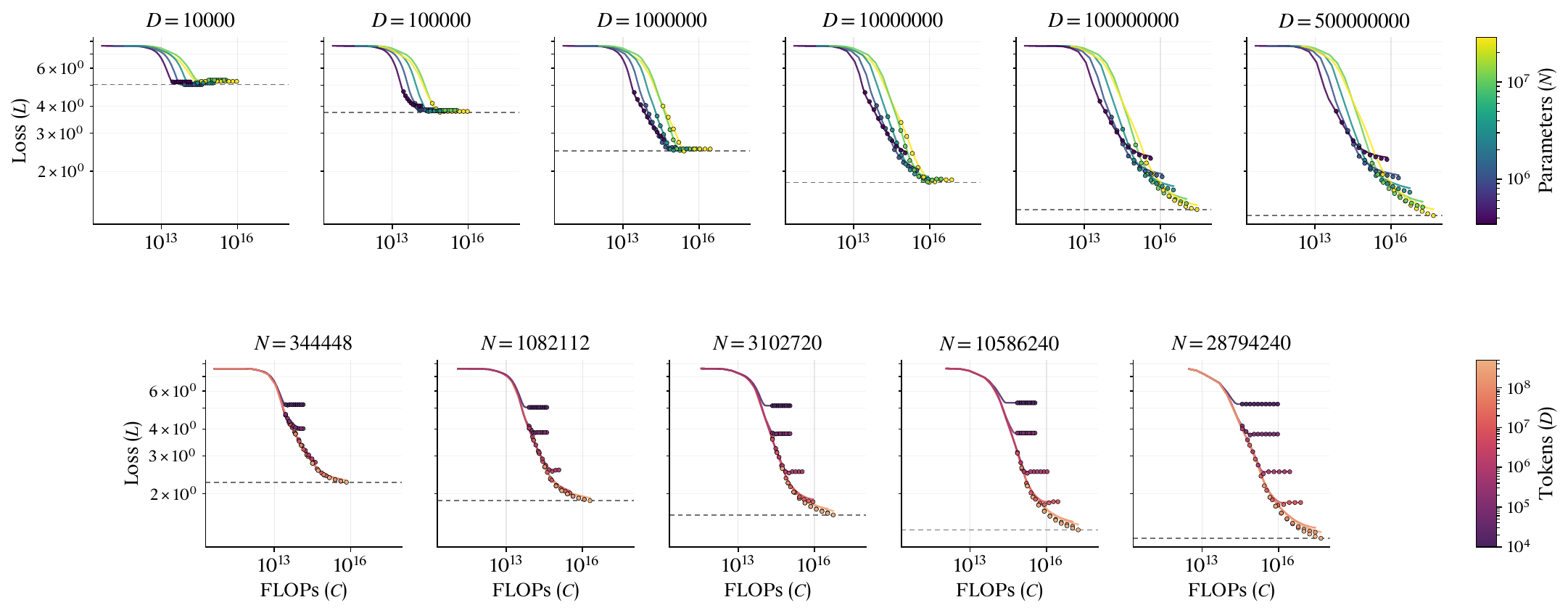}
  \captionof{figure}{TinyStories empirical running-min validation-loss surface. \emph{Top}: loss vs FLOPs at fixed $D$, colored by $N$. \emph{Bottom}: same surface at fixed $N$, colored by $D$.}
  \label{fig:tinystories-facets}
\end{minipage}
\vspace*{\fill}

\clearpage

\subsection{Held-out prediction plots}
\label{app:plots-holdout-grids}

Figure~\ref{fig:obs-vs-pred} in the main paper shows observed-versus-predicted holdout scatter on two representative datasets (Chinchilla and MNIST) under the high-$C$ holdout. The figures below show the full grid: each of our four training experiments (internal) and each of the five published LLM grids (external), under both the high-$C$ and high-$D$ holdouts of \S\ref{sec:empirical-setup}. Rows are datasets, columns are forms; blue points are training cells, orange points are held-out cells. A perfectly calibrated form would place all points on the $y = x$ diagonal.

\vspace*{\fill}
\begin{minipage}{\linewidth}
  \centering
  \includegraphics[width=\linewidth]{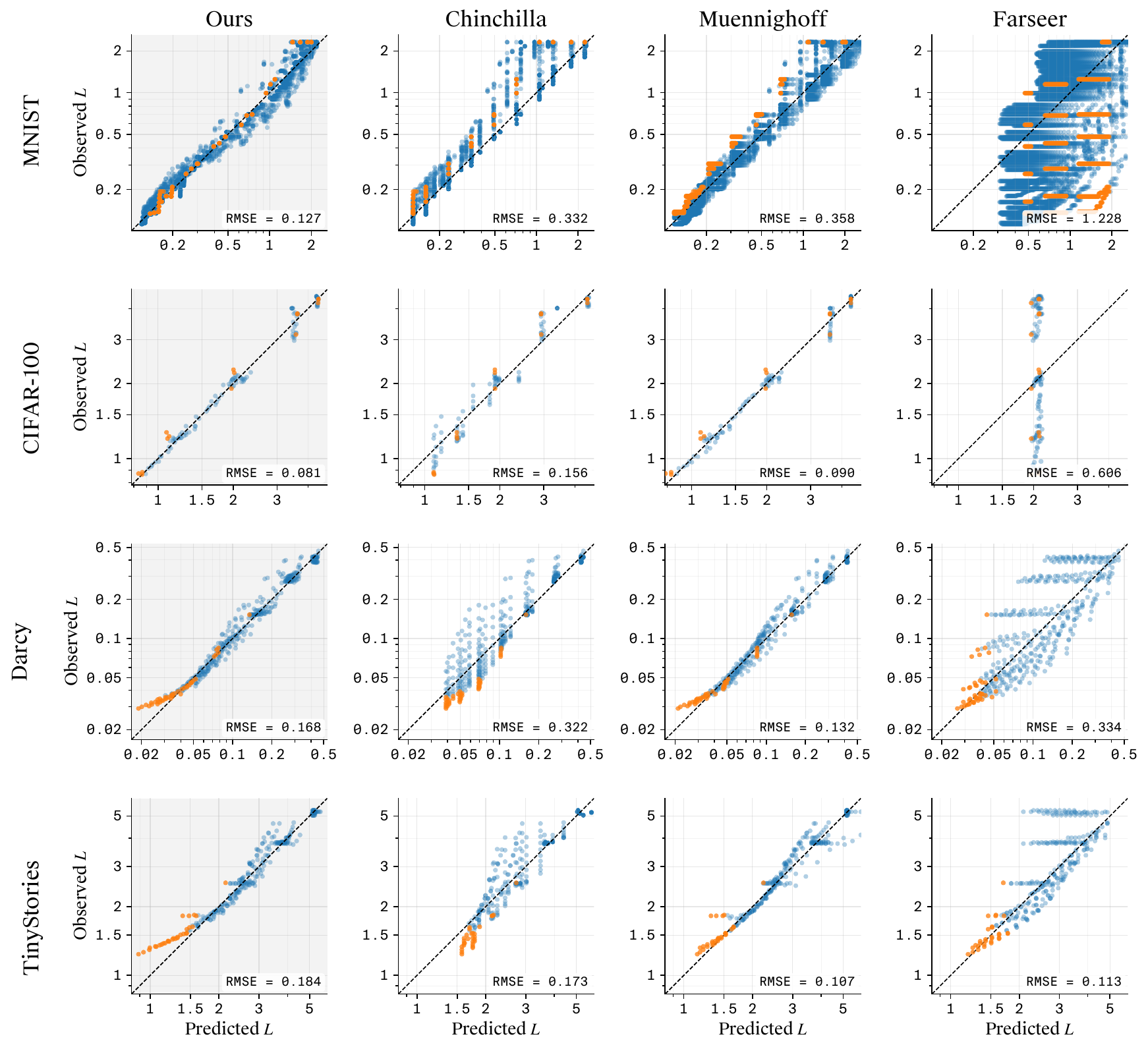}
  \captionof{figure}{\textbf{High-$C$.} Observed vs. predicted loss on the four training experiments (MNIST, CIFAR-100, Darcy, TinyStories) under the high-$C$ holdout.}
  \label{fig:holdout-grid-internal-high-c}
\end{minipage}
\vspace*{\fill}

\clearpage

\vspace*{\fill}
\begin{minipage}{\linewidth}
  \centering
  \includegraphics[width=\linewidth]{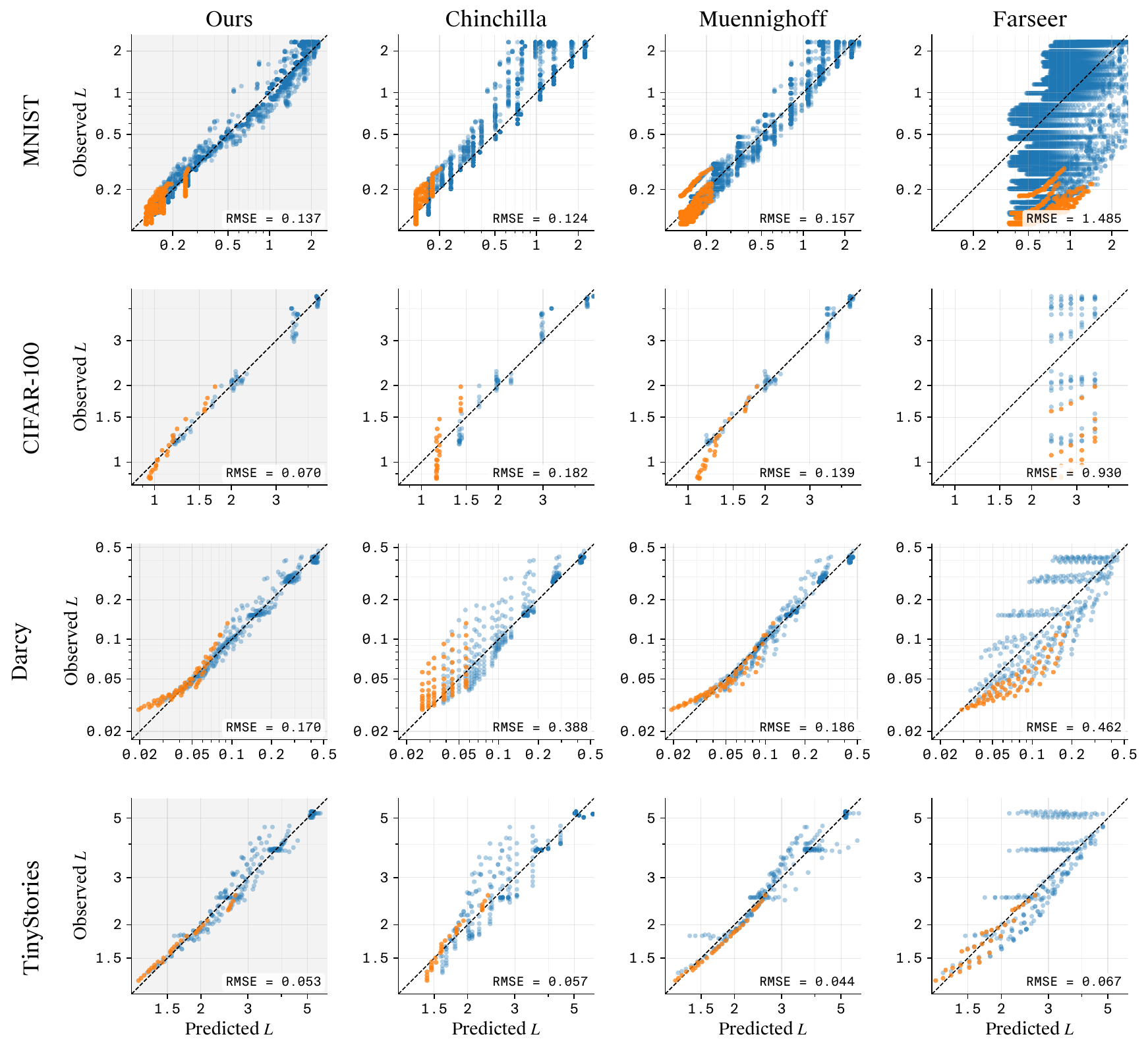}
  \captionof{figure}{\textbf{High-$D$.} Observed vs. predicted loss on the four training experiments under the high-$D$ holdout.}
  \label{fig:holdout-grid-internal-high-d}
\end{minipage}
\vspace*{\fill}

\clearpage

\vspace*{\fill}
\begin{minipage}{\linewidth}
  \centering
  \includegraphics[width=\linewidth]{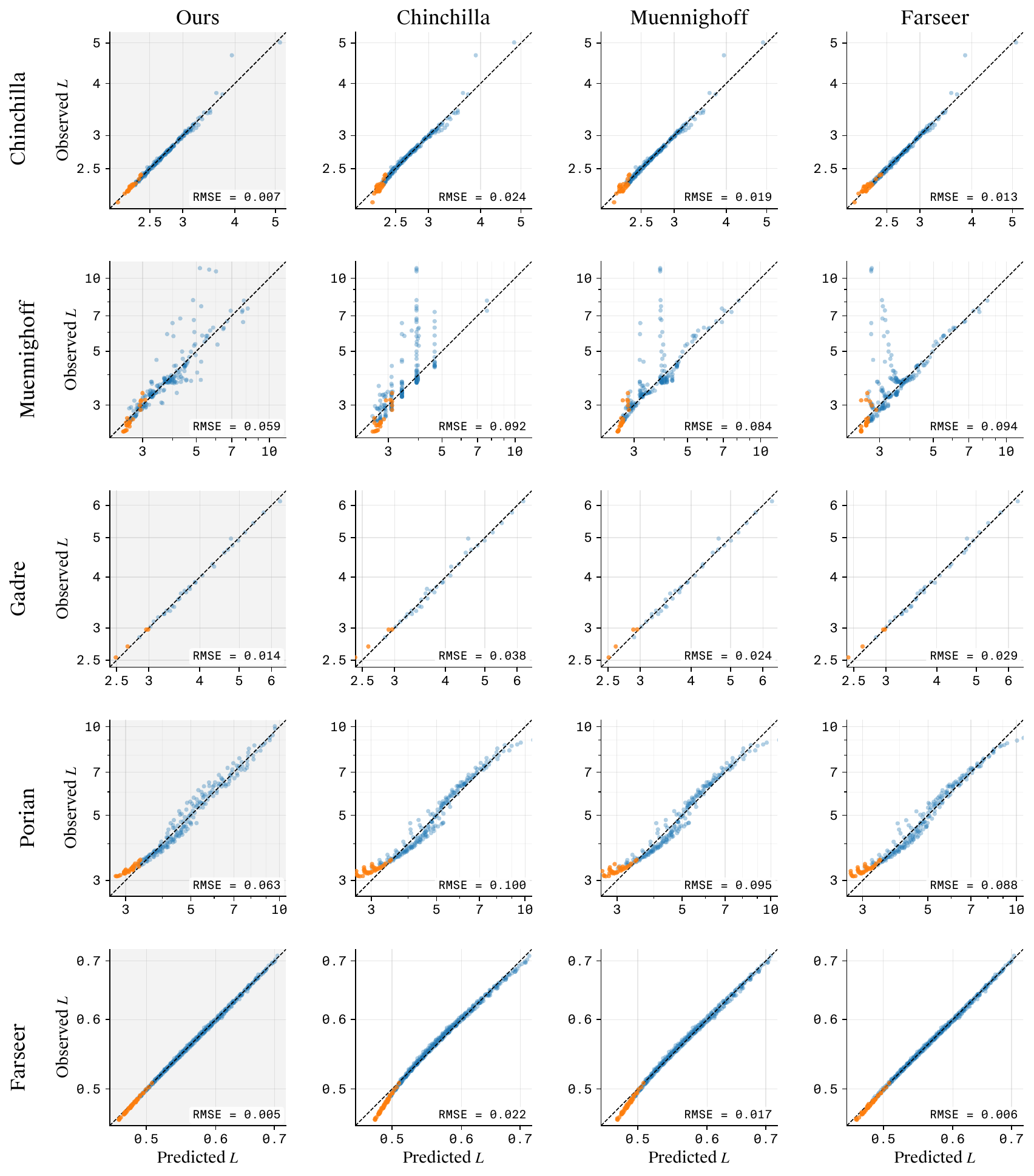}
  \captionof{figure}{\textbf{High-$C$.} Observed vs. predicted loss on the five published LLM grids (Chinchilla, Muennighoff, Gadre, Porian, Farseer) under the high-$C$ holdout.}
  \label{fig:holdout-grid-external-high-c}
\end{minipage}
\vspace*{\fill}

\clearpage

\vspace*{\fill}
\begin{minipage}{\linewidth}
  \centering
  \includegraphics[width=\linewidth]{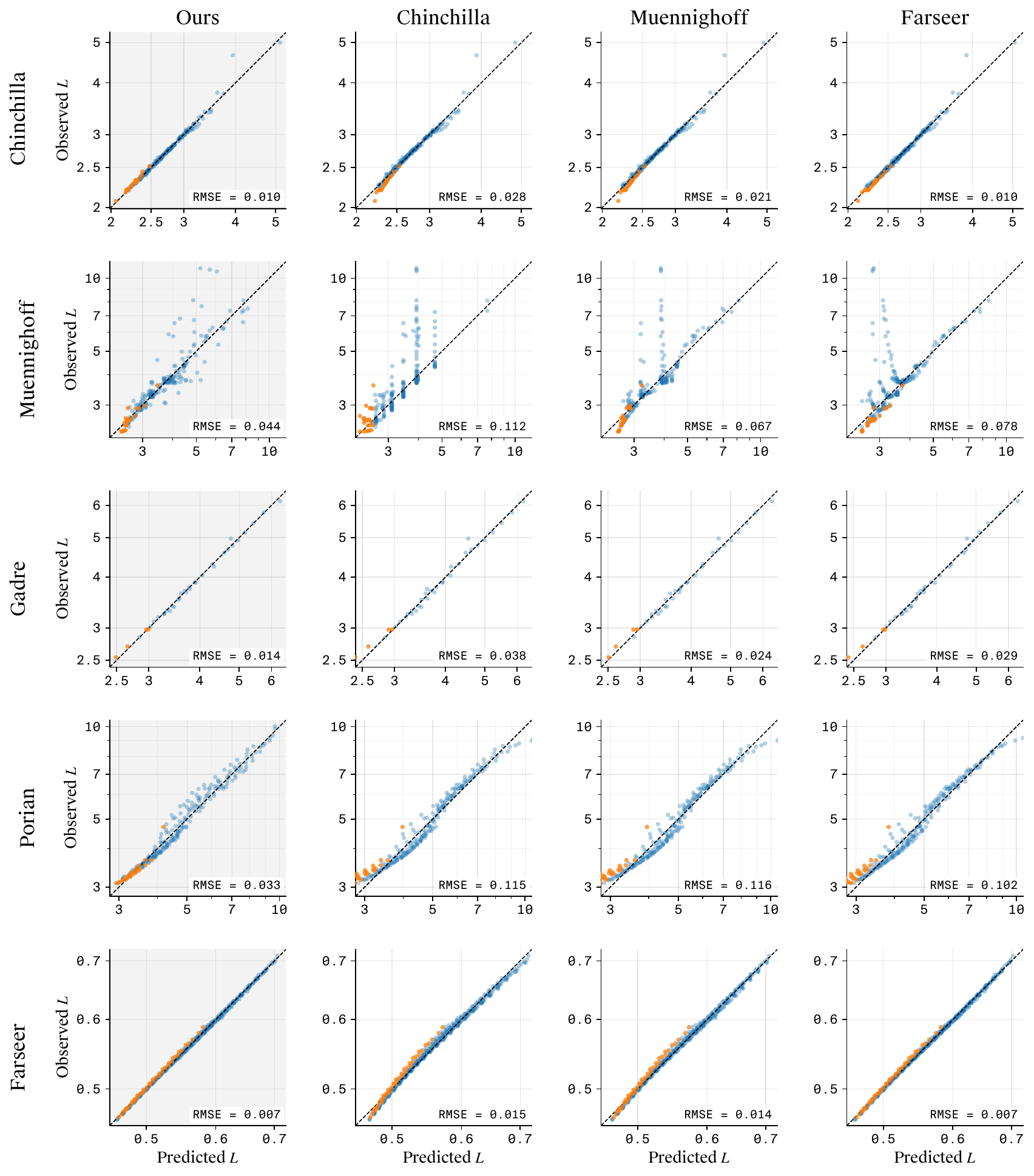}
  \captionof{figure}{\textbf{High-$D$.} Observed vs. predicted loss on the five published LLM grids under the high-$D$ holdout.}
  \label{fig:holdout-grid-external-high-d}
\end{minipage}
\vspace*{\fill}

\end{document}